\documentclass[10pt,twocolumn,letterpaper]{article}

\usepackage[accsupp]{axessibility}
\usepackage[pagenumbers]{cvpr} %

\usepackage{amsmath}
\usepackage{amssymb}
\usepackage{makecell}
\usepackage{microtype}
\usepackage{multirow}
\usepackage{enumitem}
\makeatletter
\renewcommand\paragraph{\@startsection{paragraph}{4}{\z@}%
	{0.75ex \@plus.5ex \@minus.2ex}%
	{-1em}%
	{\normalfont\normalsize\bfseries\maybe@addperiod}}
\newcommand{\maybe@addperiod}[1]{#1\@addpunct{.}}
\makeatother

\newcommand{\padd}{\phantom{0}}

\definecolor{cvprblue}{rgb}{0.21,0.49,0.74}
\usepackage[pagebackref,breaklinks,colorlinks,allcolors=cvprblue]{hyperref}

\def\paperID{6708} %
\def\confName{CVPR}
\def\confYear{2025}

\title{IRIS: Inverse Rendering of Indoor Scenes from Low Dynamic Range Images}

\author{Chih-Hao Lin$^{1,2}$ \quad Jia-Bin Huang$^{1,3}$ \quad Zhengqin Li$^{1}$ \quad Zhao Dong$^{1}$ \quad Christian Richardt$^{1}$ \\ \quad Tuotuo Li$^{1}$ \quad Michael Zollhöfer$^{1}$ \quad  Johannes Kopf$^{1}$ \quad Shenlong Wang$^{2}$ \quad Changil Kim$^{1}$ \\[0.5em]
$^{1}$Meta \quad $^{2}$University of Illinois Urbana-Champaign \quad $^{3}$University of Maryland College Park \\[0.5em]
\Large{\href{https://irisldr.github.io/}{https://irisldr.github.io/}}}

\begin{document}

\newcommand{\todocite}[1]{\textcolor{blue}{Citation needed []}}
\newcommand{\shenlongsay}[1]{\textcolor{blue}{[{\it Shenlong: #1}]}}
\newcommand{\jiabin}[1]{\textcolor{cyan}{[{\it Jia-Bin: #1}]}}

\newcommand{\mfigure}[2]{\includegraphics[width=#1\linewidth]{#2}}
\newcommand{\mpage}[2]
{
\begin{minipage}{#1\linewidth}\centering
#2
\end{minipage}
}

\newcolumntype{L}[1]{>{\raggedright\let\newline\\\arraybackslash\hspace{0pt}}m{#1}}
\newcolumntype{C}[1]{>{\centering\let\newline\\\arraybackslash\hspace{0pt}}m{#1}}
\newcolumntype{R}[1]{>{\raggedleft\let\newline\\\arraybackslash\hspace{0pt}}m{#1}}

\newcommand{\xpar}[1]{\noindent\textbf{#1}\ \ }
\newcommand{\vpar}[1]{\vspace{3mm}\noindent\textbf{#1}\ \ }

\newcommand{\ignorethis}[1]{}
\newcommand{\norm}[1]{\lVert#1\rVert}
\newcommand{\fcseven}{$\mbox{fc}_7$}

\newcommand{\topic}[1]
{
\vspace{1mm}\noindent\textbf{#1}
}

\def\naive{na\"{\i}ve\xspace}
\def\Naive{Na\"{\i}ve\xspace}

\makeatletter
\DeclareRobustCommand\onedot{\futurelet\@let@token\@onedot}
\def\@onedot{\ifx\@let@token.\else.\null\fi\xspace}

\def\iid{\emph{i.i.d}\onedot}
\def\eg{\emph{e.g}\onedot} \def\Eg{\emph{E.g}\onedot}
\def\ie{\emph{i.e}\onedot} \def\Ie{\emph{I.e}\onedot}
\def\cf{\emph{c.f}\onedot} \def\Cf{\emph{C.f}\onedot}
\def\etc{\emph{etc}\onedot} \def\vs{\emph{vs}\onedot}
\def\wrt{w.r.t\onedot} \def\dof{d.o.f\onedot}
\def\etal{et al\onedot}
\makeatother

\definecolor{citecolor}{RGB}{34,139,34}
\definecolor{mydarkblue}{rgb}{0,0.08,1}
\definecolor{mydarkgreen}{rgb}{0.02,0.6,0.02}
\definecolor{mydarkred}{rgb}{0.8,0.02,0.02}
\definecolor{mydarkorange}{rgb}{0.40,0.2,0.02}
\definecolor{mypurple}{RGB}{111,0,255}
\definecolor{myred}{rgb}{1.0,0.0,0.0}
\definecolor{mygold}{rgb}{0.75,0.6,0.12}
\definecolor{myblue}{rgb}{0,0.2,0.8}
\definecolor{mydarkgray}{rgb}{0.66,0.66,0.66}

\newcommand{\myparagraph}[1]{\vspace{-6pt}\paragraph{#1}}

\newcommand{\bbR}{{\mathbb{R}}}
\newcommand{\bK}{\mathbf{K}}
\newcommand{\bX}{\mathbf{X}}
\newcommand{\bY}{\mathbf{Y}}
\newcommand{\bk}{\mathbf{k}}
\newcommand{\bx}{\mathbf{x}}
\newcommand{\by}{\mathbf{y}}
\newcommand{\bhy}{\hat{\mathbf{y}}}
\newcommand{\bty}{\tilde{\mathbf{y}}}
\newcommand{\bG}{\mathbf{G}}
\newcommand{\bI}{\mathbf{I}}
\newcommand{\bg}{\mathbf{g}}
\newcommand{\bS}{\mathbf{S}}
\newcommand{\bs}{\mathbf{s}}
\newcommand{\bM}{\mathbf{M}}
\newcommand{\bw}{\mathbf{w}}
\newcommand{\eye}{\mathbf{I}}
\newcommand{\bU}{\mathbf{U}}
\newcommand{\bV}{\mathbf{V}}
\newcommand{\bW}{\mathbf{W}}
\newcommand{\bn}{\mathbf{n}}
\newcommand{\bv}{\mathbf{v}}
\newcommand{\bq}{\mathbf{q}}
\newcommand{\bR}{\mathbf{R}}
\newcommand{\bi}{\mathbf{i}}
\newcommand{\bj}{\mathbf{j}}
\newcommand{\bp}{\mathbf{p}}
\newcommand{\bt}{\mathbf{t}}
\newcommand{\bJ}{\mathbf{J}}
\newcommand{\bu}{\mathbf{u}}
\newcommand{\bB}{\mathbf{B}}
\newcommand{\bD}{\mathbf{D}}
\newcommand{\bz}{\mathbf{z}}
\newcommand{\bP}{\mathbf{P}}
\newcommand{\bC}{\mathbf{C}}
\newcommand{\bA}{\mathbf{A}}
\newcommand{\bZ}{\mathbf{Z}}
\newcommand{\bff}{\mathbf{f}}
\newcommand{\bF}{\mathbf{F}}
\newcommand{\bo}{\mathbf{o}}
\newcommand{\bc}{\mathbf{c}}
\newcommand{\bT}{\mathbf{T}}
\newcommand{\bQ}{\mathbf{Q}}
\newcommand{\bL}{\mathbf{L}}
\newcommand{\bl}{\mathbf{l}}
\newcommand{\ba}{\mathbf{a}}
\newcommand{\bE}{\mathbf{E}}
\newcommand{\be}{\mathbf{e}}
\newcommand{\bH}{\mathbf{H}}
\newcommand{\bd}{\mathbf{d}}
\newcommand{\br}{\mathbf{r}}
\newcommand{\bb}{\mathbf{b}}
\newcommand{\bh}{\mathbf{h}}

\newcommand{\btheta}{\bm{\theta}}
\newcommand{\bhh}{\hat{\mathbf{h}}}
\newcommand{\ci}{{\cal I}}
\newcommand{\ct}{{\cal T}}
\newcommand{\co}{{\cal O}}
\newcommand{\ck}{{\cal K}}
\newcommand{\cu}{{\cal U}}
\newcommand{\cS}{{\cal S}}
\newcommand{\cQ}{{\cal Q}}
\newcommand{\cT}{{\cal S}}
\newcommand{\cC}{{\cal C}}
\newcommand{\cE}{{\cal E}}
\newcommand{\cF}{{\cal F}}
\newcommand{\cL}{{\cal L}}
\newcommand{\X}{{\cal{X}}}
\newcommand{\Y}{{\cal Y}}
\newcommand{\cH}{{\cal H}}
\newcommand{\cP}{{\cal P}}
\newcommand{\cN}{{\cal N}}
\newcommand{\cU}{{\cal U}}
\newcommand{\cV}{{\cal V}}
\newcommand{\cX}{{\cal X}}
\newcommand{\cY}{{\cal Y}}
\newcommand{\graph}{{\cal H}}
\newcommand{\bayes}{{\cal B}}
\newcommand{\cx}{{\cal X}}
\newcommand{\cg}{{\cal G}}
\newcommand{\cm}{{\cal M}}
\newcommand{\cM}{{\cal M}}
\newcommand{\cG}{{\cal G}}
\newcommand{\cR}{\cal{R}}
\newcommand{\R}{\cal{R}}
\newcommand{\eig}{\mathrm{eig}}

\newcommand{\bbS}{\mathbb{S}}

\newcommand{\D}{{\cal D}}
\newcommand{\bfp}{{\bf p}}
\newcommand{\bfd}{{\bf d}}

\newcommand{\cv}{{\cal V}}
\newcommand{\ce}{{\cal E}}
\newcommand{\cy}{{\cal Y}}
\newcommand{\cz}{{\cal Z}}
\newcommand{\cb}{{\cal B}}
\newcommand{\cq}{{\cal Q}}
\newcommand{\cd}{{\cal D}}
\newcommand{\bcf}{{\cal F}}
\newcommand{\cI}{\mathcal{I}}

\newcommand{\ut}{^{(t)}}
\newcommand{\up}{^{(t-1)}}

\newcommand{\bpi}{\boldsymbol{\pi}}
\newcommand{\bphi}{\boldsymbol{\phi}}
\newcommand{\bPhi}{\boldsymbol{\Phi}}
\newcommand{\bmu}{\boldsymbol{\mu}}
\newcommand{\bSigma}{\boldsymbol{\Sigma}}
\newcommand{\bGamma}{\boldsymbol{\Gamma}}
\newcommand{\bbeta}{\boldsymbol{\beta}}
\newcommand{\bomega}{\boldsymbol{\omega}}
\newcommand{\blambda}{\boldsymbol{\lambda}}
\newcommand{\bkappa}{\boldsymbol{\kappa}}
\newcommand{\btau}{\boldsymbol{\tau}}
\newcommand{\balpha}{\boldsymbol{\alpha}}
\def\bgamma{\boldsymbol\gamma}

\newcommand{\prox}{{\mathrm{prox}}}

\newcommand{\pardev}[2]{\frac{\partial #1}{\partial #2}}
\newcommand{\dev}[2]{\frac{d #1}{d #2}}
\newcommand{\dw}{\delta\bw}
\newcommand{\lab}{\mathcal{L}}
\newcommand{\unlab}{\mathcal{U}}
\newcommand{\ind}{1{\hskip -2.5 pt}\hbox{I}}
\newcommand{\ff}[2]{   \cf_{\prec (#1 \rightarrow #2)}}
\newcommand{\vv}[2]{   \cv_{\prec (#1 \rightarrow #2)}}
\newcommand{\dd}[2]{   \delta_{#1 \rightarrow #2}}
\newcommand{\ld}[2]{   \lambda_{#1 \rightarrow #2}}
\newcommand{\en}[2]{  \bD(#1|| #2)}
\newcommand{\ex}[3]{  \bE_{#1 \sim #2}\left[ #3\right]} 
\newcommand{\exd}[2]{  \bE_{#1 }\left[ #2\right]}

\newcommand{\se}[1]{\mathfrak{se}(#1)}
\newcommand{\SE}[1]{\mathbb{SE}(#1)}
\newcommand{\so}[1]{\mathfrak{so}(#1)}
\newcommand{\SO}[1]{\mathbb{SO}(#1)}

\newcommand{\poselow}{\xi}
\newcommand{\pose}{\bm{\poselow}}
\newcommand{\linpose}{\pose^\ell}
\newcommand{\cbpose}{\pose^c}
\newcommand{\rateparam}{v_i}
\newcommand{\bapose}{\bm{\poselow}_i}
\newcommand{\trackingpose}{\bm{\poselow}}
\newcommand{\rotlow}{\omega}
\newcommand{\rot}{\bm{\rotlow}}
\newcommand{\translow}{v}
\newcommand{\trans}{\bm{\translow}}
\newcommand{\hnorm}[1]{\left\lVert#1\right\rVert_{\gamma}}
\newcommand{\lnorm}[1]{\left\lVert#1\right\rVert}
\newcommand{\barate}{v_i}
\newcommand{\trackingrate}{v}
\newcommand{\imgpt}{\mathbf{u}_{i,k,j}}
\newcommand{\mappt}{\mathbf{X}_{j}}
\newcommand{\timet}[1]{\bar{t}_{#1}}
\newcommand{\mf}[1]{\text{MF}_{#1}}
\newcommand{\kmf}[1]{\text{KMF}_{#1}}
\newcommand{\Exp}{\text{Exp}}
\newcommand{\Log}{\text{Log}}

\newcommand{\shiftleft}[2]{\makebox[0pt][r]{\makebox[#1][l]{#2}}}
\newcommand{\shiftright}[2]{\makebox[#1][r]{\makebox[0pt][l]{#2}}}

\begin{figure}
\twocolumn[{
    \renewcommand\twocolumn[1][]{#1}
    \maketitle
    \input{figures/new_teaser} 
    }]
\end{figure}

\begin{abstract}
Inverse rendering seeks to recover 3D geometry, surface material, and lighting from captured images, enabling advanced applications such as novel-view synthesis, relighting, and virtual object insertion.
However, most existing techniques rely on high dynamic range (HDR) images as input, limiting accessibility for general users.
In response, we introduce IRIS, an inverse rendering framework that recovers the physically based material, spatially-varying HDR lighting, and camera response functions from multi-view, low-dynamic-range (LDR) images.
By eliminating the dependence on HDR input, we make inverse rendering technology more accessible.
We evaluate our approach on real-world and synthetic scenes and compare it with state-of-the-art methods. Our results show that IRIS effectively recovers HDR lighting, accurate material, and plausible camera response functions, supporting photorealistic relighting and object insertion.

\end{abstract}
 
\section{Introduction}
\label{sec:intro}

Physically based inverse rendering enables reconstructing realistic material properties and lighting in scenes. %
This decomposition is valuable for various applications, including relighting, material editing, and realistic object insertion.
However, most existing inverse rendering methods require input images with high dynamic range (HDR) to capture the full light transport in the scene.
This poses a significant barrier to broader adoption, 
as HDR capture typically requires specialized hardware or merging multiple aligned LDR images through exposure bracketing \cite{ReinhWPDHM2010}.

Most common imaging sensors do not capture sufficient dynamic range for scenes' light-emitting and dark regions.
Moreover, cameras often convert raw sensor readings to 8-bit low-dynamic-range (LDR) images for storage and transmission.
The non-linear mapping and quantization lead to the additional loss of lighting information.
The complex light transport in indoor environments makes reconstructing the original HDR lighting -- critical for faithful inverse rendering -- even more challenging.
If material and lighting estimation could be achieved from a ``casual'' capture using standard devices like cameras or phones, inverse rendering would become far more accessible to a wider range of users.

Many state-of-the-art inverse rendering methods rely on HDR inputs to accurately recover material properties and lighting \cite{wu2023factorized, zhang2023neilf++, yu2023milo}. Some approaches attempt to overcome this issue by taking LDR images as inputs and solving for the lighting.
However, these methods often make simplifying assumptions, such as infinitely distant light sources \cite{wang2023fegr}, requiring additional inputs like emitter masks \cite{ZhuHYLLXWTHBW2023}, or neglecting multi-bounce light transport \cite{ZhuHYLLXWTHBW2023, yao2022neilf, zhang2023neilf++}.
As a result, these methods struggle to handle the complex light transport in indoor scenes and often fail to produce high-quality surface material and lighting. %

To address these challenges, we introduce IRIS, an inverse rendering method for indoor scenes, using only multi-view LDR images. %
IRIS models tone mapping (i.e., HDR-to-LDR conversion), allowing us to work directly with LDR inputs.
IRIS automatically identifies emitters and reconstructs spatially varying HDR lighting through physically based inverse rendering, which is crucial for faithful material estimation. %
Jointly estimating lighting, surface material, and camera response function (CRF) leads to unstable optimization due to ambiguities. To tackle this, we design a novel optimization strategy that effectively resolves these ambiguities, enabling high-quality estimation.
Our core contributions are:
\begin{enumerate}[topsep=0pt]
\item
IRIS faithfully estimates spatially-varying HDR lighting, physically-based materials, and camera response function from LDR images and outperforms several state-of-the-art inverse rendering methods.

\item
We explicitly model LDR image formation in our (inverse) rendering pipeline so that LDR images can be used directly, %
broadening the accessibility of high-quality inverse rendering.

\item
Finally, IRIS is extensively evaluated on synthetic and real-world scenes, demonstrating diverse and realistic view synthesis, relighting, and object insertion.
\end{enumerate}

\section{Related Work}

\begin{figure}[t]
    \centering
    \includegraphics[width=0.8\linewidth]{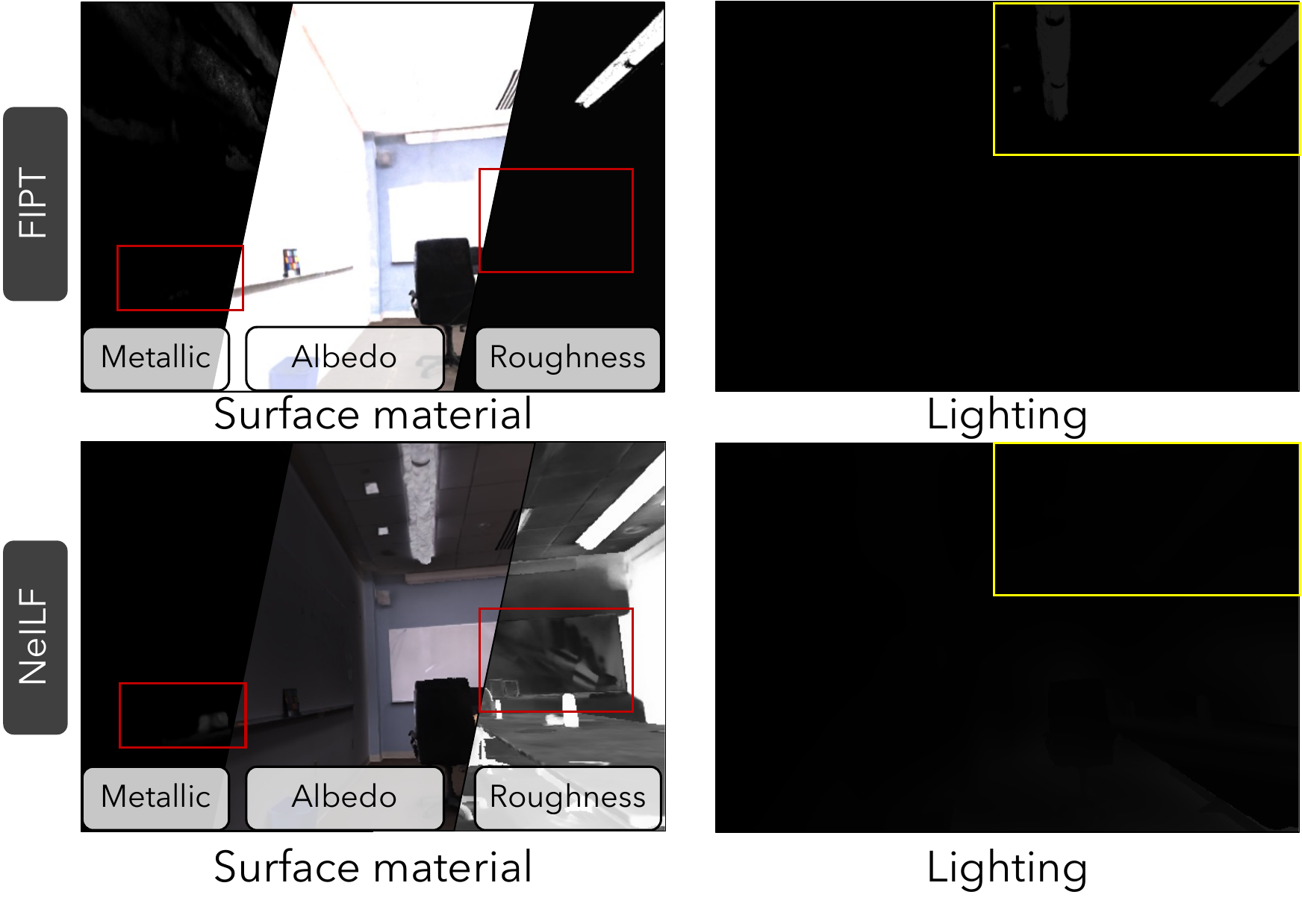}
    \vspace{-5mm}
    \caption{\textbf{Limitation of SOTA.}
    A typical image formation process causes the loss of lighting information, posing challenges in inverse rendering. FIPT \cite{wu2023factorized} assumes HDR input and NeILF \cite{yao2022neilf} ignores multi-bounce light transport. Both methods fail to estimate accurate material (red boxes) and HDR lighting (yellow boxes). We demonstrate significantly better results (\cref{fig:teaser}).
    }
    \label{fig:motivation}
    \vspace{-6mm}
\end{figure}

\topic{Data-driven inverse rendering.}
Inverse rendering seeks to reconstruct a scene's geometry, material, and lighting from single or multiple images.
This process is notoriously challenging and ill-posed due to the inherent ambiguity.
Numerous methods employ deep priors learned from large-scale datasets for different tasks, including intrinsic image decomposition \cite{li2018cgintrinsics,li2018learning}, SVBRDF estimation \cite{deschaintre2018single,li2018materials,li2017modeling,ZhuLMPC2022, ZengDGHHLYH2024}, lighting estimation \cite{gardner2017learning,li2020inverse,srinivasan2020lighthouse}, lighting editing \cite{li2022physically} and relighting \cite{PhiliMGD2021}.
These learning-based methods take a single or a few images \cite{BiXSMSHHKR2020,srinivasan2020lighthouse} as inputs, reducing capturing requirements of classical measurement-based methods \cite{goldman2009shape}. 
However, high-quality datasets are essential for data-driven methods to achieve compelling generalization ability.
Synthetic datasets are often used for training because ground-truth labeling is challenging to obtain in real environments.
Despite efforts to create photorealistic synthetic datasets \cite{li2018cgintrinsics,li2021openrooms,roberts:2021}, the inherent diversity of real-world environments, especially in complex indoor scenes, still causes a substantial domain gap. 
Furthermore, these methods do not reconstruct complete 3D scenes and fail to render consistent novel views under novel illuminations.
Our method utilizes prediction from a data-driven method \cite{ZhuLMPC2022} for initialization and reconstructs 3D geometry, material and lighting with a carefully designed optimization framework, demonstrating superior performance for real-world scenes.

\topic{Optimization-based inverse rendering.} 
Differentiable path tracing \cite{Mitsuba3} enables the direct optimization of scene parameters with physically-based multi-bounce light transport.
Recent methods, including %
MILO \cite{yu2023milo}, IPT \cite{azinovic2019inverse}, and follow-up work \cite{nimierdavid2021material}
exploit this
to jointly optimize BRDF and lighting emission, accounting for global light transport.
FIPT \cite{wu2023factorized} utilizes a factored light transport formulation for BRDF estimation and emitter detection.
Nevertheless, these methods assume piecewise constant materials or are designed to take high-dynamic range images as input, and the robustness is limited in real-world scenes.
Several recent NeRF-based approaches \cite{zhang2021nerfactor, jin2023tensoir, ZhangLWBS2021, munkberg2022extracting, hasselgren2022shape, sun2023neural, wang2023fegr, lin2023urbanir} parameterize materials as a neural field and optimize with a volume rendering formulation.
They tend to assume distant lighting and cannot handle spatially varying lighting in indoor environments well.
NeILF \cite{yao2022neilf} and NeILF++ \cite{zhang2023neilf++} represent spatially-varying lighting as a neural field, I$^2$-SDF \cite{ZhuHYLLXWTHBW2023} constructs lighting from given emitter masks.
While handling more complex lighting, these works require additional assumptions (e.g., HDR input, emitter masks), and single-bounce ray tracing cannot model complex light transport.
In contrast, our method automatically identifies emitters and reconstructs HDR lighting from casual (LDR) images, while achieving high-quality material and lighting estimation that outperforms prior work in this practical scenario.

\topic{CRF and HDR estimation.} 
The complete information about the illumination is usually lost within the photography pipeline. 
Estimating the HDR lighting and CRF enables us to edit photos and achieve realistic object insertion.
\citet{debevec1997recovering} propose to take multiple photos with varying exposure levels to recover the CRF and HDR image.
Some previous works \cite{liu2020single, marnerides2018expandnet, yang2018image, zhang2017learning} predict HDR images with a data-driven approach by learning the LDR--HDR mapping from a single image.
However, these methods rely on HDR data for training, and the estimation is inconsistent across views and scenes.
To handle the HDR--LDR mapping, FIPT \cite{wu2023factorized} and RawNeRF \cite{MildeHMSB2022} assume a known gamma correction function; NeILF++ \cite{zhang2023neilf++} learn a single gamma correction parameter; HDR-NeRF \cite{HuangZFLWW2022} parameterizes tone-mapping function with an MLP.
However, these works are designed to take HDR images or LDR with multiple exposure levels as input.
We leverage physically-based rendering to recover spatially varying BRDF, HDR illumination, and CRF from LDR images.
Our method generates realistic view synthesis, relighting, and object insertion. %

\section{Background: LDR image formation}
\label{sec:background}

Real-world radiance exhibits high dynamic range (HDR) and contains bright light sources and dark regions. To preserve the complete lighting information, HDR images are typically produced with a high-end camera, laborious exposure bracketing \cite{debevec1997recovering}, and are stored in specialized format (16 or 32 bits per color channel). However, capturing HDR is often impractical for casual users and is typically absent in standard photography assets. Due to the limitations of standard sensors and displays,  the radiance is frequently processed with multiple stages to produce low dynamic range (LDR) images stored in more lightweight formats (such as JPEG or PNG at 8 bits and RAW at 12 bits per color channel). We describe the major steps in this process below.

\topic{Dynamic range clipping.}
Given the scene irradiance $E$ and exposure time $\Delta t$, the observed radiance is clipped at a certain maximum value due to the sensor limitation, which could be formulated as:
$Z_c = \min (E\Delta t, 1)$. Due to the clipping operation, the lighting information is lost for over-exposed/saturated pixels.

\topic{Non-linear mapping of CRF.}
To enhance the image's visual quality and match the human perception of the scene, the clipped intensity is further transformed with a non-linear camera response function (CRF) to produce LDR pixel values:
$Z = \texttt{CRF}(Z_c) = \texttt{CRF}(\min (E\Delta t, 1))$.
Please refer to \citet{debevec1997recovering} for a more detailed imaging pipeline explanation.

\topic{Challenges in inverse rendering.}
Real-world scenes often exhibit extreme variances in radiance, making it challenging to capture these scenes in LDR images without under-exposing or saturating them. Due to the clipping and non-linear CRF mapping, critical lighting information is lost, posing challenges in previous works, illustrated in \cref{fig:motivation}.
In this work, we propose an inverse rendering framework that jointly recovers HDR lighting and CRF from LDR images.

\begin{figure*}[t]
    \centering
    \includegraphics[width=1.0\textwidth]{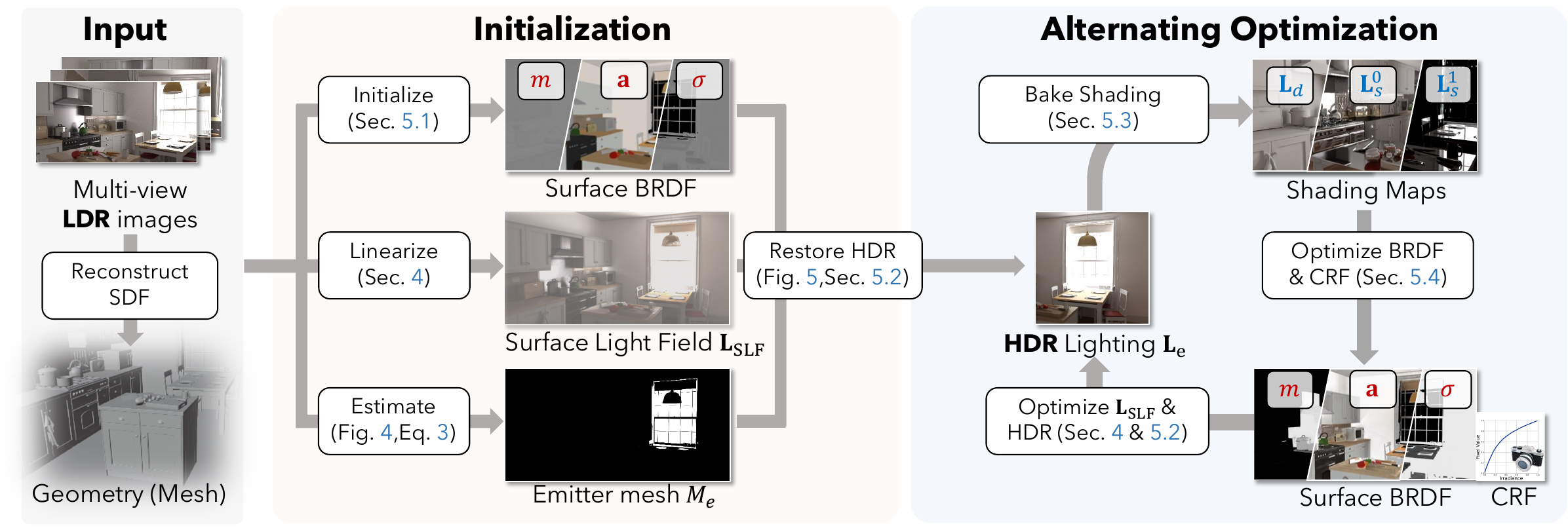}
    \vspace{-5mm}
    \caption{\textbf{Framework Overview.}
        Given multi-view posed LDR images and a surface mesh, our inverse rendering pipeline is divided into two main stages.
        In the initialization stage, we initialize the BRDF (\cref{sec:brdf_init}), extract a surface light field (\cref{sec:representation}), and estimate emitter geometry (\cref{eq:emitter_mask}).
        In the optimization stage, we first recover HDR radiance from the LDR input (\cref{sec:hdr_restoration}), then bake shading maps (\cref{sec:shading}), and jointly optimize BRDF and CRF parameters (\cref{sec:train_brdf}).
        The improved parameters are used to refine the emission again.
        These three steps are repeated until convergence.
        }
    \label{fig:framework}
    \vspace{-4mm}
\end{figure*}

\section{Representation and rendering}\label{sec:representation}

\topic{Camera response function (CRF).}
The CRF is crucial for recovering original radiance from 8-bit LDR images. However, the CRF varies with each camera and is often unknown to users.
Thus, deducing the CRF from LDR image observations is critical to our solution.
We use  Gross \etal.'s \cite{GrossN2004} empirical model of response (EMoR) for CRF modeling \cite{liu2020single}.
They collected a database of 201 real-world CRFs of various devices and parameterized them as vectors.
A mean curve $\bar{\bg}$ and PCA basis $\bg_b$ are calculated from the database, and we parameterize a learnable CRF as 
\begin{equation}\label{eq:crf_parameter}
    \bg = \bar{\bg} + \displaystyle\Sigma_b w_b\bg_b, \bg \in \bbR^{1024} \text{,}  
\end{equation}
where $\{w_b\}$ weigh each basis curve and are learnable parameters.
The non-linear function is discretized as $\texttt{CRF}(x) = \texttt{LERP}(\{g_k\}_k, x)$, which linearly interpolates the uniformly sampled CRF $\{g_k\}_{k=0}^{1023} \in [0, 1]^{1024}$, independently per color channel.
We make this design choice because it covers most of the CRF space, and the simplicity makes it easy to enforce regularization, which brings huge advantage in the complex optimization of inverse rendering.

\topic{Spatially-varying lighting.}
While environment maps handle object-centric and outdoor scenes well \cite{jin2023tensoir, ZhangLWBS2021, SriniDZTMB2021, wang2023fegr}, they cannot capture complex spatially-varying lighting of indoor scenes.
To handle this, we define direct lighting on the scene mesh and measure emission as view-independent radiance: $\bL_\text{e}(\bx) \!\in\! \bbR^3$.
The parameterization of $\bL_\text{e}$ is detailed in our supplementary material.
We also define an emitter mask $M_\text{e}(\bx) \!\in\! \{0, 1\}$ to denote whether a surface point $\bx$ is on an emitter.
Global illumination is represented as a view-independent surface light field (SLF): $\bL_{\text{SLF}}(\bx) \!\in\! \bbR^3$.
The sRGB pixel colors of input LDR images are linearized with an inverse CRF mapping: $\bI_{\text{linear}} \!=\! \texttt{CRF}^{-1}(\bI_{\text{LDR}})$.
If a voxel contains no surface, it is set to zero; otherwise, it collects the average linearized radiances across all input images with ray casting.
This greatly accelerates the approximation of global illumination without requiring multi-bounce ray tracing.

\topic{Material BRDF.}
We represent spatially-varying materials using the Cook–Torrance BRDF \cite{cook1982reflectance}.
The surface albedo $\ba(\bx) \!\in\! \bbR^3$, roughness $\sigma(\bx) \!\in\! \bbR$, and metallicity $m(\bx) \!\in\! \bbR$ model the diffuse reflectance $\bk_\text{d} \!=\! \ba \!\cdot\! (1 \!-\! m)$ and specular reflectance $\bk_\text{s} \!= 0.04 \!\cdot\! (1 \!-\! m)$, and are parameterized as a neural field $\mathbf{f} \colon \bx \mapsto (\ba, \sigma, m)$, implemented with multi-resolution hash encoding and an MLP \cite{muller2022instant}.

\topic{Factorized light transport.}
To render images from scene geometry, material, and lighting information, We follow the rendering equation \cite{kajiya1986rendering} to model physically-based light transport for realistic rendering:
\begin{equation}\label{eq:rendering}
        \bL_\text{o}(\bx, \bomega_\text{o}) \hspace{-.2em}=\hspace{-.2em}\bL_\text{e}(\bx, \bomega_\text{o}) + \hspace{-.2em}\int_{\Omega^+}\hspace{-.5em}\bL_\text{i}(\bx, \bomega_\text{i})f(\bx, \bomega_\text{i}, \bomega_\text{o})d\bomega_\text{i} \text{,}
\end{equation}
where $\bL_\text{o}$ is the radiance observed along a ray at a 3D position $\bx$ in direction $\bomega_\text{o}$,
$\bL_\text{e}$ is the emission term,
$\bL_\text{r} = \int_{\Omega^+}\bL_\text{i}(\bx, \omega_\text{i})f(\bx, \bomega_\text{i}, \bomega_\text{o})d\bomega_\text{i}$ is the reflectance term, and
$f(\bx, \bomega_\text{i}, \bomega_\text{o})$ is the BRDF. The recursive nature of the 
rendering equation is approximated with Monte–Carlo path tracing \cite{kajiya1986rendering, lafortune1996mathematical}.
However, modeling multiple bounces is still computationally expensive and unstable in the inverse rendering process.
To enhance the efficiency and robustness of the optimization, we follow FIPT \cite{wu2023factorized} and adopt factorized light transport \cite{krivanek2022practical, seyb2020design, wang2021learning}, which precomputes shading maps $\bL_\text{d}$, $\bL_\text{s}^0$, $\bL_\text{s}^1$ and linearly interpolates with roughness to obtain final rendering.
Detailed notation and formulation are available in the supplemental material.

\section{Inverse Rendering from LDR Images}
Given multi-view posed LDR images $\{\mathbf{I}_i\}$ captured with the same or varying exposure levels $\{\Delta t_i\}$, our goal is to estimate spatially-varying HDR lighting, surface material (as albedo, roughness, and metallic), and the camera response function (CRF) shared by all input images.
We assume that the primary light sources are observed in the input images, following the recent literature \cite{wu2023factorized, ZhuHYLLXWTHBW2023, yu2023milo}.

Previous approaches perform differentiable path tracing during the material and lighting optimization \cite{azinovic2019inverse, nimierdavid2021material}, but the number of samples is limited due to heavy computation. 
This introduces high variance and unstable estimation. 
The factorized formulation \cite{wu2023factorized, krivanek2022practical, seyb2020design, wang2021learning} splits light transport into BRDF parameters and shading, and enables updating them alternately.
While the method is shown to be more stable and produce high-quality decompositions, it requires HDR images as input to aggregate and bake shading maps, detect emitters, and estimate HDR emissions.

To address this ill-posed optimization problem, we propose a multi-stage optimization strategy alternatingly estimating spatially varying lighting, material, and CRF.
Specifically, we 
(1) initialize the BRDF with an off-the-shelf
monocular inverse rendering
method (\cref{sec:brdf_init}), 
(2) recover HDR lighting from LDR with physically-based rendering (\cref{sec:hdr_restoration}),
(3) bake the diffuse and specular shading (\cref{sec:shading}), and
(4) jointly estimate BRDF and CRF (\cref{sec:train_brdf}). 
We repeat Steps 2--4 in that order until all estimates are converged. 
\cref{fig:framework} illustrates our framework. %

\topic{Geometry reconstruction.}
From input images with camera poses, we reconstruct the geometry with BakedSDF \cite{yariv2023bakedsdf}, and use normal estimation from off-the-shelf method \cite{eftekhar2021omnidata} for regulating surface geometry, following MonoSDF \cite{yu2022monosdf}.

\subsection{BRDF Initialization}\label{sec:brdf_init}

For each input image $\bI_i$, we predict a 2D albedo map $\bA_i$ using an off-the-shelf single-image albedo estimation method \cite{ZhuLMPC2022} and averaging per-pixel albedo values within each semantic part.
Given $\{ \bA_i \}$, we initialize albedo $\ba(\bx)$ by minimizing the error between the projected albedo and the estimated albedo across all input images: $\min_{\ba} \sum_i \| \pi_i(\ba) - \bA_i \|_2$.
Here, $\pi_i$ represents the perspective projection onto the image space of $\bI_i$ by intersecting camera rays with the surface mesh of the input scene and reading $\ba$ at their first intersections.
We obtain semantic category labels from part IDs in synthetic scenes or using Mask2Former \cite{cheng2021mask2former} for real scenes.
Roughness and metallicity are both initialized to 0.5. %

\begin{figure}[t]
    \centering
    \includegraphics[width=0.8\linewidth]{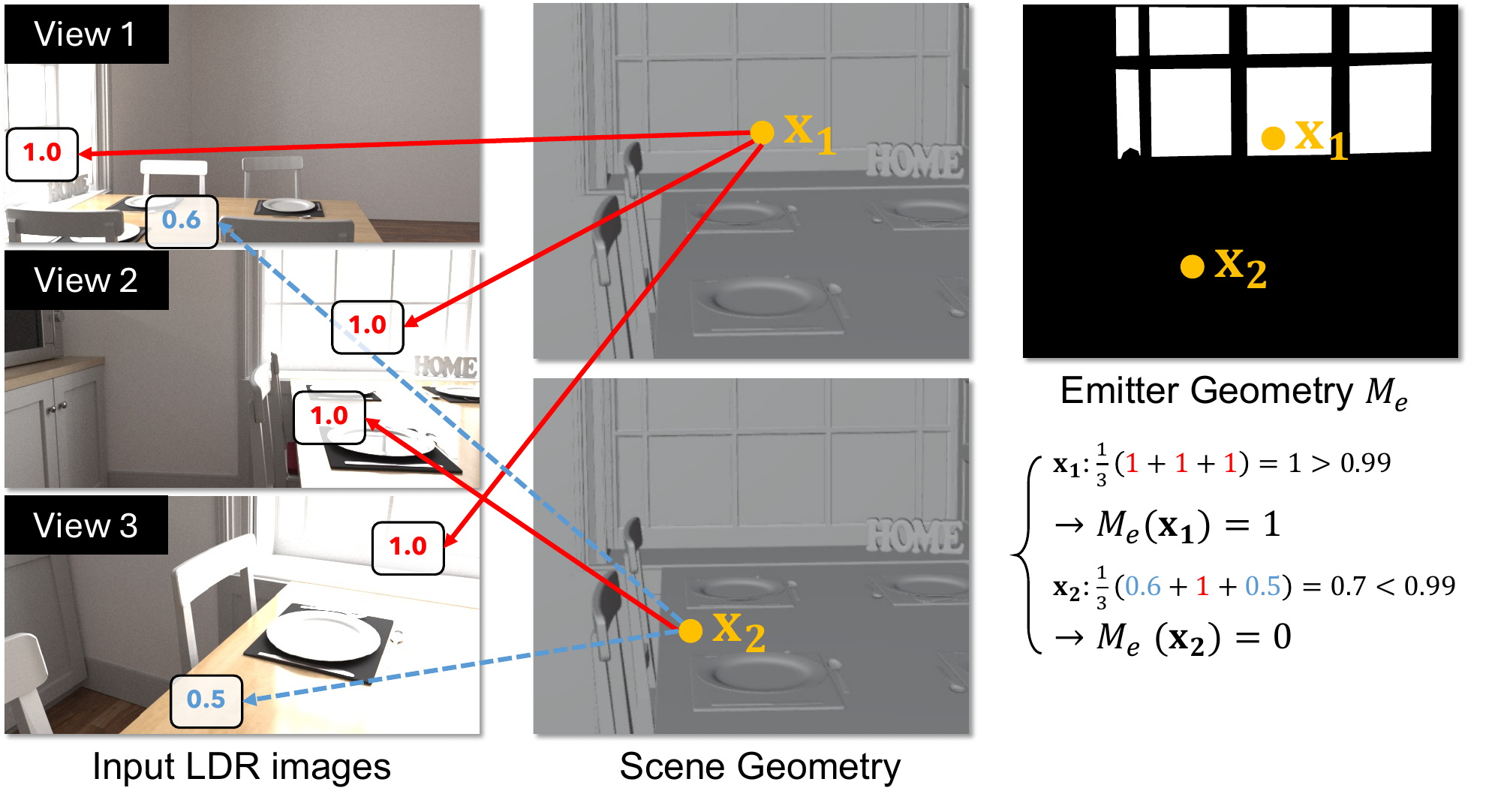}
    \vspace{-4mm}
    \caption{\textbf{Emitter geometry estimation $\boldsymbol M_\text{e}(\bx)$.}
        The point $\bx_1$ on the window is saturated across all input views and thus identified as an emitter.
        The point $\bx_2$ on the table is reflective and saturated in some views (e.g., view 2) but not in others and is NOT an emitter.
    }
    \label{fig:emitter_mask}
    \vspace{-5mm}
\end{figure}

\subsection{HDR Emission Restoration}\label{sec:hdr_restoration}

Given that emitters appear bright across views and produce saturated pixels in LDR images, we estimate the binary emitter geometry $M_\text{e}(\bx)$ on the surface mesh as follows:
\begin{equation}\label{eq:emitter_mask}
    M_\text{e}(\bx) = \begin{cases}
        1 & \quad \text{if } \frac{1}{N}\sum_i v_i(\bx)\bI_i(\pi_i(\bx)) \geq 0.99 \\
        0 & \quad \text{otherwise,}
    \end{cases}
\end{equation}
where $v(\cdot) \in \{0, 1\}$ indicates visibility.
See \cref{fig:emitter_mask}.

\begin{figure}[t]
    \centering
    \includegraphics[width=0.72\linewidth]{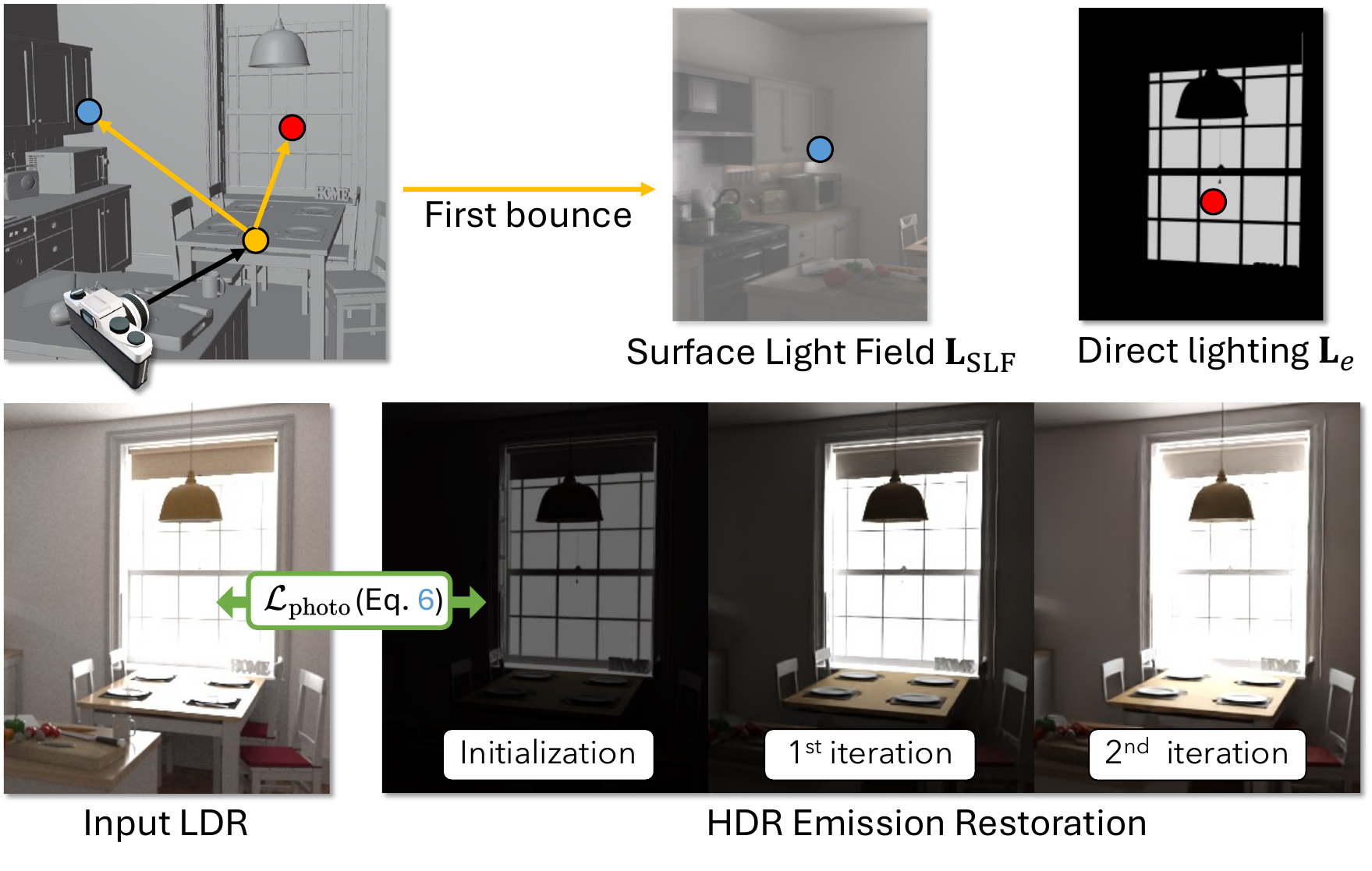}
    \vspace{-4mm}
    \caption{\textbf{HDR emission restoration.}
        \textbf{Top:} Ray sampling process.
        Learnable direct lighting $\bL_\text{e}$ is retrieved if a ray hits an emitter (e.g., window) or $\bL_{\text{SLF}}$ is retrieved otherwise (e.g., wall).
        \textbf{Bottom:} HDR restoration process.
        By performing differentiable physically-based rendering, the photometric loss enhances the emitter intensity $\bL_\text{e}$.
    }
    \label{fig:emitter_restoration}
    \vspace{-3mm}
\end{figure}

The HDR intensity is crucial for inverse rendering but is missing in the LDR images.
To recover the HDR lighting, we perform physically-based rendering with single-bounce path tracing to generate the images \eqref{eq:rendering}.
To estimate the radiance at the point $\bx$ observed from a camera ray, we trace $N \!=\! 128$ secondary rays from $\bx$, sampled with the current BRDF estimates. 
The resulting radiance is approximated by:
\begin{align}
    \bL_\text{r}(\bx, \bomega_\text{o}) &= \displaystyle\sum_{i=1}^{N}\bL_{\text{end}}(\bx')f(\bx, \bomega_\text{i}, \bomega_\text{o}) \text{,} \\
\label{eq:L_end}
    \bL_{\text{end}}(\bx) &= \begin{cases}
        \bL_{\text{e}}(\bx) & \quad \text{if }  M_\text{e}(\bx) = 1\\
        \bL_{\text{SLF}}(\bx) & \quad \text{otherwise,}
    \end{cases}
\end{align}
where $\bx' = \bx + d \cdot \bomega_\text{i}$ is the intersected point of the secondary ray along direction $\bomega_\text{i}$.
Intuitively, we retrieve the radiance from the \emph{learnable} emission radiance $\bL_\text{e}(\bx)$ if the secondary ray hits an emitter directly, or otherwise from the pre-computed surface light field $\bL_{\text{SLF}}(\bx)$ to approximate the global illumination.
\cref{fig:emitter_restoration} shows an illustration. 

To recover the HDR emitter radiance, we minimize the photometric loss with gradient descent: $\min_{\bL_\text{e}} \mathcal{L}_{\text{photo}}$, where
\begin{equation}
    \mathcal{L}_{\text{photo}} =  \sum_i \left\| \texttt{CRF}(\min(\bL_\text{o}(\bx, \omega_\text{o}) \Delta t_i, 1)) - \bI_i \right\|_2 \, \text{.} \label{eq:l_photo}
\end{equation}
The emitter radiance $\bL_\text{e}(\bx)$ is enhanced to match the captured images, while the BRDF and CRF are fixed in this stage.
Intuitively, in order to minimize the photometric loss with physically-based rendering, the emitter intensity $\bL_{\text{e}}(\bx)$ should be increased significantly and enhance the overall brightness of the rendered image.
Therefore, the HDR lighting is recovered in this process.

\subsection{Shading Baking}
\label{sec:shading}

After the HDR emission is recovered in the previous stage, we bake the diffuse and specular shading maps $\bL_\text{d}$, $\bL_\text{s}^0$, $\bL_\text{s}^1$, following factorized light transport (details in supplementary material). 
Given a surface point intersected by a camera ray, multiple rays are sampled according to the current (fixed) BRDF.
The incident radiance is computed with multi-bounce path tracing, where a ray keeps tracing if it hits a non-emissive specular surface and stops otherwise. 
The radiance of the endpoint follows \cref{eq:L_end}.

\subsection{BRDF \& CRF Optimization}\label{sec:train_brdf}
With the baked shading maps $\bL_\text{d}$, $\bL_\text{s}^0$, and $\bL_\text{s}^1$, the albedo, metallicity, roughness, and CRF are jointly optimized. 
The optimization involves the following objective function:
\begin{equation}
    \min_{\ba, m, \sigma, \bg} \mathcal{L}_{\text{photo}} + \lambda_{\text{a}}\mathcal{L}_{\text{albedo}} + \lambda_{\text{c}}\mathcal{L}_{\text{CRF}} + \lambda_{\text{m}}\mathcal{L}_{\text{mat}} \, \text{,}
\end{equation}
where $\lambda_\text{a}=0.01$, $\lambda_\text{c}=0.001$, and $\lambda_{\text{m}}=0.005$.

\topic{Photometric loss.}
We perform physically-based rendering, and the photometric loss $\mathcal{L}_{\text{photo}}$ is computed as in Eq. \ref{eq:l_photo}.

\topic{Albedo regularization.}
Motivated by the shift-scale-inva\-riant loss in MonoSDF \cite{yu2022monosdf}, we adopt a shift-invariant loss against the monocular albedo maps $\{\bA_i\}$:
\begin{equation}\label{eq:loss_albedo}
    \mathcal{L}_{\text{albedo}} = \sum_i \| \pi_i (\ba) - s_i \bA_i \|_2 \, \text{,}
\end{equation}
where $s_i$ is the scale used to align the two albedo maps.

\topic{CRF regularization.}
Based on the fair assumption that the learned CRF should not deviate significantly from the mean curve $\bar{\bg}$ (Eq. \ref{eq:crf_parameter}), and that the curve should be monotonically increasing, we regularize the $L_2$ norm of the estimated coefficients of the PCA bases, and enforce the monotonicity of the sampled CRF as follows:
\begin{equation}\label{eq:loss_crf}
    \mathcal{L}_{\text{CRF}} = \|\bw\|_2 + \displaystyle\sum_i\max(g_{i-1} - g_i, 0) \, \text{.}
\end{equation}

\topic{Material regularization.}
We apply roughness-metallicity regularization $\mathcal{L}_{\text{mat}}$, which enforces the consistency of surface roughness and metallic within the same semantic instances (e.g. chairs, tables).
While the shading and emission are initially inaccurate and lead to erroneous BRDFs, we alternately update the BRDFs and produce better emission and shading, improving the BRDFs. Our alternating strategy yields more stable optimization and better estimation.

\section{Experiments}

We evaluate our method on both synthetic and real-world data and compare it against several baseline methods. 
We analyze the quality of our
inverse rendering,
relighting, and novel-view synthesis capability. 
Please see our supplementary material for additional results and evaluations.
We will release our code and data.

\topic{Datasets.}
We evaluate our method and baselines on the ScanNet++ \cite{yeshwanthliu2023scannetpp} and FIPT \cite{wu2023factorized} datasets.
ScanNet++ contains 380 real-world indoor scenes and collects multi-view LDR images with a DSLR camera at a constant exposure level.
We select four scenes that cover a variety of indoor scenes (e.g., office, bathroom).
The FIPT dataset consists of four synthetic and two real-world scenes.
The synthetic scenes provide ground-truth geometry and material properties for evaluation.
We used the LDR images provided in the dataset, where the lighting is clipped in overly-exposed regions, compressed with a nonlinear CRF, quantized into 256 levels, and stored in 8-bit PNG format.
We adopt the constant exposure setting to provide a fair comparison with baselines, as they cannot handle varying exposure.
We further demonstrate that IRIS handles varying exposure levels of input LDR images and estimates CRF jointly.

\topic{Baselines.}
We compare our method with the following baselines.
\emph{\citet{li2022physically}} estimate BRDF, geometry (depth \& normal), and lighting condition of an indoor scene from a single image using a data-driven approach. 
In addition to the single image, the model also takes emitter masks as input, and we provide the mask estimated by our method \eqref{eq:emitter_mask}.
\emph{NeILF \cite{yao2022neilf}} targets inverse rendering from known geometry and multi-view images. The spatially-varying BRDF and illumination are parameterized as neural fields \cite{mildenhall2020nerf}. 
Physically-based rendering and single-bounce ray tracing are performed during the training and rendering process. 
The HDR-to-LDR tone mapping is modeled as a learnable gamma correction.
The original FIPT \cite{wu2023factorized} is designed for HDR input, %
and fails to estimate the emission mask from LDR images.
Thus, we additionally provide the emission mask estimated by our method \eqref{eq:emitter_mask}. 
We denote this modified version as ``\emph{FIPT*}''.

To further evaluate our performance, we compare the original FIPT with HDR input if the HDR information is available. This privileged information will make illumination estimation much more accurate.

\subsection{Inverse Rendering of Real Scenes}

\begin{figure*}[t]
    \centering\setlength{\tabcolsep}{0.1em}
    \resizebox{1.0\textwidth}{!}{%
    \begin{tabular}{@{}clcccc@{}}
    
    Input LDR Image & & Reconstruction & Diffuse reflectance $\bk_\text{d}$ & Roughness $\sigma$ & Emission \\[0.2em]

    \frame{\includegraphics[width=0.2\textwidth]{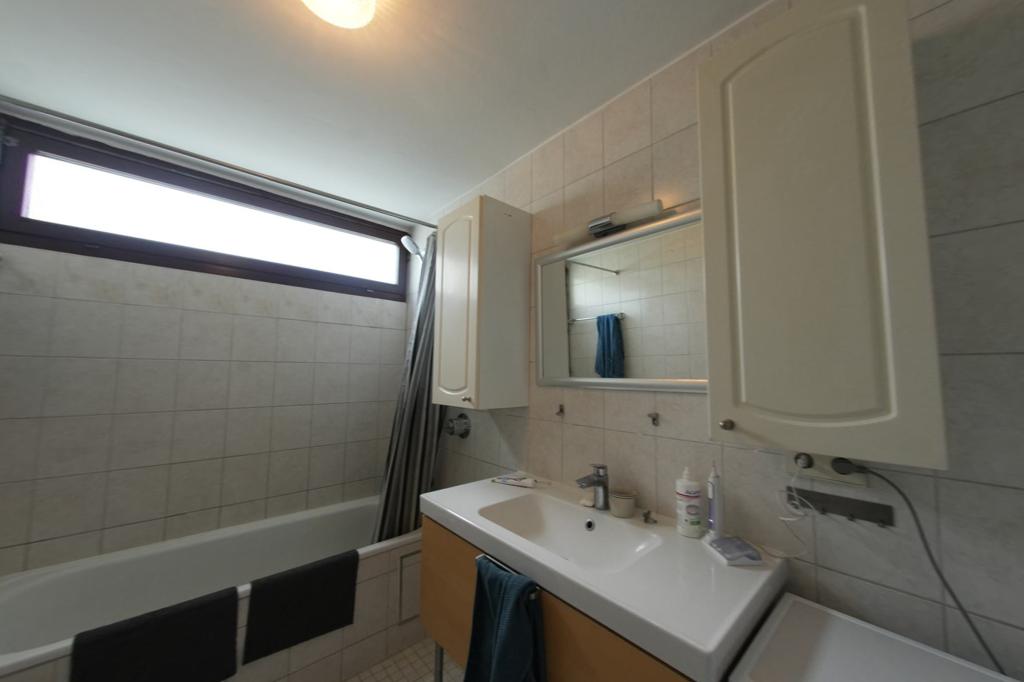}}& \raisebox{2.5\normalbaselineskip}[0pt][0pt]{\rotatebox[origin=c]{90}{NeILF~\cite{yao2022neilf}}} & \frame{\includegraphics[width=0.2\textwidth]{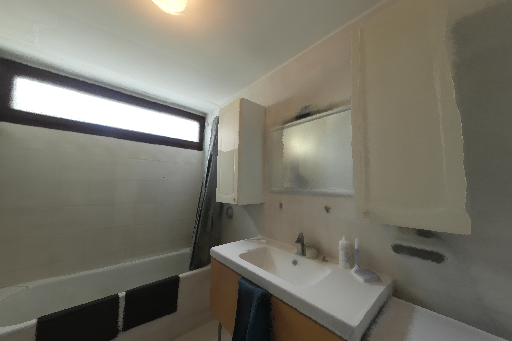}} & \frame{\includegraphics[width=0.2\textwidth]{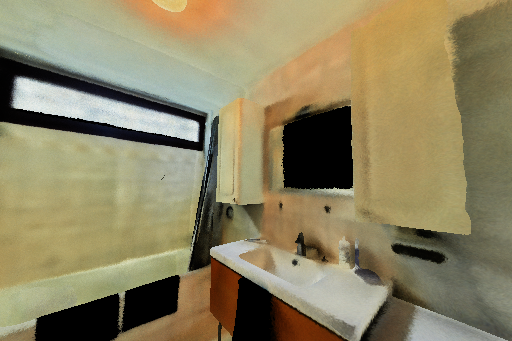}} & \frame{\includegraphics[width=0.2\textwidth]{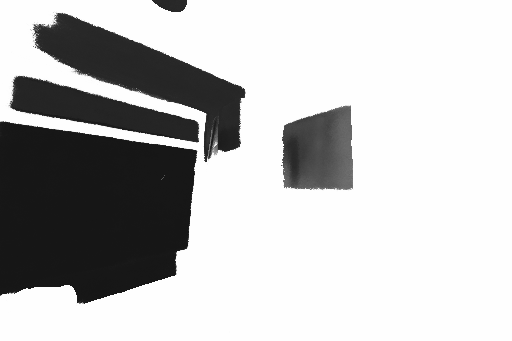}} & \frame{\includegraphics[width=0.2\textwidth]{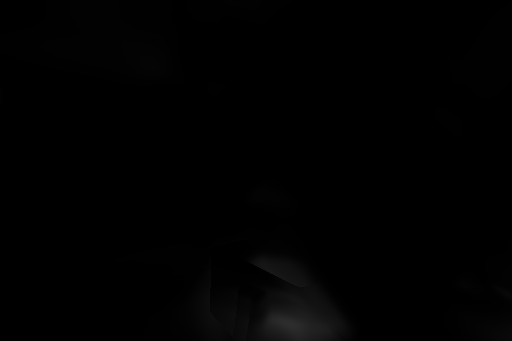}} \\
     & \raisebox{2.5\normalbaselineskip}[0pt][0pt]{\rotatebox[origin=c]{90}{FIPT*~\cite{wu2023factorized}}} & \frame{\includegraphics[width=0.2\textwidth]{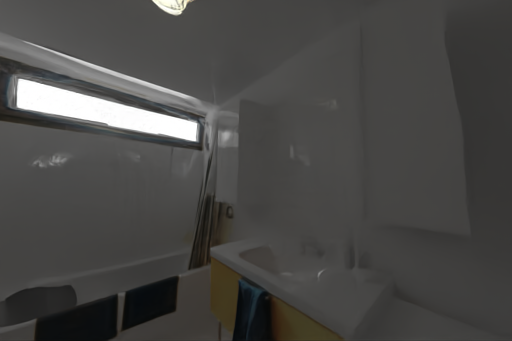}} & \frame{\includegraphics[width=0.2\textwidth]{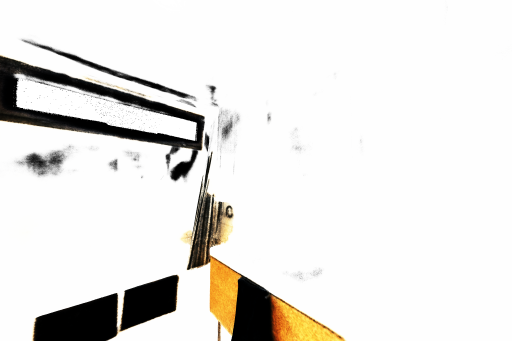}} & \frame{\includegraphics[width=0.2\textwidth]{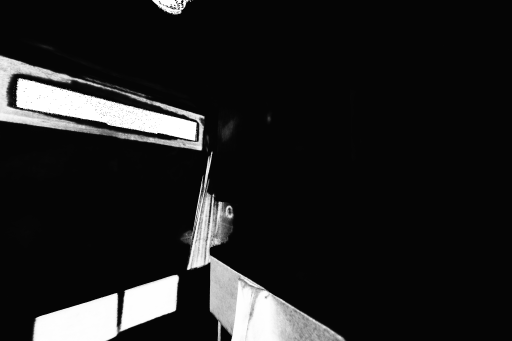}} & \frame{\includegraphics[width=0.2\textwidth]{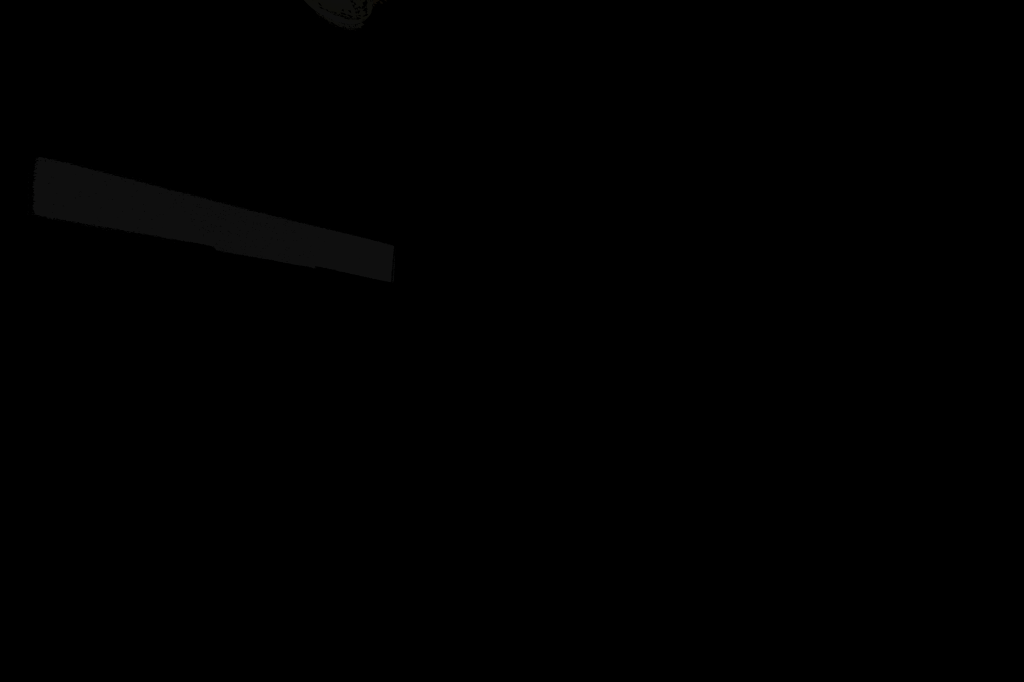}} \\
     & \raisebox{2.5\normalbaselineskip}[0pt][0pt]{\rotatebox[origin=c]{90}{Ours}} & \frame{\includegraphics[width=0.2\textwidth]{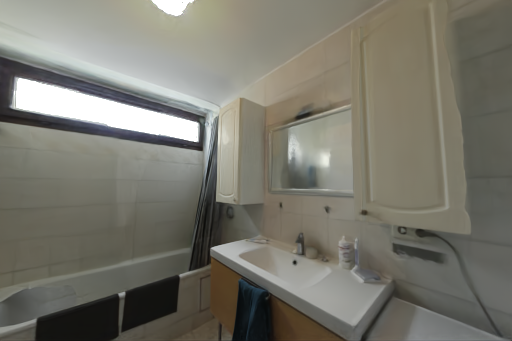}} & \frame{\includegraphics[width=0.2\textwidth]{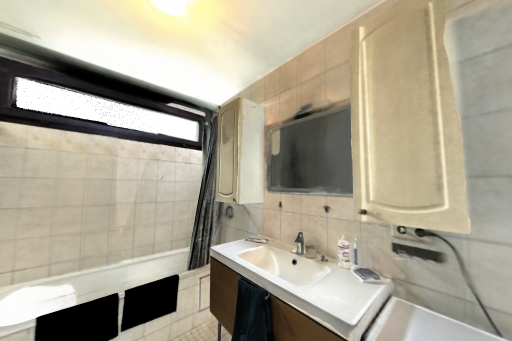}} & \frame{\includegraphics[width=0.2\textwidth]{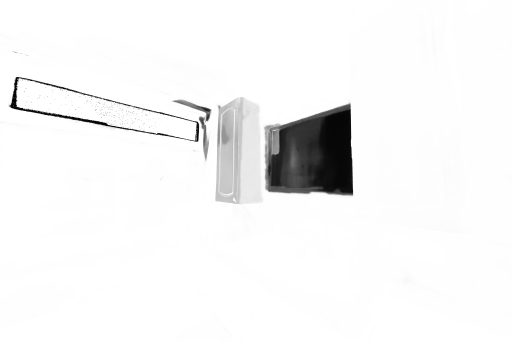}} & \frame{\includegraphics[width=0.2\textwidth]{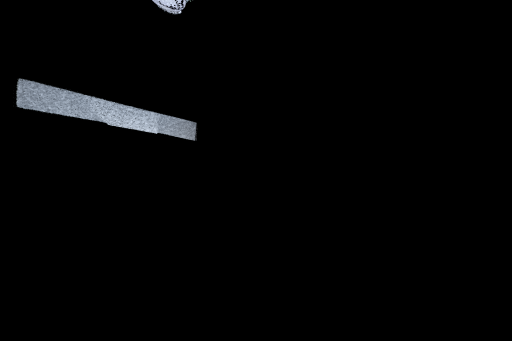}} \\
    
    \end{tabular}%
    }
    \vspace{-3mm}
    \caption{\textbf{Decompositions of real scenes} from ScanNet++ \cite{yeshwanthliu2023scannetpp}. %
    In addition to good reconstruction, IRIS produces a detailed diffuse map, identifies the mirror as a low-roughness region, and recovers HDR lighting from windows and ceiling light. Please note that the emission maps are normalized to visualize the relative intensity.
    }
    \vspace{-3mm}
    \label{fig:real_intrinsic}
\end{figure*}

\begin{figure*}[t]
    \centering\setlength{\tabcolsep}{0.1em}
    \resizebox{1.0\textwidth}{!}{%
    \begin{tabular}{@{}clcccc@{}}
    
    Input LDR Image & & Reconstruction & Relighting 1 & Relighting 2 & Object Insertion \\[0.2em]

    \frame{\includegraphics[width=0.2\textwidth]{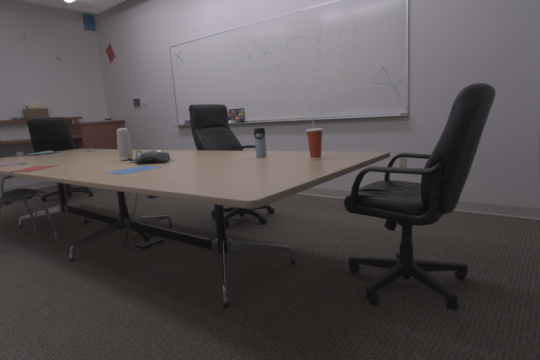}}& \raisebox{2.5\normalbaselineskip}[0pt][0pt]{\rotatebox[origin=c]{90}{FIPT*~\cite{wu2023factorized}}} &\frame{\includegraphics[width=0.2\textwidth]{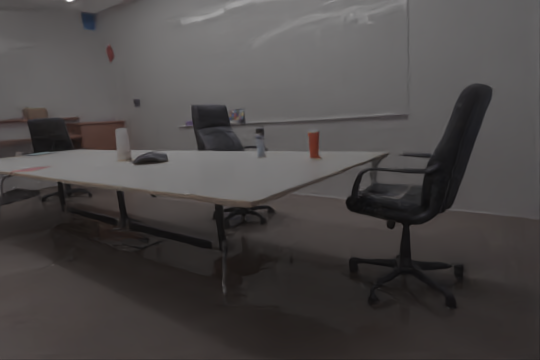}} &\frame{\includegraphics[width=0.2\textwidth]{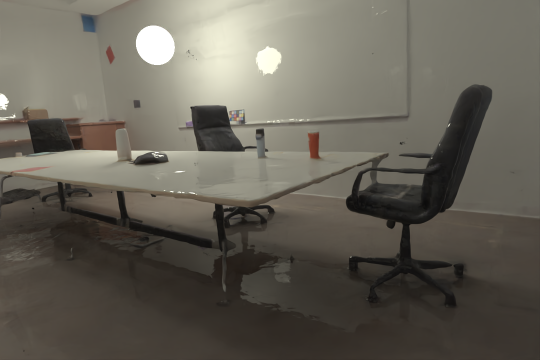}} &\frame{\includegraphics[width=0.2\textwidth]{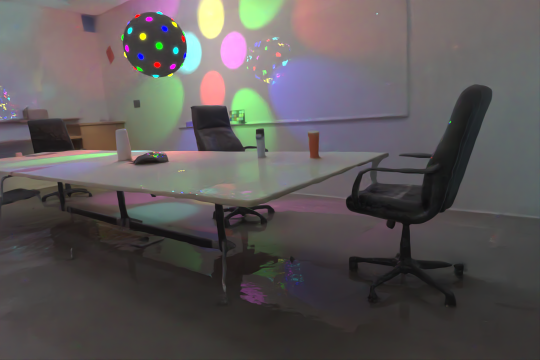}} &\frame{\includegraphics[width=0.27\textwidth]{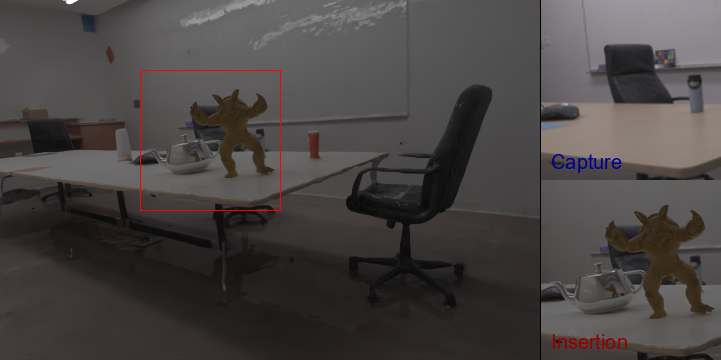}} \\
     & \raisebox{2.5\normalbaselineskip}[0pt][0pt]{\rotatebox[origin=c]{90}{Ours}} &\frame{\includegraphics[width=0.2\textwidth]{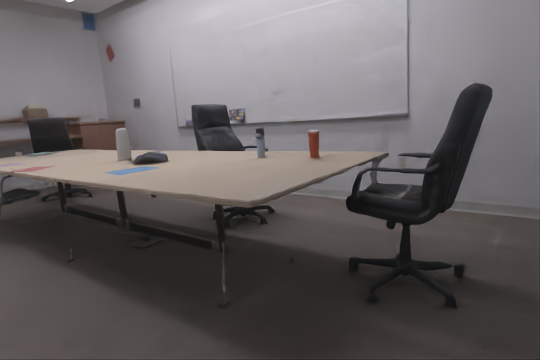}}
     &\frame{\includegraphics[width=0.2\textwidth]{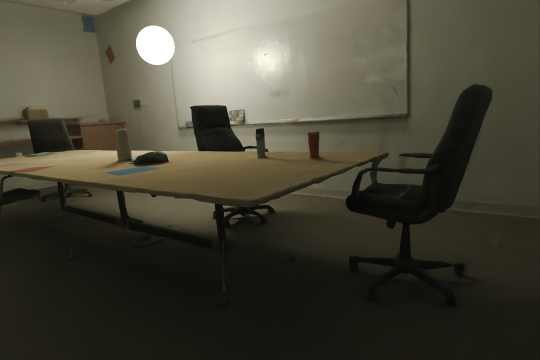}} &\frame{\includegraphics[width=0.2\textwidth]{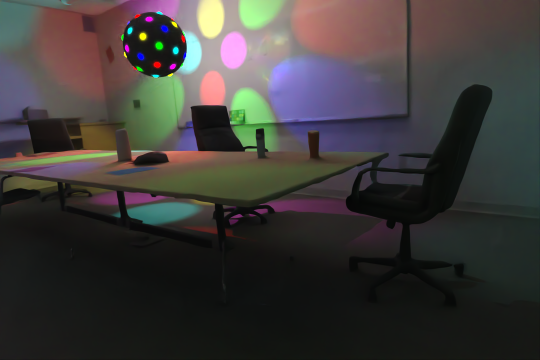}} &\frame{\includegraphics[width=0.27\textwidth]{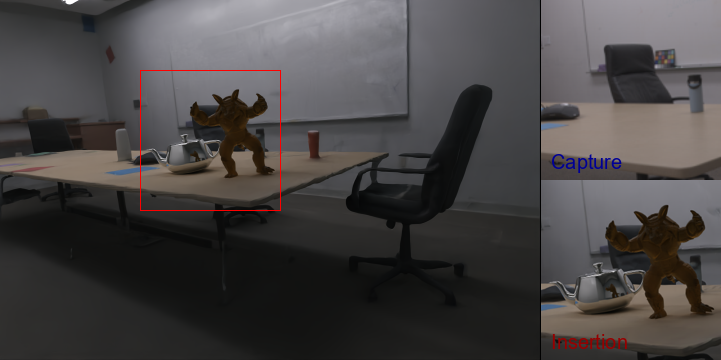}} \\

    \frame{\includegraphics[width=0.2\textwidth]{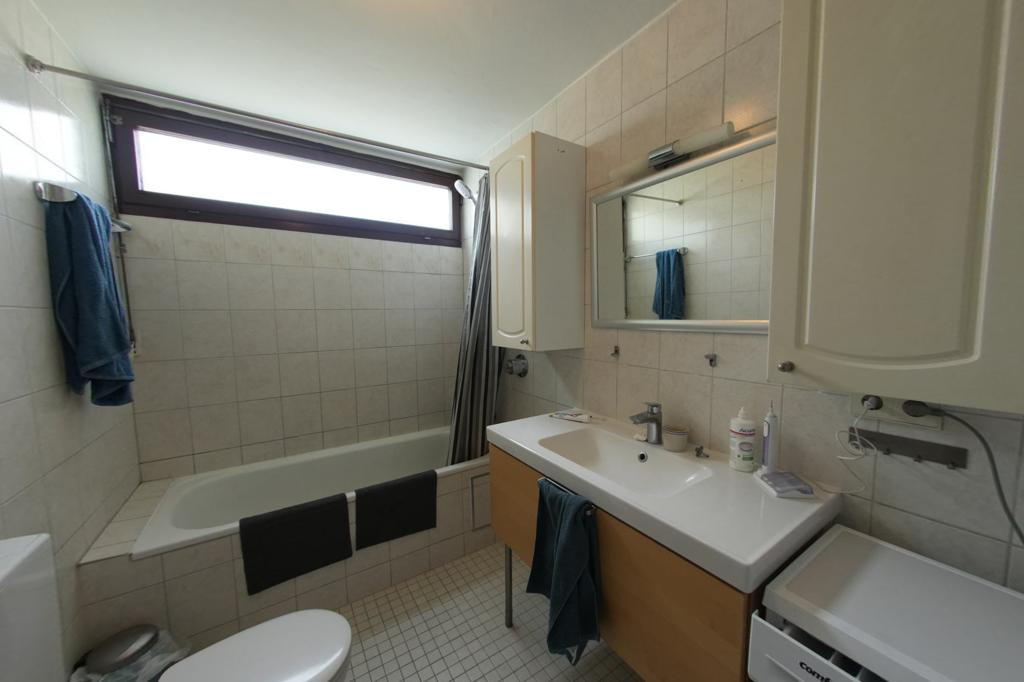}}& \raisebox{2.5\normalbaselineskip}[0pt][0pt]{\rotatebox[origin=c]{90}{FIPT*~\cite{wu2023factorized}}} &\frame{\includegraphics[width=0.2\textwidth]{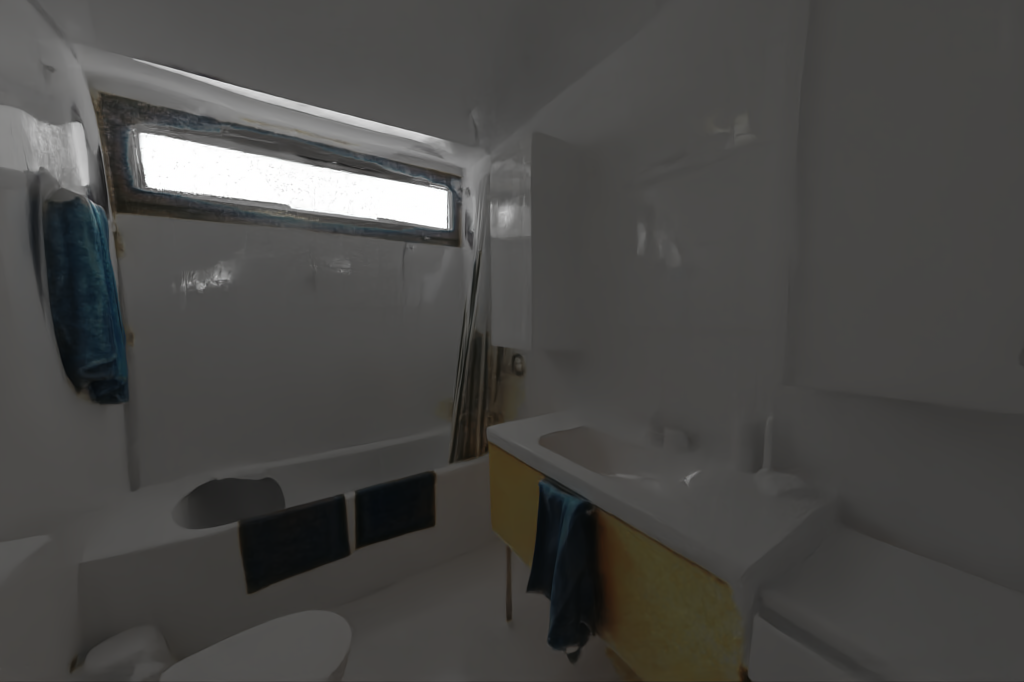}} &\frame{\includegraphics[width=0.2\textwidth]{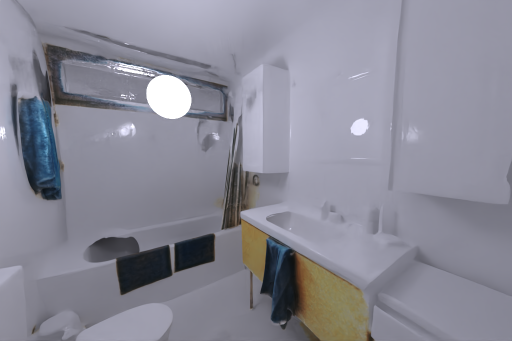}} &\frame{\includegraphics[width=0.2\textwidth]{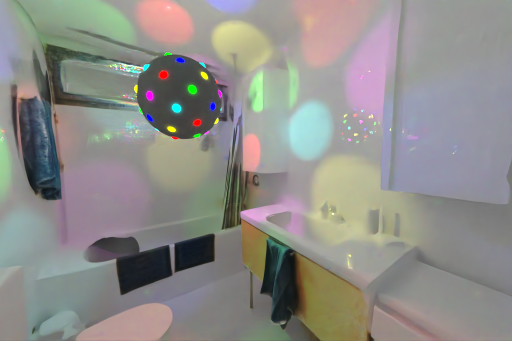}} &\frame{\includegraphics[width=0.27\textwidth]{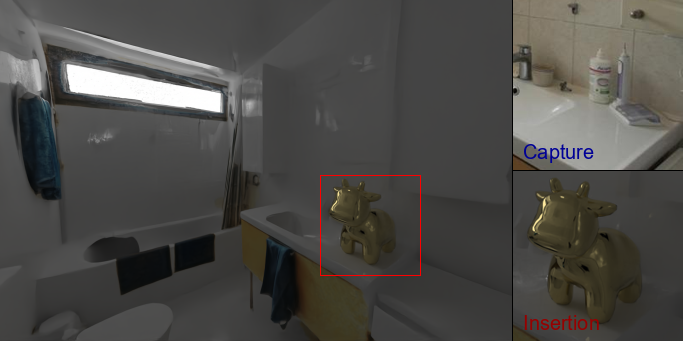}} \\
     & \raisebox{2.5\normalbaselineskip}[0pt][0pt]{\rotatebox[origin=c]{90}{Ours}} &\frame{\includegraphics[width=0.2\textwidth]{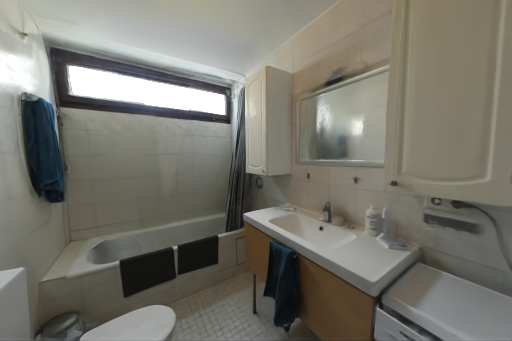}}
     &\frame{\includegraphics[width=0.2\textwidth]{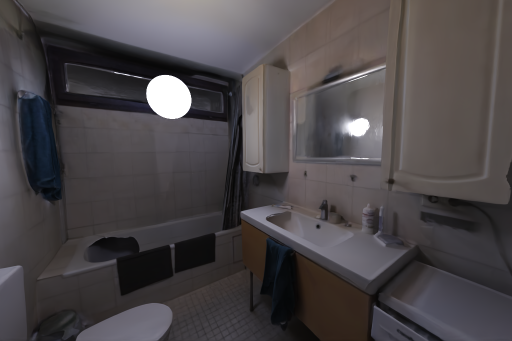}} &\frame{\includegraphics[width=0.2\textwidth]{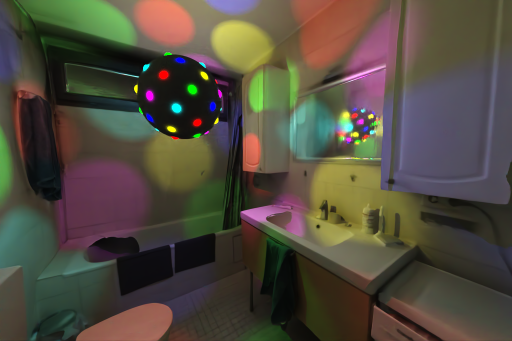}} &\frame{\includegraphics[width=0.27\textwidth]{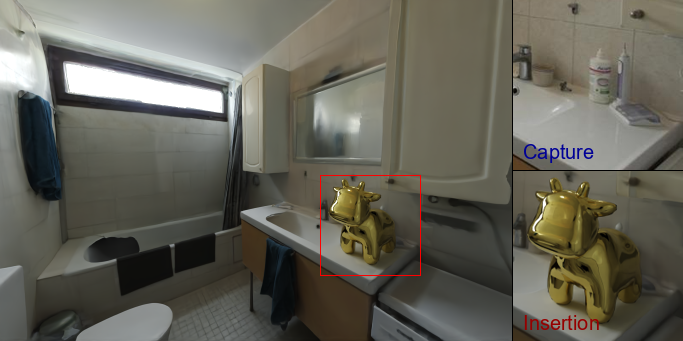}} \\

    \end{tabular}%
    }
    \vspace{-3mm}
    \caption{\textbf{Relighting and object insertion with real scenes} from the ScanNet++ \cite{yeshwanthliu2023scannetpp} and FIPT datasets \cite{wu2023factorized}.
    When inserting new light sources (Relighting 1 \& 2), IRIS not only changes shadows but also produces reflection of the light on the whiteboard (first scene) and mirror (second scene), enhancing realism substantially. The recovered HDR lighting also strengthens the reflection and shadow of inserted objects (``Object Insertion'' column).}
    \vspace{-5mm}
    \label{fig:real_relight}
\end{figure*}

\begin{figure}[t]
    \centering\setlength{\tabcolsep}{0.1em}
    \resizebox{1.0\linewidth}{!}{
    \begin{tabular}{cccc}
       Reconstruction & Relight 1 & Relight 2 & Relight 3\\
        
        \includegraphics[width=0.3\linewidth]{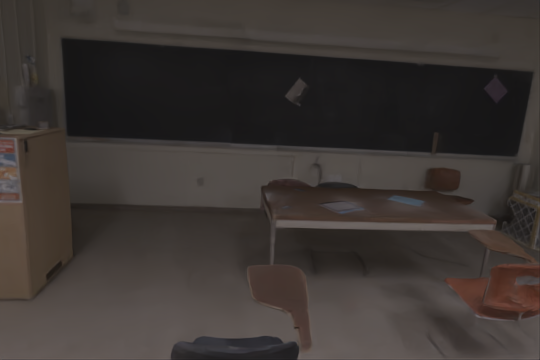} & \includegraphics[width=0.3\linewidth]{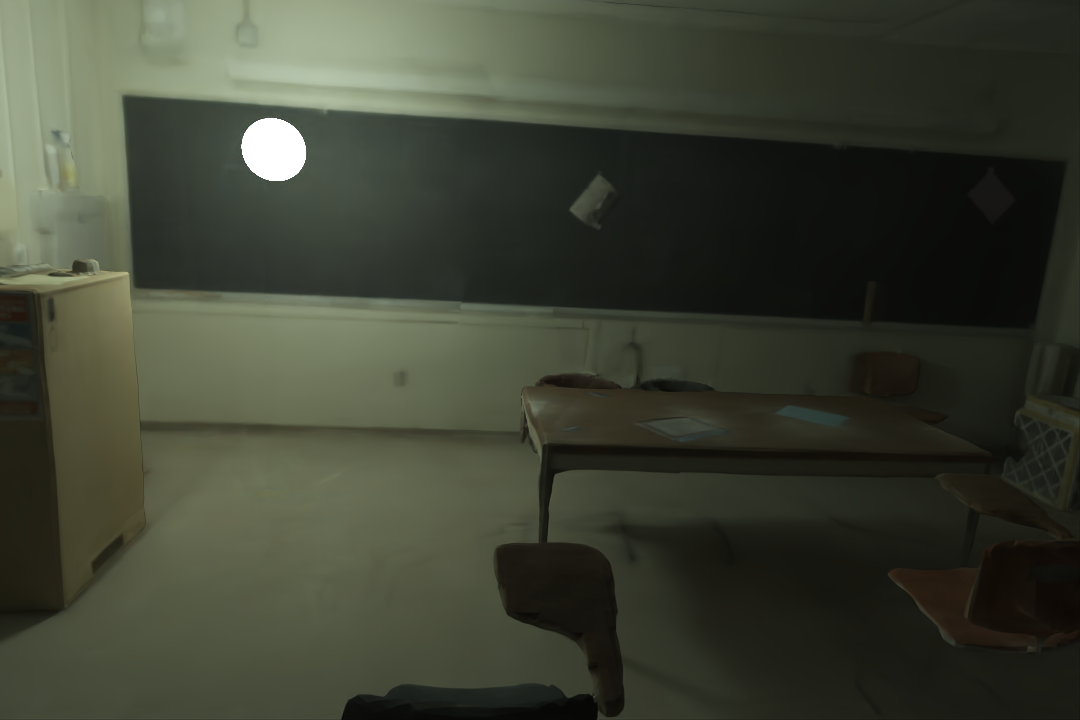} & \includegraphics[width=0.3\linewidth]{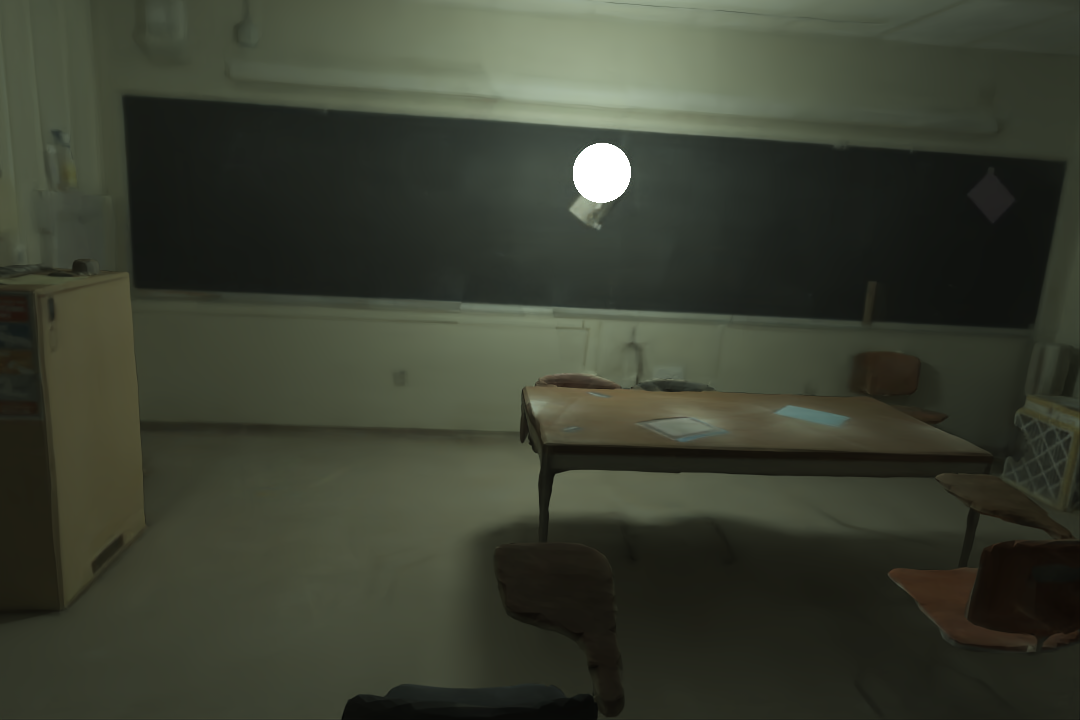} & \includegraphics[width=0.3\linewidth]{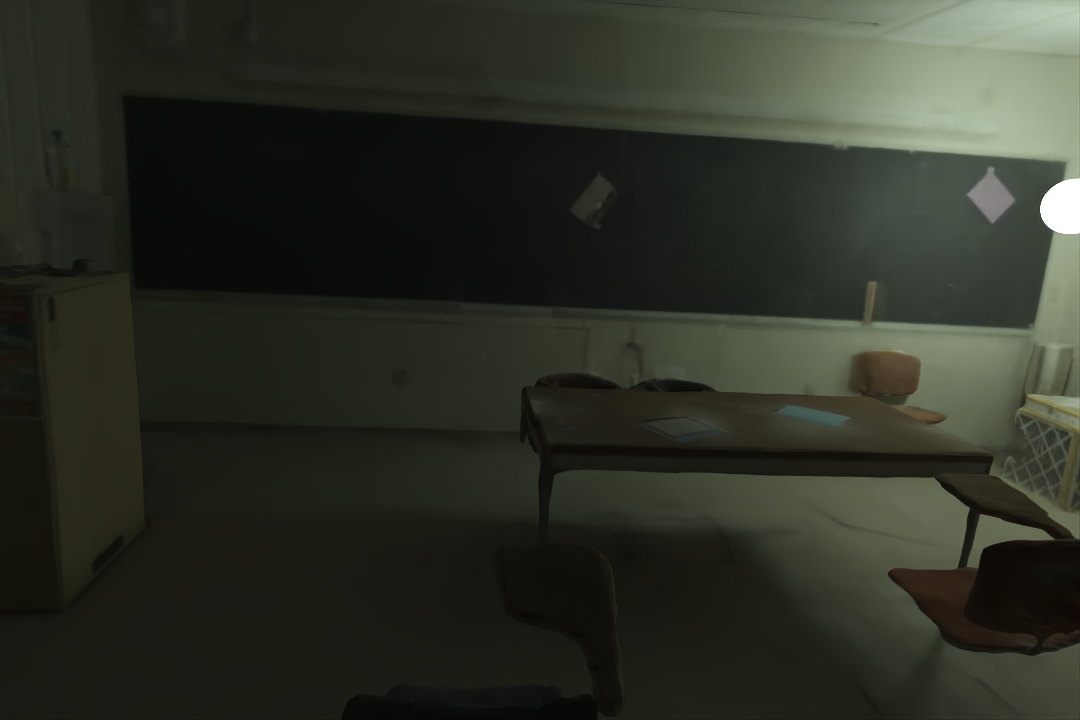} 
    \end{tabular}
    }
    \vspace{-3mm}
    \caption{\textbf{Relighting with a moving light ball} in a real scene.}
    \label{fig:move}
    \vspace{-3mm}
\end{figure}

We provide a qualitative evaluation of inverse rendering
in \cref{fig:real_intrinsic}.
Conducting single-bounce ray tracing in the neural light field, NeILF \cite{yao2022neilf} fails to remove the shadow in diffuse reflectance $\bk_\text{d}$, estimates lower roughness $\sigma$ on the walls than the mirror, and produces an incorrect emission mask.
FIPT* struggles to recover surface material from LDR input, where accurate lighting information is missing.
Please note that FIPT* takes emitter masks from \cref{eq:emitter_mask} as additional input, but does not recover HDR emission in the optimization procedure.
On the other hand, IRIS recovers an almost shading-free diffuse field.
Regarding roughness, our method effectively identifies specular surfaces with low roughness on the mirror, demonstrating better specular/rough material separation.
Additionally, our proposed HDR emission estimation proves effective for real scenes, recovering HDR lighting sources (i.e., window and ceiling light).
The restored HDR lighting makes the light transport more accurate and significantly enhances the performance of inverse rendering.

\subsection{View Synthesis and Relighting of Real Scenes}
We compare the relighting with FIPT* in \cref{fig:real_relight}.
FIPT* fails to estimate accurate surface material from LDR input and tends to recover low roughness for most surfaces, which unrealistically reflects the inserted light source and objects.
To be more specific, the ground of the office (first scene) and the wall of the bathroom (second scene) look overly smooth and specular, which is not consistent with the observation and produces unrealistic reflections.
In contrast, IRIS produces better relighting and object insertion quality thanks to better inverse rendering.
The reflections on the specular surfaces (e.g., whiteboard, mirror) and the shading on the wall demonstrate the strength of IRIS for realistic relighting. Furthermore, the recovered HDR lighting can produce naturalistic specular reflection and shadows for inserted objects. Additionally, we insert moving light sources in the scene, shown in \cref{fig:move}, where the shadow on the ground and shading on the table change consistently. IRIS enables the photorealistic rendering of novel objects under diverse lighting and from arbitrary viewpoints, showcasing the significant potential for content creation applications.

\begin{table}[t]
    \caption{\label{tab:syn_intrinsic}%
        \textbf{BRDF-emission comparison on synthetic data.}
        Numbers are averaged across four synthetic scenes with GT \cite{wu2023factorized}.
        The best metrics among LDR methods are highlighted in bold.
    }
    \centering\setlength{\tabcolsep}{4pt}
    \resizebox{0.8\linewidth}{!}{%
    \begin{tabular}{lcccccc}
    \toprule
        & & $\bk_\text{d}$ & $\ba'$ & $\sigma$ & \multicolumn{2}{c}{$\bL_\text{e}$} \\
        Method & Input & \multicolumn{3}{c}{PSNR $\uparrow$} & IoU $\uparrow$ & L2 $\downarrow$\\
    \midrule
        Li et al~\cite{li2022physically} & LDR & 16.73 & 14.01 & 11.30 & 0.35 & 2.29 \\
       NeILF~\cite{yao2022neilf}& LDR & 16.85 & 14.02 & 16.96 & --- & --- \\
       FIPT* & LDR & 15.49 & \padd 9.74 & \padd 4.99 &  \textbf{0.69} & 0.28\\
       Ours& LDR  & \textbf{22.33} & \textbf{17.92} & \textbf{21.38} & \textbf{0.69} & \textbf{0.12} \\
       \midrule %
       FIPT~\cite{wu2023factorized}& HDR & 29.95 & 25.98 & 26.37 & 0.86 & 0.03 \\
    \bottomrule
    \end{tabular}%
    }
    \vspace{-5mm}
\end{table}

\subsection{Quantitative Evaluation}
Quantitative evaluation of inverse rendering is difficult for real-world scenes as ground truth is not available.
As a result, we evaluate IRIS and baselines on the synthetic scenes from FIPT, where ground truth of surface material, illumination, and relighting are provided. Please note that we use the ground-truth mesh during optimization for all methods to minimize the influence of imperfect geometry and conduct a fair comparison.
We compute the metrics of inverse rendering
and provide the comparison in \cref{tab:syn_intrinsic}, which aggregates the results from the 4 synthetic FIPT scenes \cite{wu2023factorized}.

We compare the roughness $\sigma$, diffuse reflectance $\bk_\text{d}$ among the diffuse surfaces, and material reflectance: $\ba'$.
We report the PSNR for these BRDF estimates.
For emission, we calculate the intersection over the union (IoU) of the emission masks and compare the $L_2$ error of estimated emission maps. 
All the estimates are rendered in image space and compared with ground truth. 

\Cref{tab:syn_intrinsic} shows that our method surpasses all baseline methods that use LDR inputs regarding 
inverse rendering
quality. 
Compared to the state-of-the-art single-view method \cite{li2022physically}, our method excels in all metrics across various scenes, indicating the superior decomposition of material properties from multi-view observations. 
Furthermore, even when the estimated emitter mask is provided for FIPT*, it fails to recover emission radiance, leading to compromised BRDF estimation accuracy.
The $L_2$ error of emission maps shows that our proposed HDR restoration (\cref{sec:hdr_restoration}) successfully recovers the HDR lighting from LDR observations, which leads to better light transport modeling and more accurate BRDF estimation. 
We also compare with FIPT taking HDR images as input (bottom row), which serves as a reference, indicating complete lighting information is crucial for accurate inverse rendering. IRIS recovers HDR and achieves the smallest metric gaps among all methods taking LDR input.

We compare novel-view synthesis and relighting of all competing algorithms and report results in \cref{tab:nvs_synthetic}.
The novel-view synthesis results show that IRIS is comparable with baseline methods \cite{yao2022neilf}, while exhibiting superior material and lighting estimation.
Our main focus is relighting, and IRIS outperforms all baselines with LDR input, demonstrating the effectiveness of inverse rendering.

\begin{figure}
    \centering\setlength{\tabcolsep}{3pt}
    \newcommand{\cropimg}[1]{\includegraphics[width=0.2\linewidth,trim=0 10 50 80,clip]{figures/images/crf_varying/#1}}
    \resizebox{1.0\linewidth}{!}{%
    \begin{tabular}{cccc}

    \footnotesize{CRF 1} & \footnotesize{CRF 2} & \footnotesize{CRF 3} & \footnotesize{CRF 4}       
    \\

    \cropimg{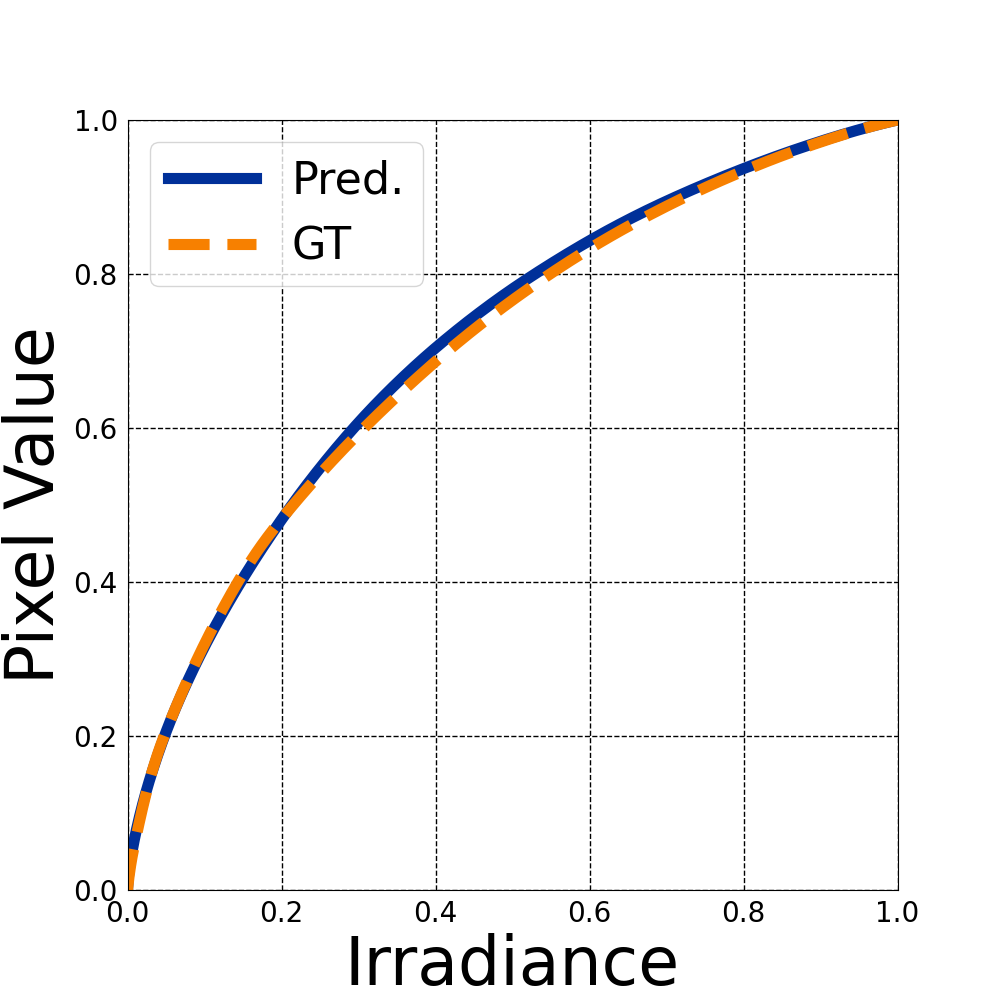} & \cropimg{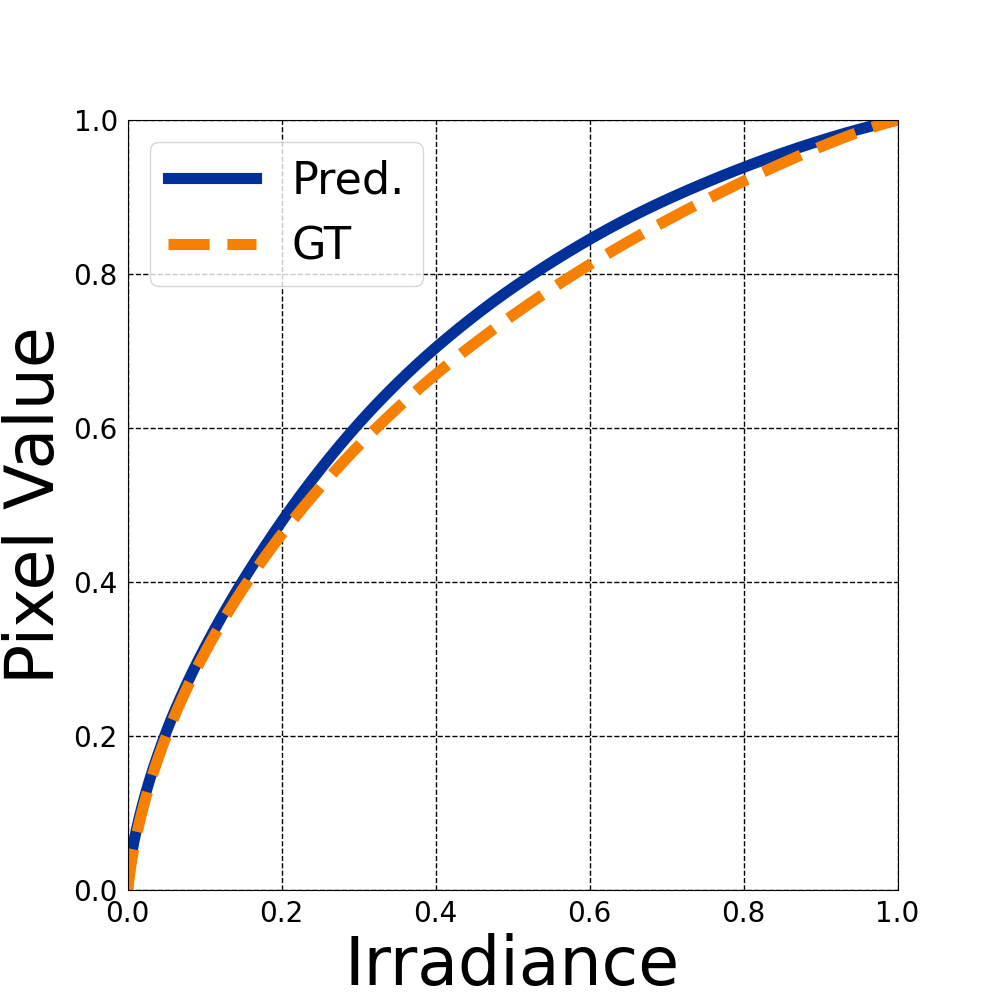} & \cropimg{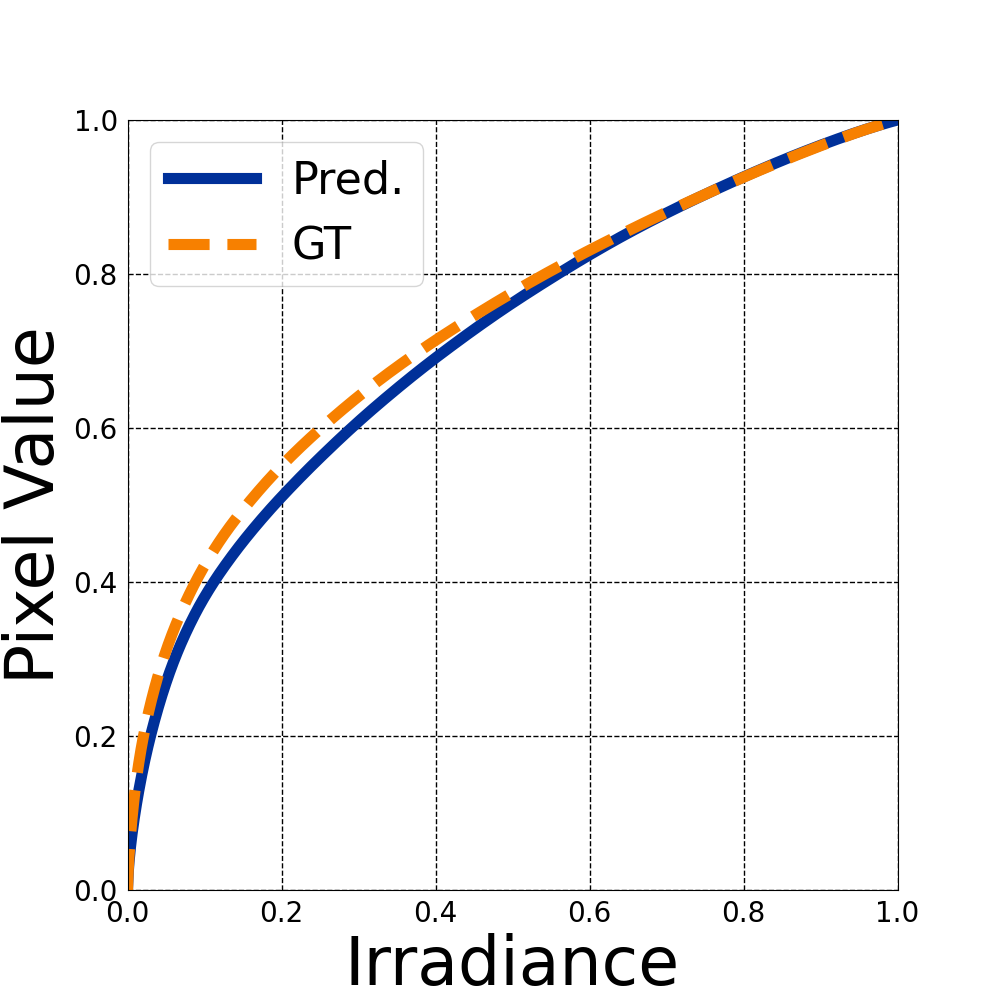} & \cropimg{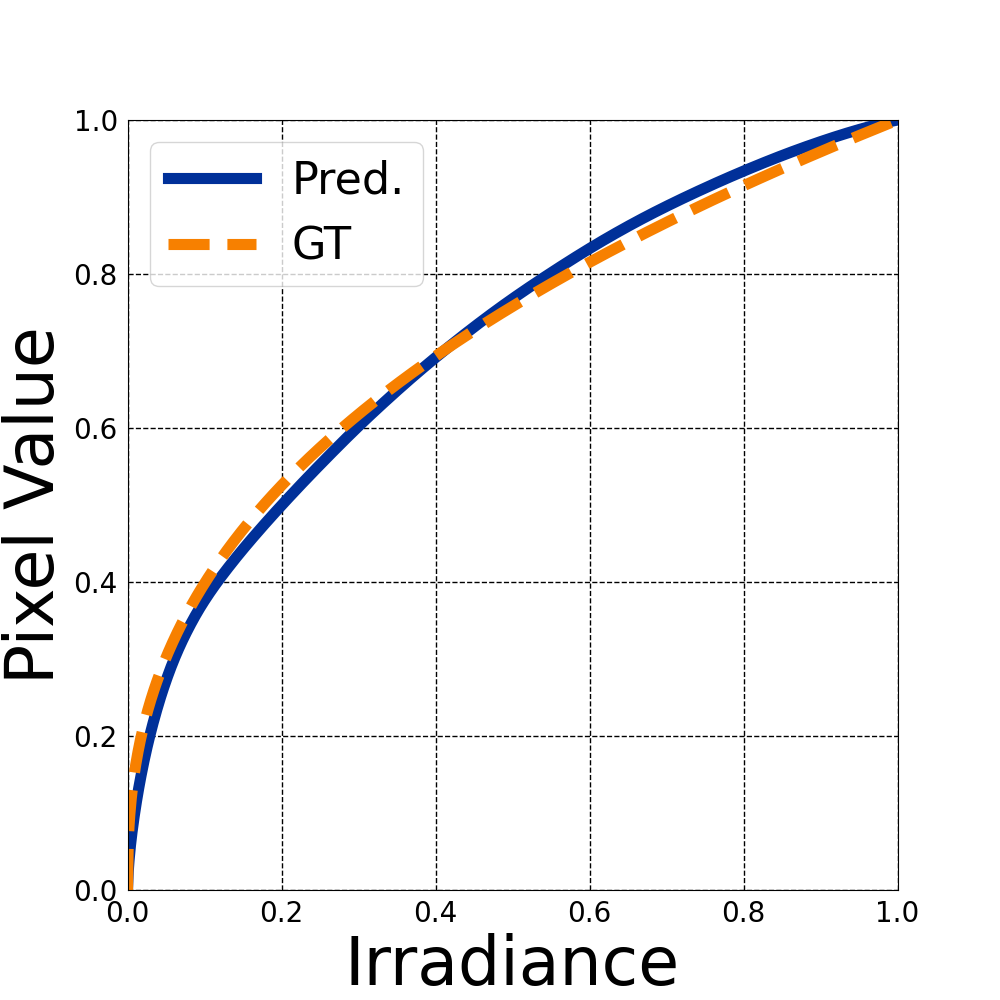} \\
      
    \end{tabular}%
    }
    \vspace{-2mm}
    \caption{\textbf{CRF estimation from real-world devices.}
    The orange dashed line is the ground truth CRF, and the blue line our estimate.
    }
    \label{fig:crf_varying}
\end{figure}

\begin{table}[t]
    \caption{\label{tab:nvs_synthetic}%
        \textbf{Quantitative results of novel-view synthesis (NVS) and relighting} on synthetic scenes with ground truth.
    }
    \centering\setlength{\tabcolsep}{4pt}
    \resizebox{0.8\linewidth}{!}{%
    \begin{tabular}{llcc@{~~}c@{~~}c}
    \toprule
    & Method & Input & PSNR $\uparrow$ & SSIM $\uparrow$ & LPIPS $\downarrow$ \\
    \midrule 
    \multirow{4}{*}{NVS}
    & NeILF~\cite{yao2022neilf}         & LDR & \bf 30.031 &     0.899 & \bf 0.180 \\
    & FIPT*                             & LDR &     15.180 &     0.744 &     0.437 \\
    & Ours                              & LDR &     29.468 & \bf 0.903 &     0.181 \\
    \cmidrule{2-6}
    & FIPT~\cite{wu2023factorized}      & HDR &     28.760 &     0.902 &     0.186 \\
    \midrule 
    \multirow{4}{*}{Relighting}
    & Li et al.~\cite{li2022physically} & LDR &     22.444 &     0.813 &     0.398 \\
    & FIPT*                             & LDR &     11.626 &     0.705 &     0.359 \\
    & Ours                              & LDR & \bf 22.861 & \bf 0.882 & \bf 0.173 \\
    \cmidrule{2-6}
    & FIPT~\cite{wu2023factorized}      & HDR &     29.012 &     0.902 &     0.139 \\
    \bottomrule 
    \end{tabular}%
    }
    \vspace{-3mm}
\end{table}

\subsection{Ablation Study}

\begin{table}
    \caption{\label{}%
        \textbf{Ablation study}
        }
    \vspace{-3mm}
    \centering\setlength{\tabcolsep}{4pt}
    \resizebox{1.0\linewidth}{!}{%
    \begin{tabular}{l|ccc|cc|c}
    \toprule
        & $\bk_\text{d}$ & $\ba'$ & $\sigma$ & \multicolumn{2}{c|}{$\bL_\text{e}$} & CRF \\
        Method & \multicolumn{3}{c|}{PSNR $\uparrow$} & IoU $\uparrow$ & L2 $\downarrow$ & L2 $\downarrow$ \\
    \midrule
     $-$ Emitter mask $M_e$ (\cref{eq:emitter_mask}) & 16.50 & 11.30 & \phantom{0}4.81 & 0.00 & 0.48 & 8.91 \\
     $-$ CRF modeling (\cref{eq:crf_parameter}) & 18.18 & 13.59 & 17.52 & \textbf{0.69} & \textbf{0.12} & --- \\
     $-$ HDR restoration (\cref{sec:hdr_restoration}) & 18.20 & 14.40 & \phantom{0}7.31 & \textbf{0.69} & 0.28 & 6.23 \\
     $- \mathcal{L}_{\text{albedo}}$ (\cref{eq:loss_albedo}) & 21.59 & 17.36 & 17.43 & \textbf{0.69} & 0.15 & 2.42 \\
     $- \mathcal{L}_{\text{CRF}}$ (\cref{eq:loss_crf}) & 14.49 & 12.81 & \textbf{25.59} & \textbf{0.69} & 0.13 & 9.40 \\
    Full model (Ours) & \textbf{22.33} &  \textbf{17.92} & 21.38 & \textbf{0.69} & \textbf{0.12} & \textbf{2.35} \\
       
    \bottomrule
    \end{tabular}%
    }
    \label{tab:ablation}
    \vspace{-5mm}
\end{table}

We validate our design choices and provide an ablation study in \cref{tab:ablation}, where we compute PSNR for material ($\bk_\text{d}$, $\mathbf{a}'$, $\sigma$) and L2 for CRF curves and emission maps $\bL_\text{e}$.
The emitter mask estimation in \cref{eq:emitter_mask} identifies the emitter from LDR observations and improves the emission map in the second row.
CRF estimation is critical for modeling HDR-to-LDR conversion and helps inverse rendering in the third row.
Nevertheless, the HDR restoration (\cref{sec:hdr_restoration}) plays an important role in recovering accurate light transport from the captured images.
Once the HDR light is restored by our proposed approach, the material (especially roughness $\sigma$), CRF, and emission map are improved significantly.
The results demonstrate the significance of each component and show that inverse rendering from casually captured LDR images benefits from our proposed framework.

\subsection{CRF Estimation}
To further validate the robustness of our method, we generate LDR images from FIPT’s synthetic HDR scenes using different CRFs collected from real-world devices \cite{grossberg2003space}.
In this evaluation, we perform inverse rendering with IRIS and plot the CRF.
The results in \cref{fig:crf_varying} show that our method can generalize to various imaging pipelines.
Please refer to the supplementary details for more analysis of CRF estimation.

\section{Conclusion}
We presented IRIS, an inverse rendering framework for indoor scenes from casually captured LDR images.
Our key innovations include HDR lighting restoration, CRF estimation, surface material estimation, and alternating optimization to recover all scene properties accurately.
IRIS estimates accurate surface material, spatially varying HDR illumination, and CRF from real-world LDR images. 
As a result, IRIS enables photorealistic rendering under diverse lighting and from arbitrary viewpoints, showcasing the significant potential for content creation applications.

\clearpage
\maketitlesupplementary
\appendix

\begin{abstract}
This supplementary document shows additional details of our method and more results.
We refer readers to our webpage, which shows more results that allow for easy comparisons with the baseline methods on all scenes we use.
\end{abstract}

\begin{figure*}[t]
    \centering\setlength{\tabcolsep}{0.1em}
    \resizebox{1.0\textwidth}{!}{
    \begin{tabular}{lccccc}
       & $t=1$ & $t=2$ & $t=3$ & $t=4$ & $t=5$ \\
       
       \raisebox{4.0\normalbaselineskip}[0pt][0pt]{\rotatebox[origin=c]{90}{ Reconstruction}} & 
       \frame{\includegraphics[width=0.3\textwidth]{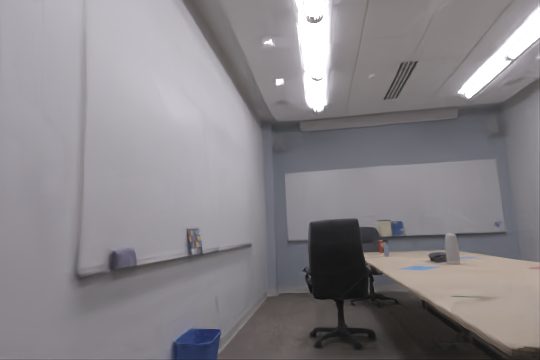}} &\frame{\includegraphics[width=0.3\textwidth]{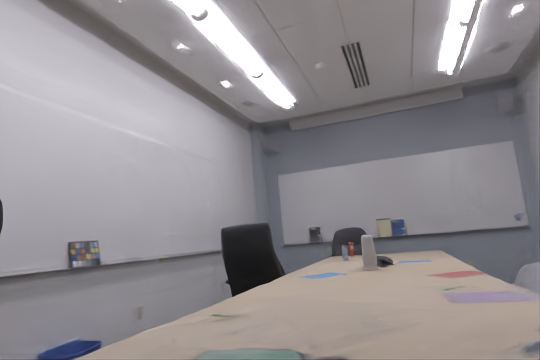}} &\frame{\includegraphics[width=0.3\textwidth]{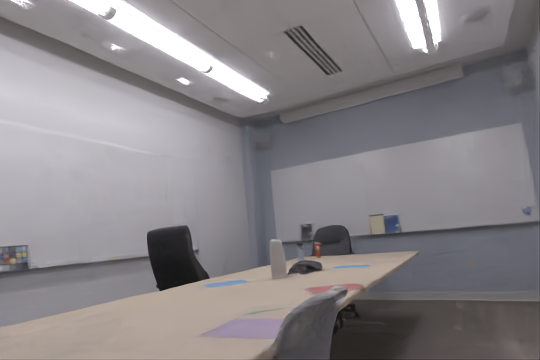}} &\frame{\includegraphics[width=0.3\textwidth]{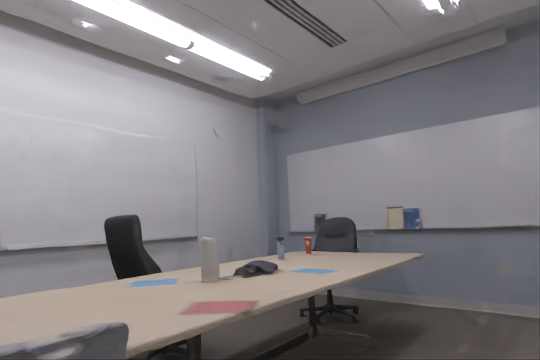}} &\frame{\includegraphics[width=0.3\textwidth]{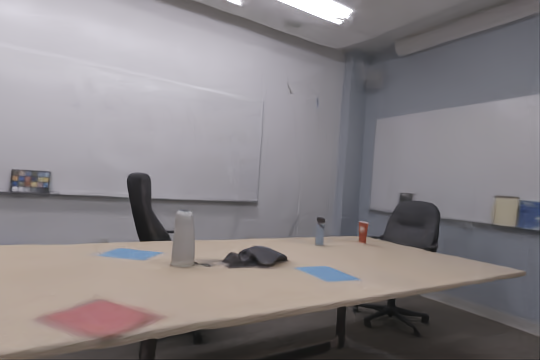}} \\

       \raisebox{4.0\normalbaselineskip}[0pt][0pt]{\rotatebox[origin=c]{90}{ Relighting 1}} & 
       \frame{\includegraphics[width=0.3\textwidth]{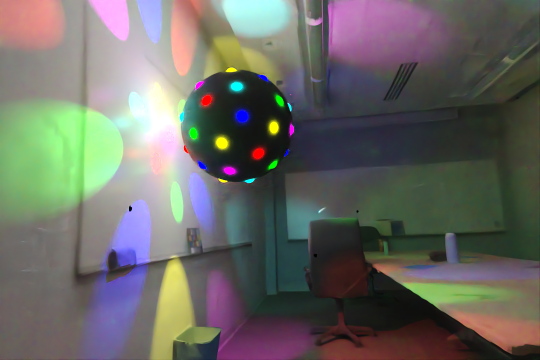}} &\frame{\includegraphics[width=0.3\textwidth]{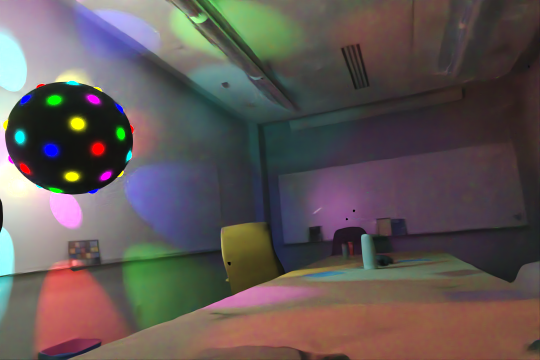}} &\frame{\includegraphics[width=0.3\textwidth]{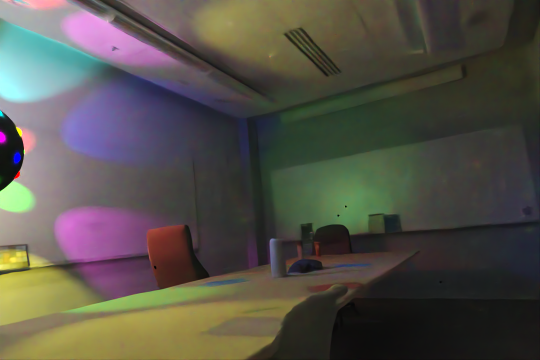}} &\frame{\includegraphics[width=0.3\textwidth]{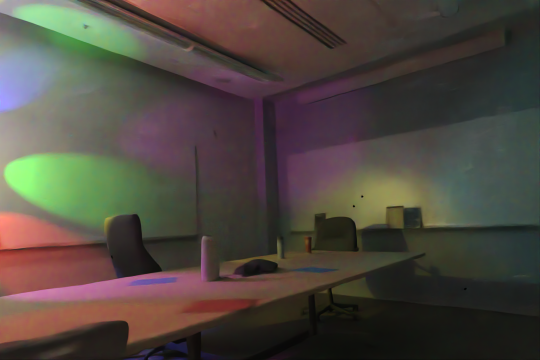}} &\frame{\includegraphics[width=0.3\textwidth]{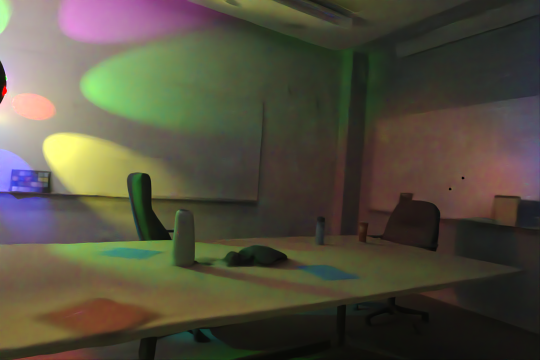}} \\

       \raisebox{4.0\normalbaselineskip}[0pt][0pt]{\rotatebox[origin=c]{90}{ Relighting 2}} & 
       \frame{\includegraphics[width=0.3\textwidth]{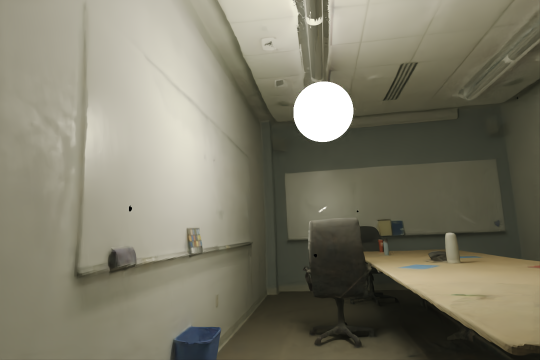}} &\frame{\includegraphics[width=0.3\textwidth]{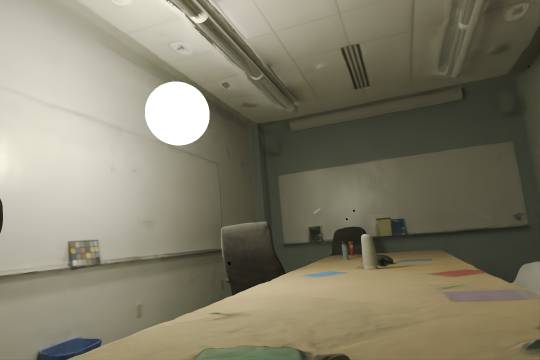}} &\frame{\includegraphics[width=0.3\textwidth]{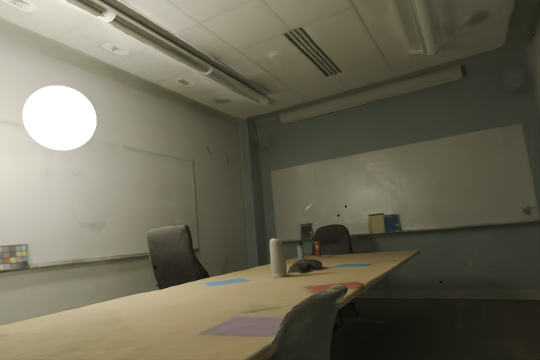}} &\frame{\includegraphics[width=0.3\textwidth]{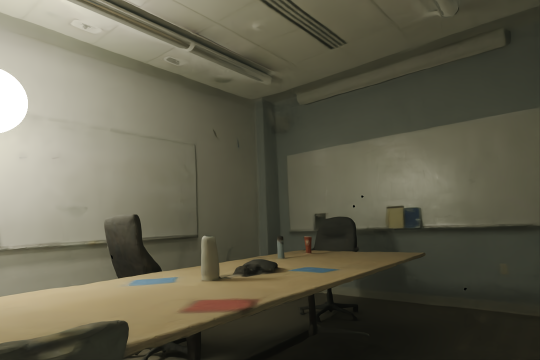}} &\frame{\includegraphics[width=0.3\textwidth]{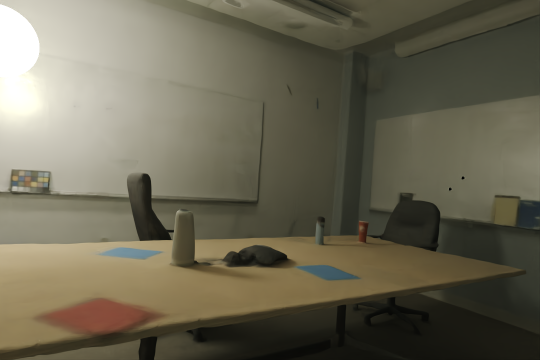}} \\

      \raisebox{4.0\normalbaselineskip}[0pt][0pt]{\rotatebox[origin=c]{90}{Insertion}} & 
       \frame{\includegraphics[width=0.3\textwidth]{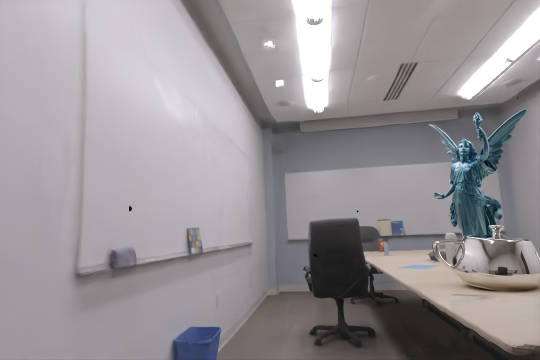}} &\frame{\includegraphics[width=0.3\textwidth]{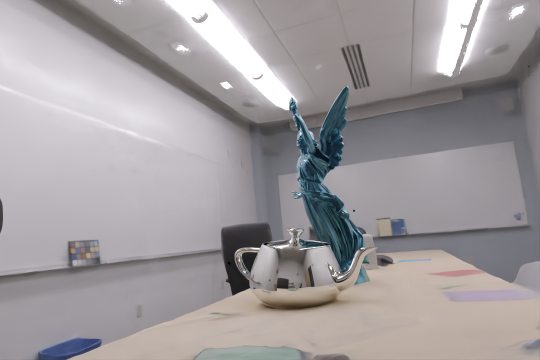}} &\frame{\includegraphics[width=0.3\textwidth]{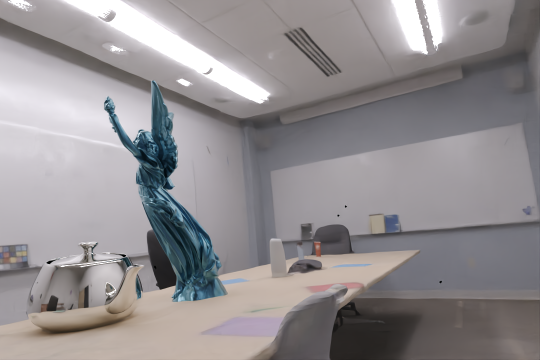}} &\frame{\includegraphics[width=0.3\textwidth]{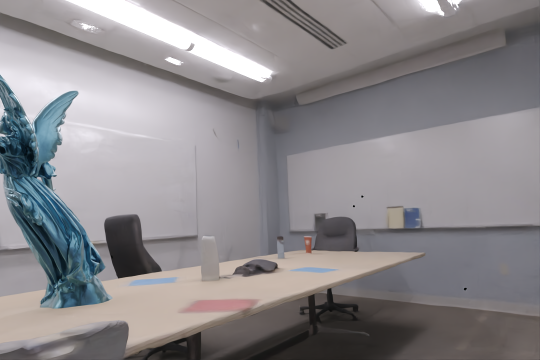}} &\frame{\includegraphics[width=0.3\textwidth]{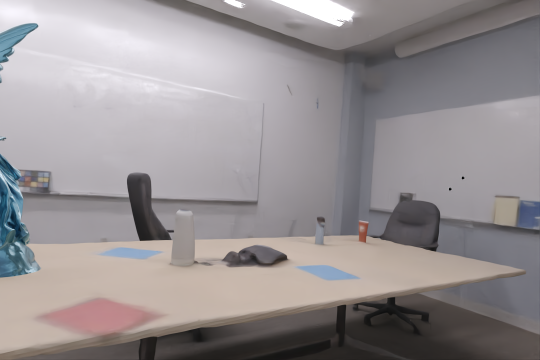}} \\
        
    \end{tabular}
    }
    \caption{\textbf{Relighting and object insertion in `conference room`.}  The inserted new light sources are reflected on the whiteboard surface, demonstrating the accuracy of the material estimation of IRIS.}
    \label{tab:supp_relight_conferenceroom}
\end{figure*}

\begin{figure*}[t]
    \centering\setlength{\tabcolsep}{0.1em}
    \resizebox{1.0\textwidth}{!}{
    \begin{tabular}{lccccc}
       & $t=1$ & $t=2$ & $t=3$ & $t=4$ & $t=5$ \\
       
       \raisebox{4.0\normalbaselineskip}[0pt][0pt]{\rotatebox[origin=c]{90}{Reconstruction}} & 
       \frame{\includegraphics[width=0.3\textwidth]{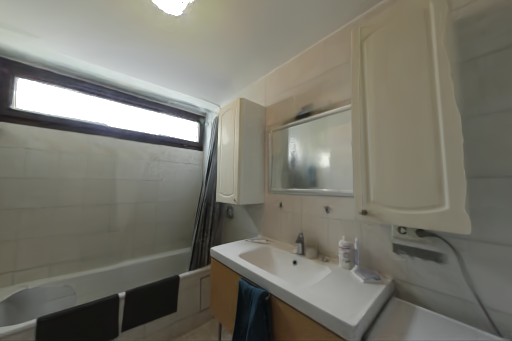}} &\frame{\includegraphics[width=0.3\textwidth]{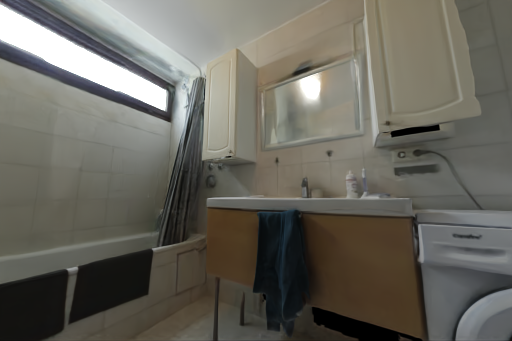}} &\frame{\includegraphics[width=0.3\textwidth]{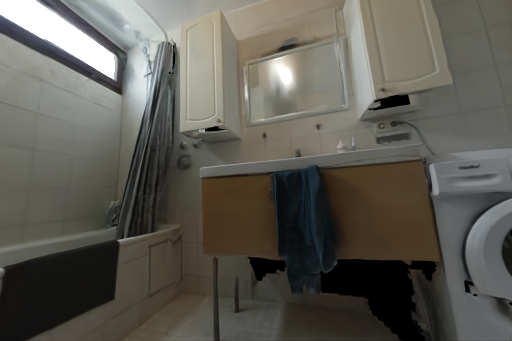}} &\frame{\includegraphics[width=0.3\textwidth]{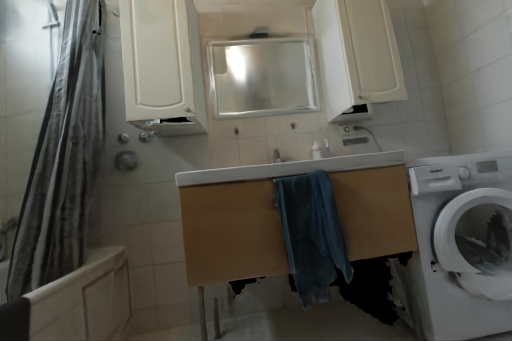}} &\frame{\includegraphics[width=0.3\textwidth]{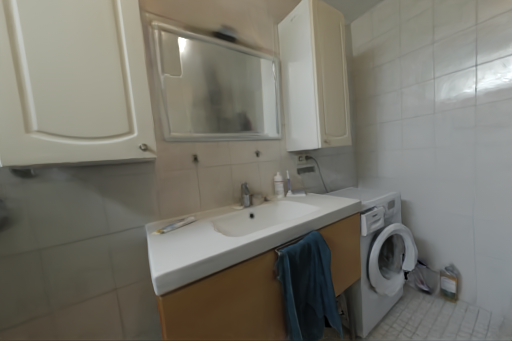}} \\

       \raisebox{4.0\normalbaselineskip}[0pt][0pt]{\rotatebox[origin=c]{90}{Relighting 1}} & 
       \frame{\includegraphics[width=0.3\textwidth]{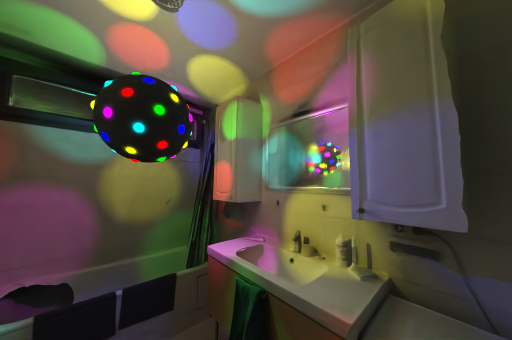}} &\frame{\includegraphics[width=0.3\textwidth]{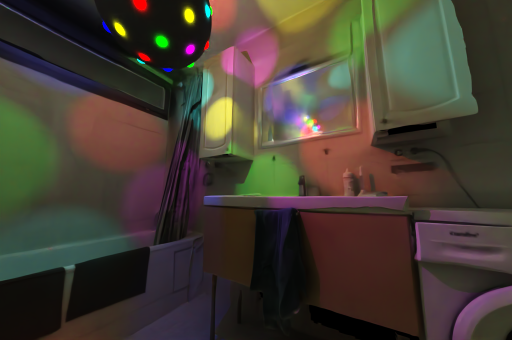}} &\frame{\includegraphics[width=0.3\textwidth]{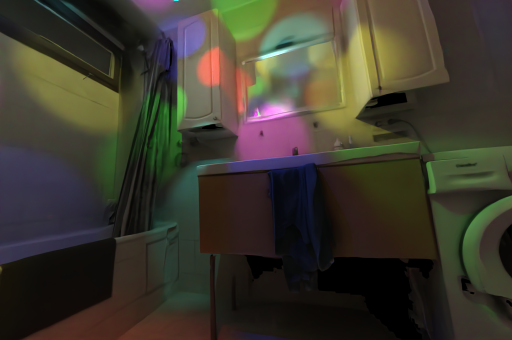}} &\frame{\includegraphics[width=0.3\textwidth]{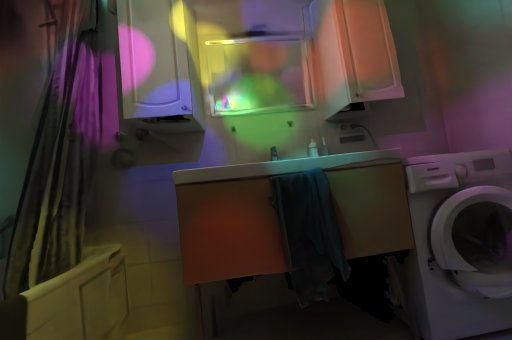}} &\frame{\includegraphics[width=0.3\textwidth]{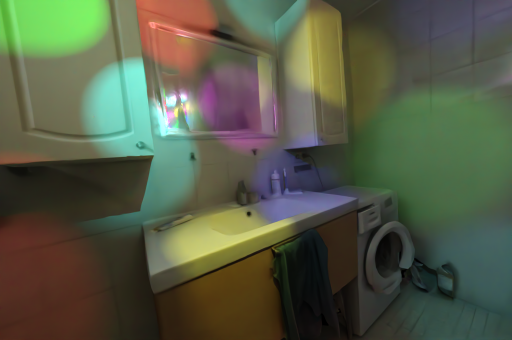}} \\

       \raisebox{4.0\normalbaselineskip}[0pt][0pt]{\rotatebox[origin=c]{90}{Relighting 2}} & 
       \frame{\includegraphics[width=0.3\textwidth]{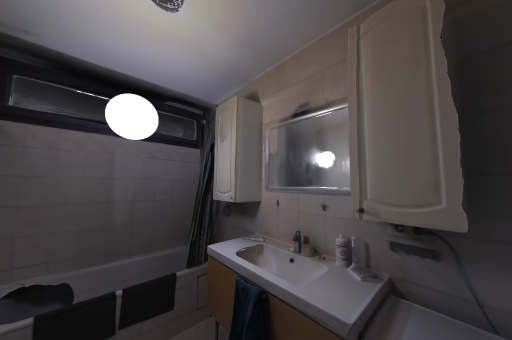}} &\frame{\includegraphics[width=0.3\textwidth]{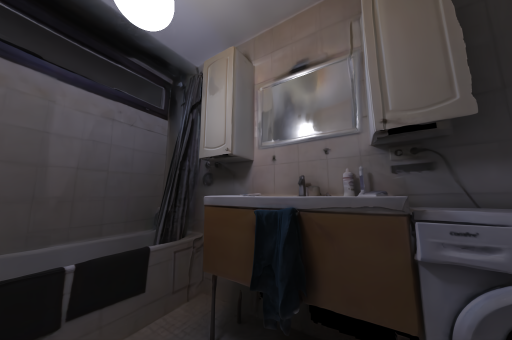}} &\frame{\includegraphics[width=0.3\textwidth]{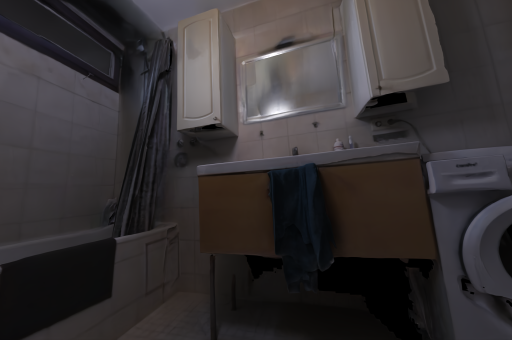}} &\frame{\includegraphics[width=0.3\textwidth]{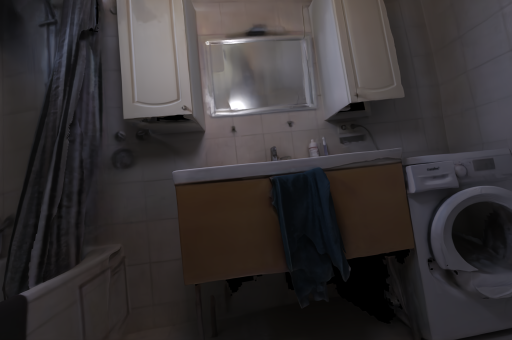}} &\frame{\includegraphics[width=0.3\textwidth]{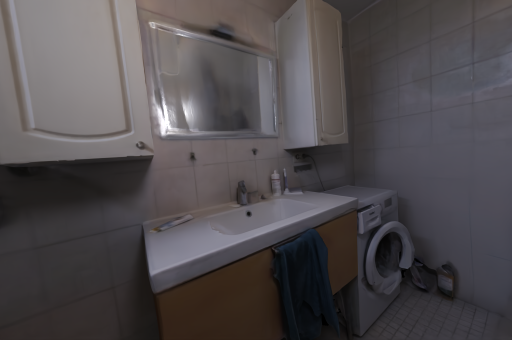}} \\

       \raisebox{4.0\normalbaselineskip}[0pt][0pt]{\rotatebox[origin=c]{90}{Insertion}} & 
       \frame{\includegraphics[width=0.3\textwidth]{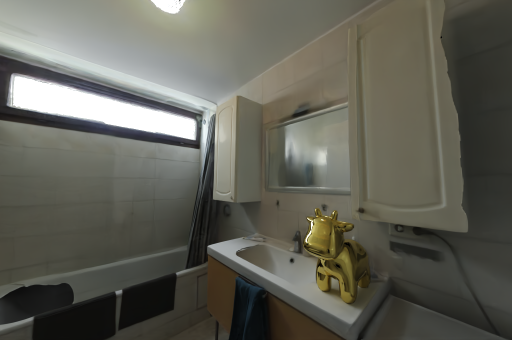}} &\frame{\includegraphics[width=0.3\textwidth]{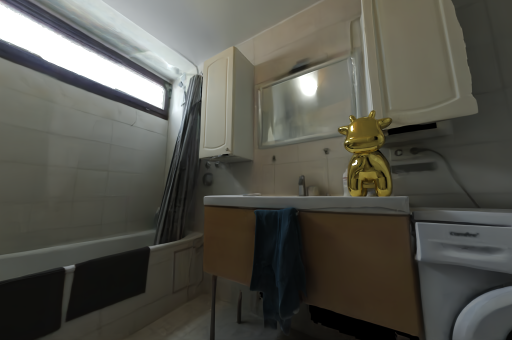}} &\frame{\includegraphics[width=0.3\textwidth]{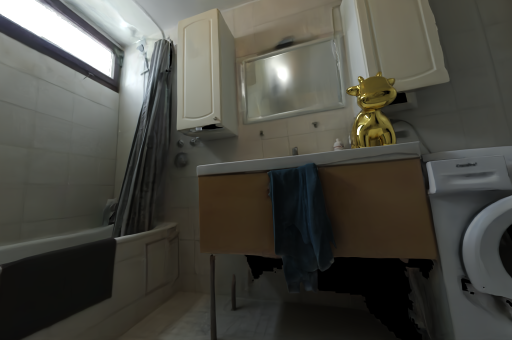}} &\frame{\includegraphics[width=0.3\textwidth]{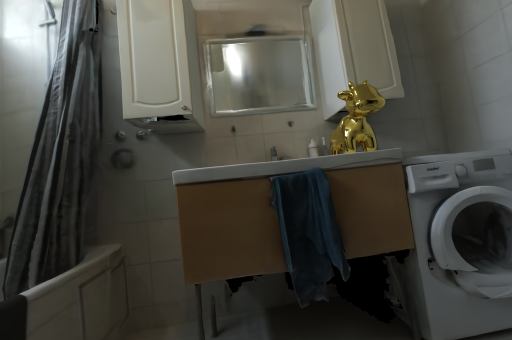}} &\frame{\includegraphics[width=0.3\textwidth]{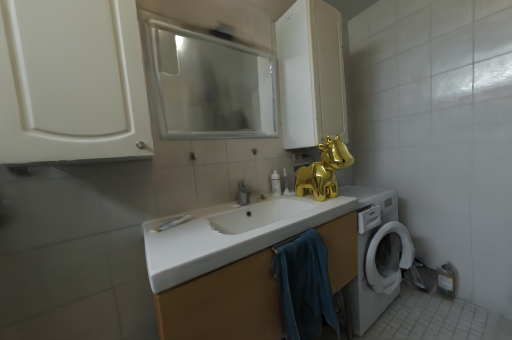}} \\

    \end{tabular}
    }
    \caption{\textbf{Relighting and object insertion in `bathroom`.} The mirror is estimated as a low-roughness surface, and it reflects the new light sources and enhances the realism of relighting significantly. The inserted object also exhibits reflection of HDR lighting recovered by IRIS.}
    \label{tab:supp_relight_bathroom2}
\end{figure*}

\begin{figure*}[t]
    \centering\setlength{\tabcolsep}{0.1em}
    \resizebox{1.0\textwidth}{!}{
    \begin{tabular}{lccccc}
       & $t=1$ & $t=2$ & $t=3$ & $t=4$ & $t=5$ \\
       
       \raisebox{4.0\normalbaselineskip}[0pt][0pt]{\rotatebox[origin=c]{90}{Reconstruction}} & 
       \frame{\includegraphics[width=0.3\textwidth]{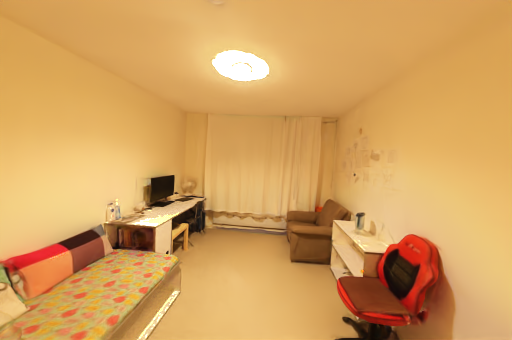}} &\frame{\includegraphics[width=0.3\textwidth]{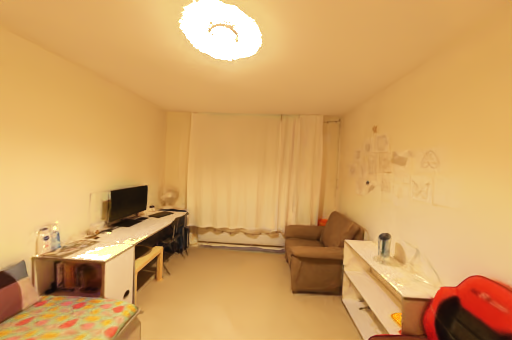}} &\frame{\includegraphics[width=0.3\textwidth]{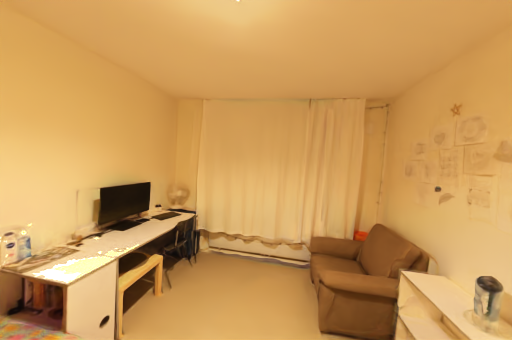}} &\frame{\includegraphics[width=0.3\textwidth]{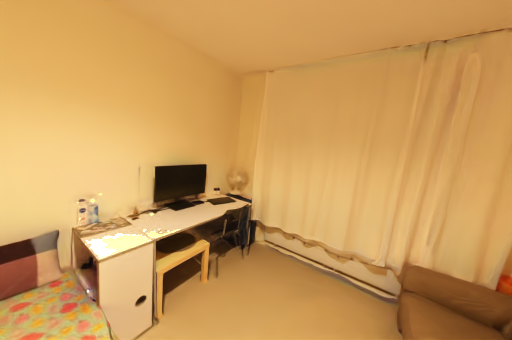}} &\frame{\includegraphics[width=0.3\textwidth]{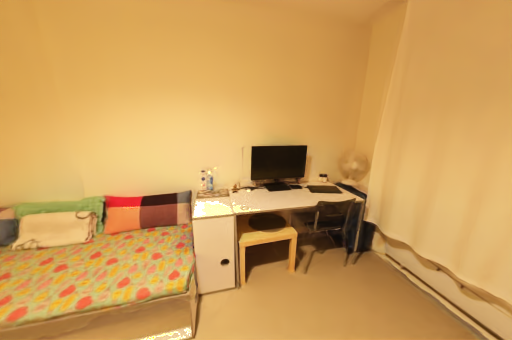}} \\

       \raisebox{4.0\normalbaselineskip}[0pt][0pt]{\rotatebox[origin=c]{90}{Relighting 1}} & 
       \frame{\includegraphics[width=0.3\textwidth]{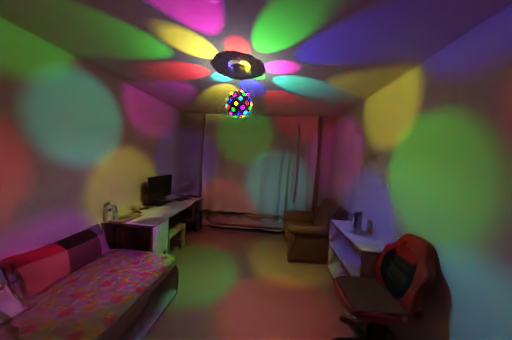}} &\frame{\includegraphics[width=0.3\textwidth]{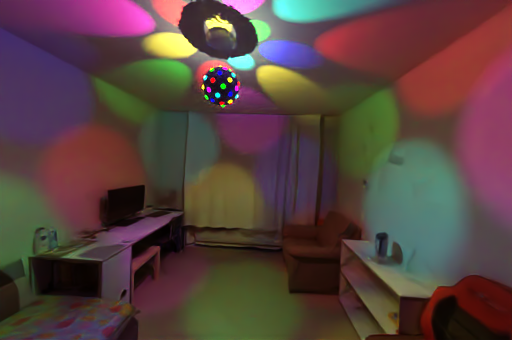}} &\frame{\includegraphics[width=0.3\textwidth]{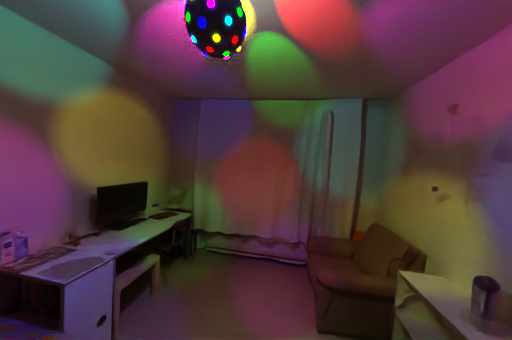}} &\frame{\includegraphics[width=0.3\textwidth]{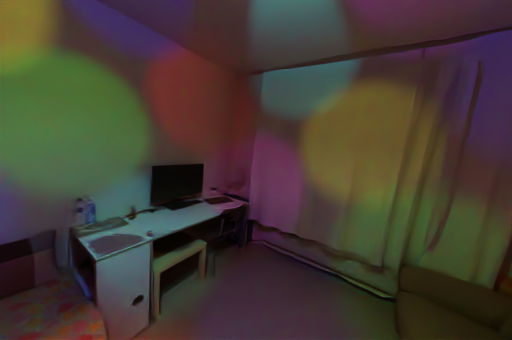}} &\frame{\includegraphics[width=0.3\textwidth]{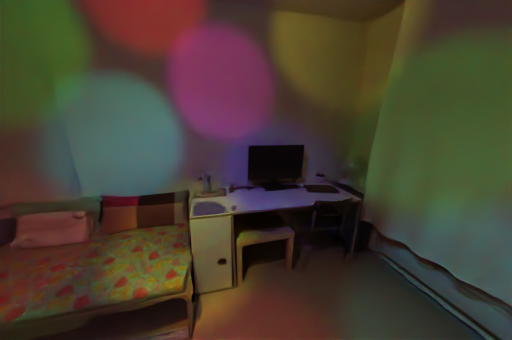}} \\

       \raisebox{4.0\normalbaselineskip}[0pt][0pt]{\rotatebox[origin=c]{90}{Relighting 2}} & 
       \frame{\includegraphics[width=0.3\textwidth]{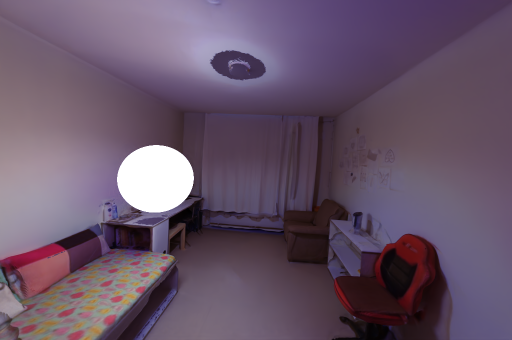}} &\frame{\includegraphics[width=0.3\textwidth]{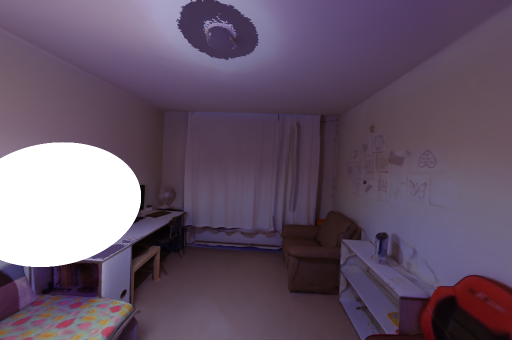}} &\frame{\includegraphics[width=0.3\textwidth]{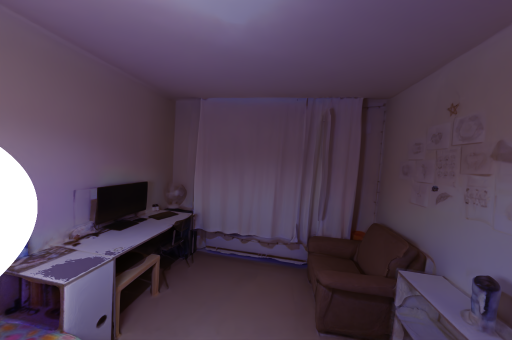}} &\frame{\includegraphics[width=0.3\textwidth]{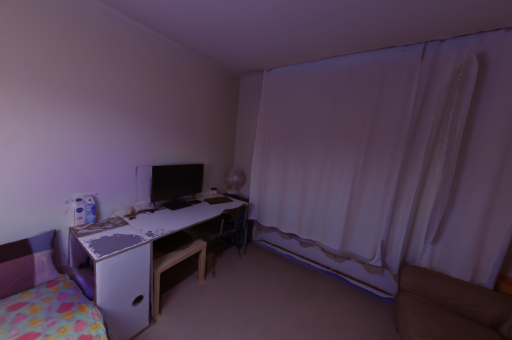}} &\frame{\includegraphics[width=0.3\textwidth]{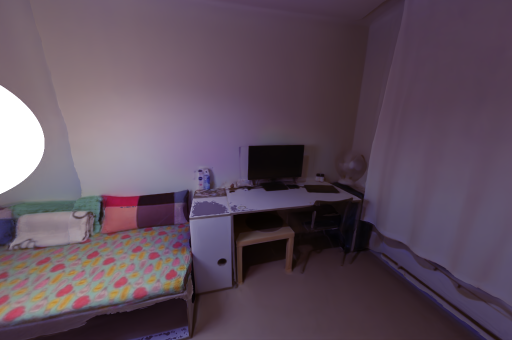}} \\

       \raisebox{4.0\normalbaselineskip}[0pt][0pt]{\rotatebox[origin=c]{90}{Insertion}} & 
       \frame{\includegraphics[width=0.3\textwidth]{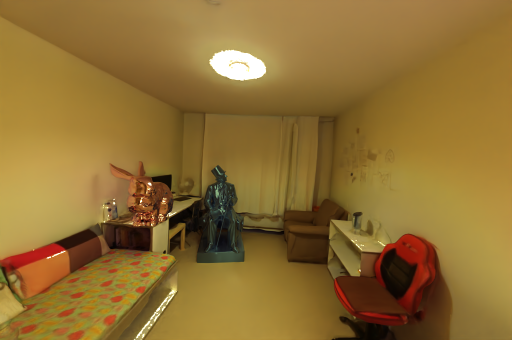}} &\frame{\includegraphics[width=0.3\textwidth]{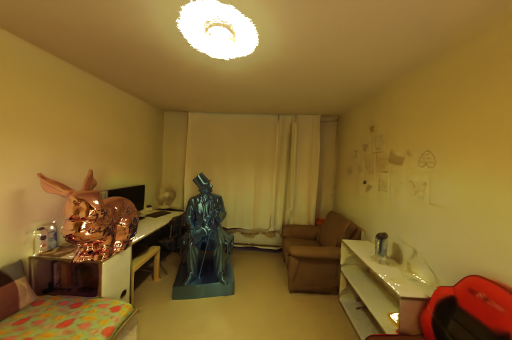}} &\frame{\includegraphics[width=0.3\textwidth]{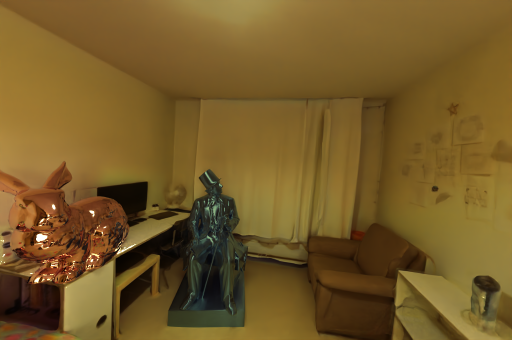}} &\frame{\includegraphics[width=0.3\textwidth]{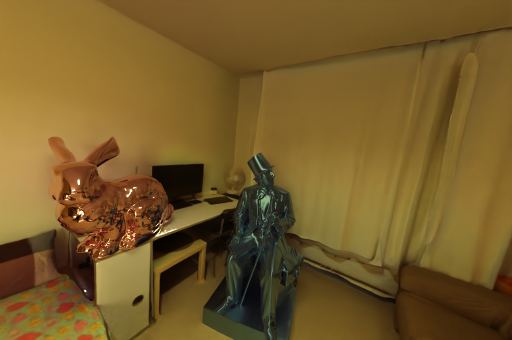}} &\frame{\includegraphics[width=0.3\textwidth]{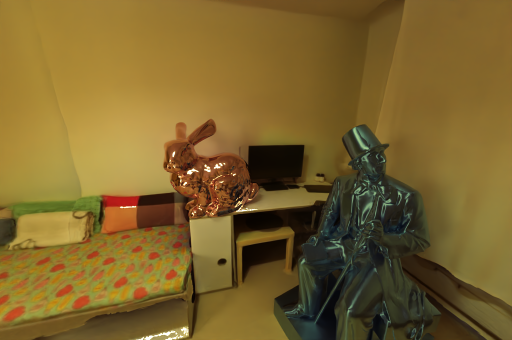}} \\

    \end{tabular}
    }
    \caption{\textbf{Relighting and object insertion in `bedroom`.} The Disco ball rotates and casts colorful lights in different directions, creating realistic relighting results in the real-world scene.}
    \label{tab:supp_relight_room2}
\end{figure*}

\section{Relighting and Object Insertion Results}

\begin{figure}[h]
    \centering\setlength{\tabcolsep}{0.1 em}
    \resizebox{1.0\linewidth}{!}{%
    \begin{tabular}{@{}clcccc@{}}
    
    \footnotesize{Input} & & \footnotesize{Recon.} & \footnotesize{Diffuse $\bk_\text{d}$} & \footnotesize{Roughness $\sigma$} & \footnotesize{Relight.} \\ %

    \frame{\includegraphics[width=0.2\linewidth]{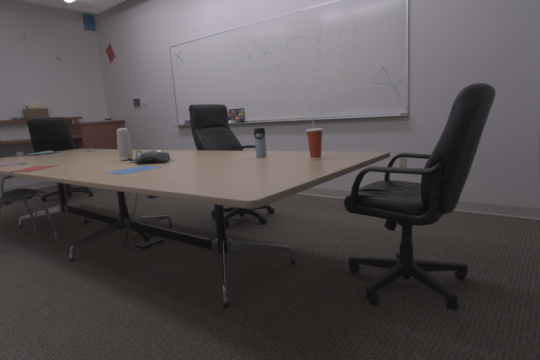}}& \raisebox{1.2\normalbaselineskip}[0pt][0pt]{\rotatebox[origin=c]{90}{\footnotesize{Li et al.}}} & \frame{\includegraphics[width=0.2\linewidth]{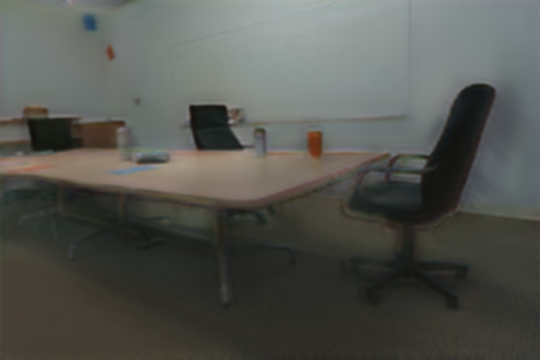}} & \frame{\includegraphics[width=0.2\linewidth]{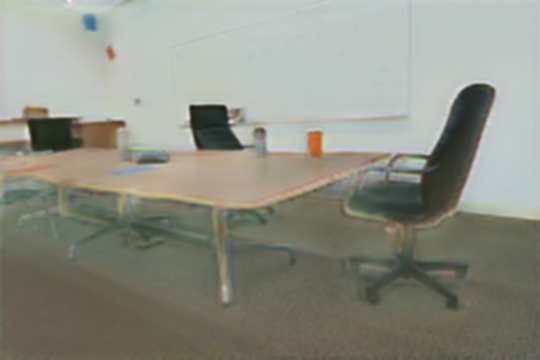}} & \frame{\includegraphics[width=0.2\linewidth]{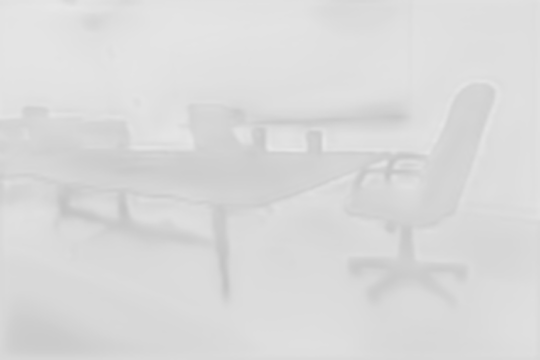}} & \frame{\includegraphics[width=0.2\linewidth]{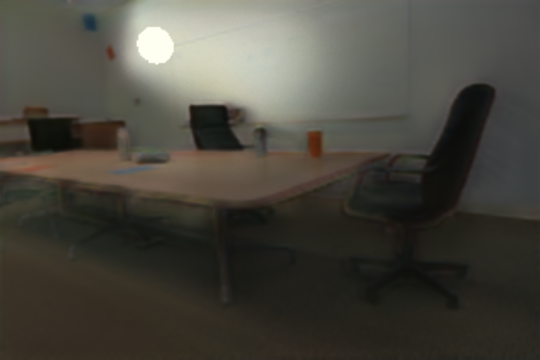}} \\
     & \raisebox{1.2\normalbaselineskip}[0pt][0pt]{\rotatebox[origin=c]{90}{\footnotesize{NeILF++}}} & \frame{\includegraphics[width=0.2\linewidth]{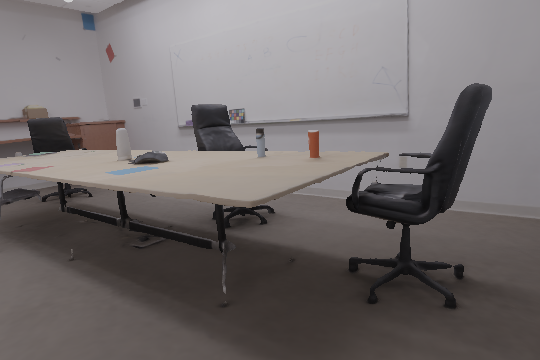}} & \frame{\includegraphics[width=0.2\linewidth]{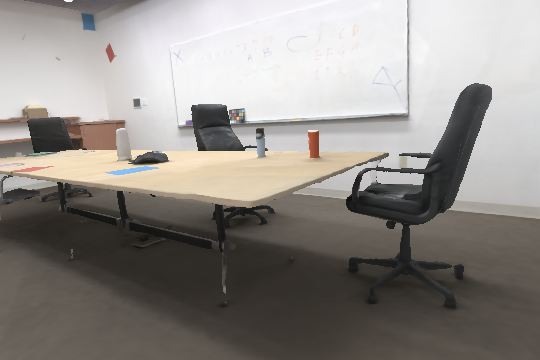}} & \frame{\includegraphics[width=0.2\linewidth]{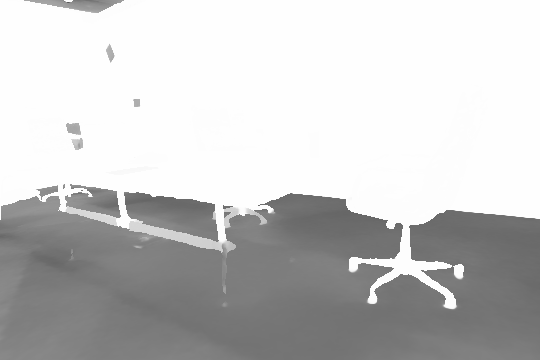}} & \frame{\includegraphics[width=0.2\linewidth]{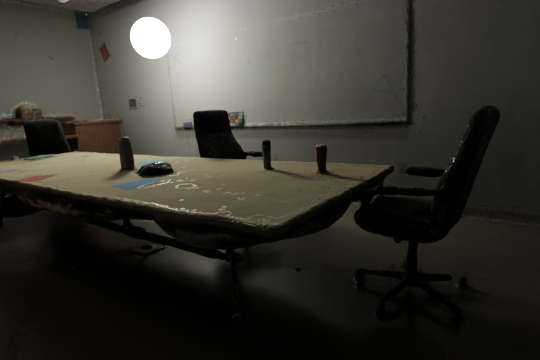}} \\
     & \raisebox{1.2\normalbaselineskip}[0pt][0pt]{\rotatebox[origin=c]{90}{\footnotesize{Ours}}} & \frame{\includegraphics[width=0.2\linewidth]{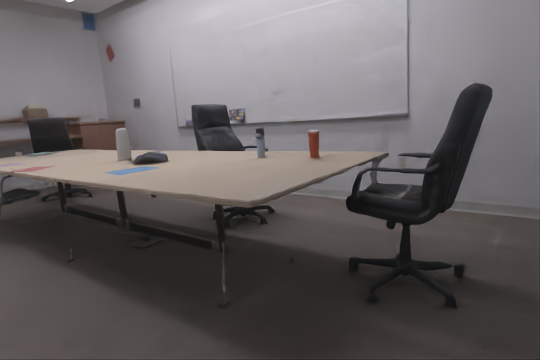}} & \frame{\includegraphics[width=0.2\linewidth]{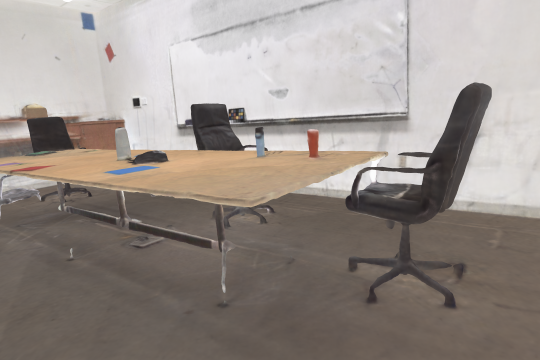}} & \frame{\includegraphics[width=0.2\linewidth]{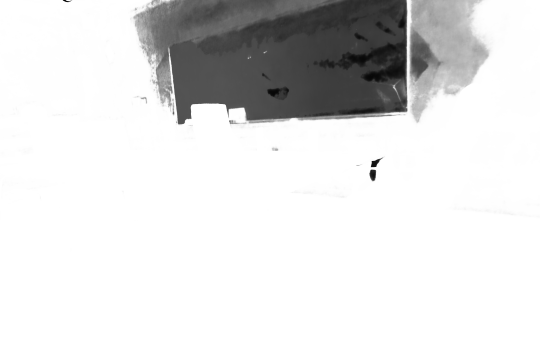}} & \frame{\includegraphics[width=0.2\linewidth]{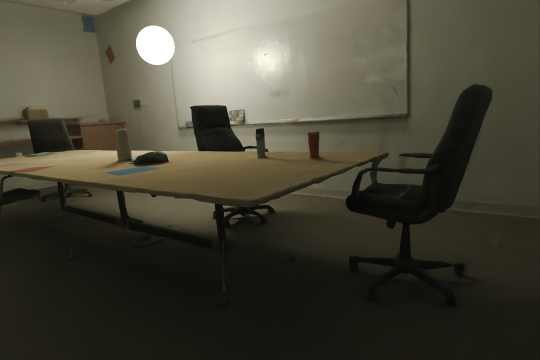}} \\
    
    \end{tabular}%
    }
    \vspace{-3mm}
    \caption{\textbf{Comparison with Li et al~\cite{li2022physically} and NeILF++~\cite{zhang2023neilf++}.}}
    \label{fig:compare_li_neilfpp}
    \vspace{-3mm}
\end{figure}

\begin{figure}[t]
    \centering\setlength{\tabcolsep}{0.1em}
    \resizebox{1.0\linewidth}{!}{
    \begin{tabular}{lcccc}
       & $t=0$ & $t=1$ & $t=2$ & $t=3$\\
       
       \raisebox{2\normalbaselineskip}[0pt][0pt]{\rotatebox[origin=c]{90}{ Recon.}}  
       & 
       \frame{\includegraphics[width=0.3\textwidth]{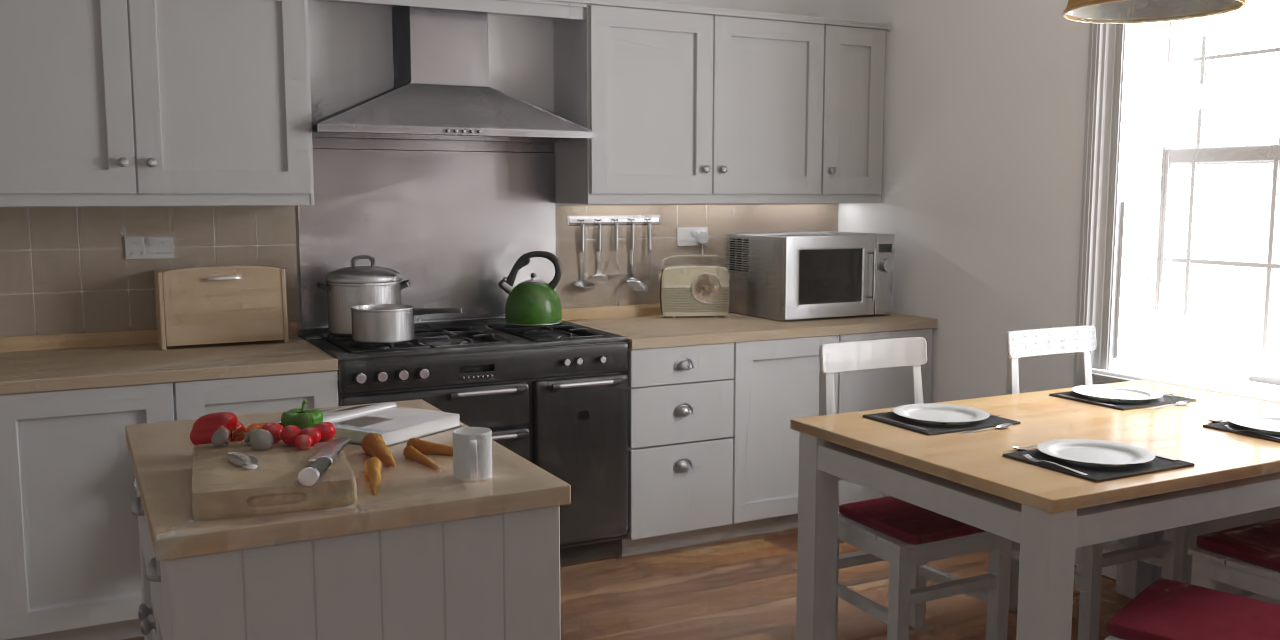}} & \frame{\includegraphics[width=0.3\textwidth]{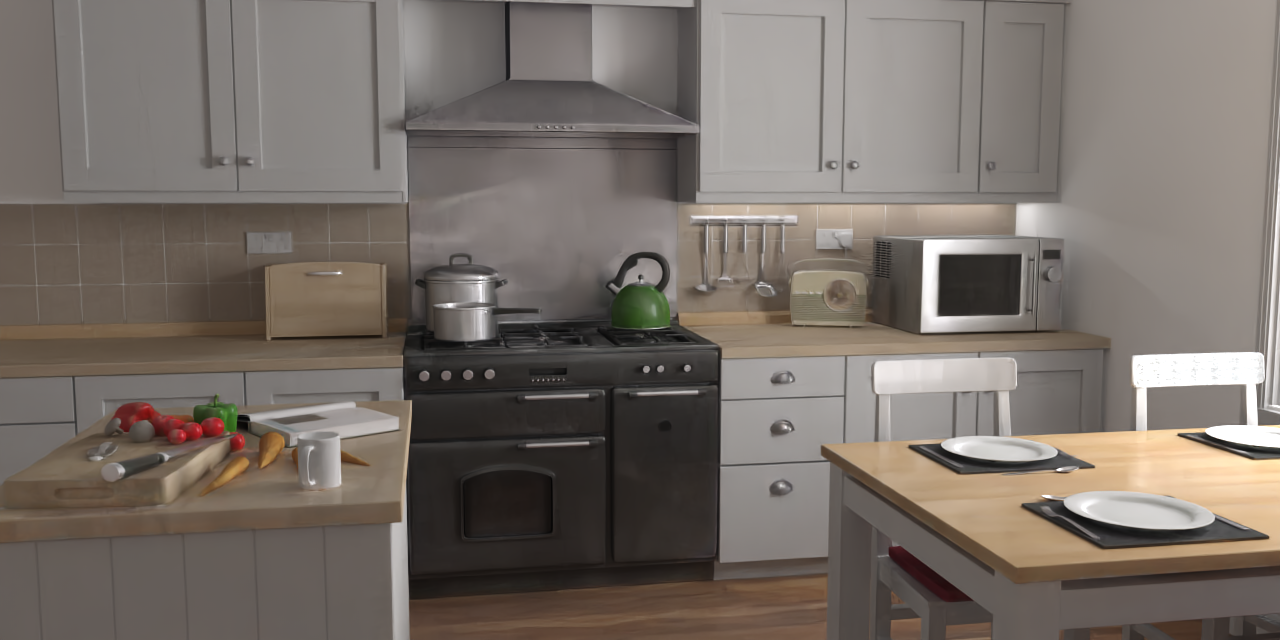}} & \frame{\includegraphics[width=0.3\textwidth]{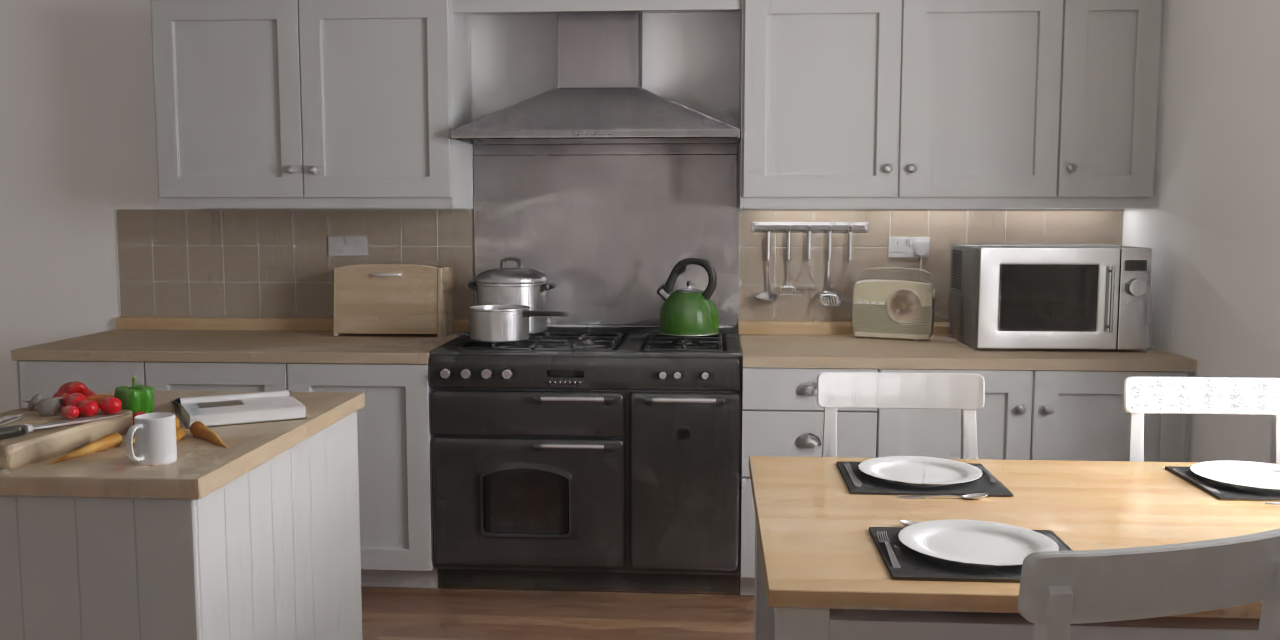}} & \frame{\includegraphics[width=0.3\textwidth]{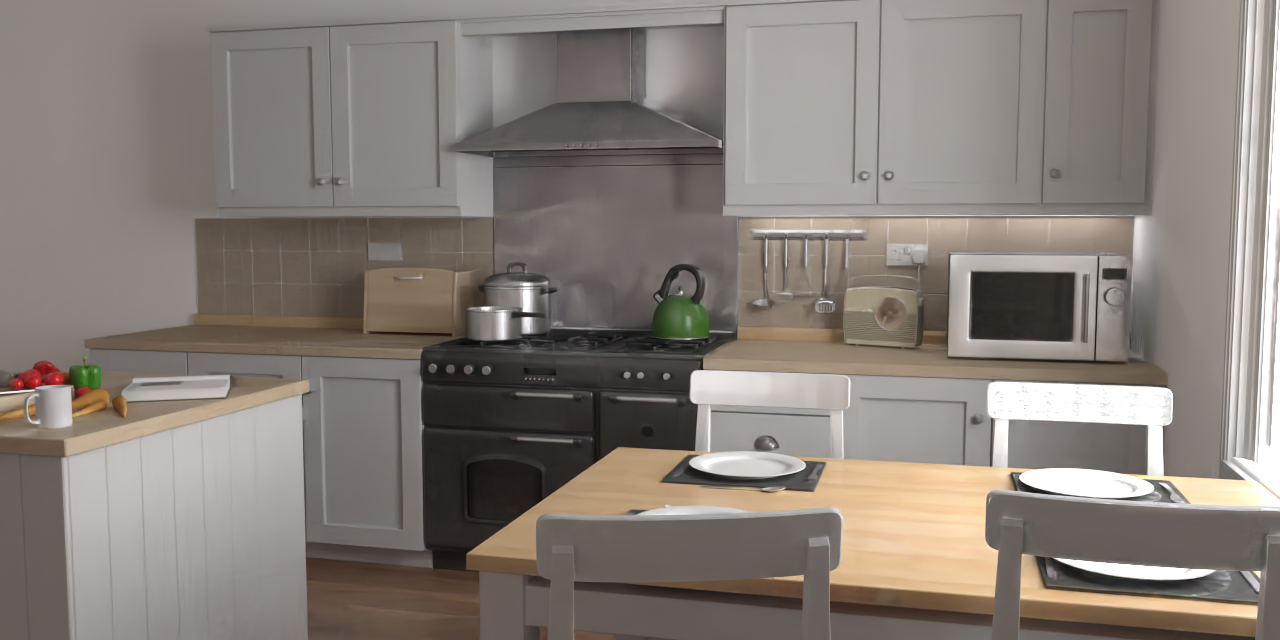}}  \\

        \raisebox{2\normalbaselineskip}[0pt][0pt]{\rotatebox[origin=c]{90}{ Relight}} 
        & \frame{\includegraphics[width=0.3\textwidth]{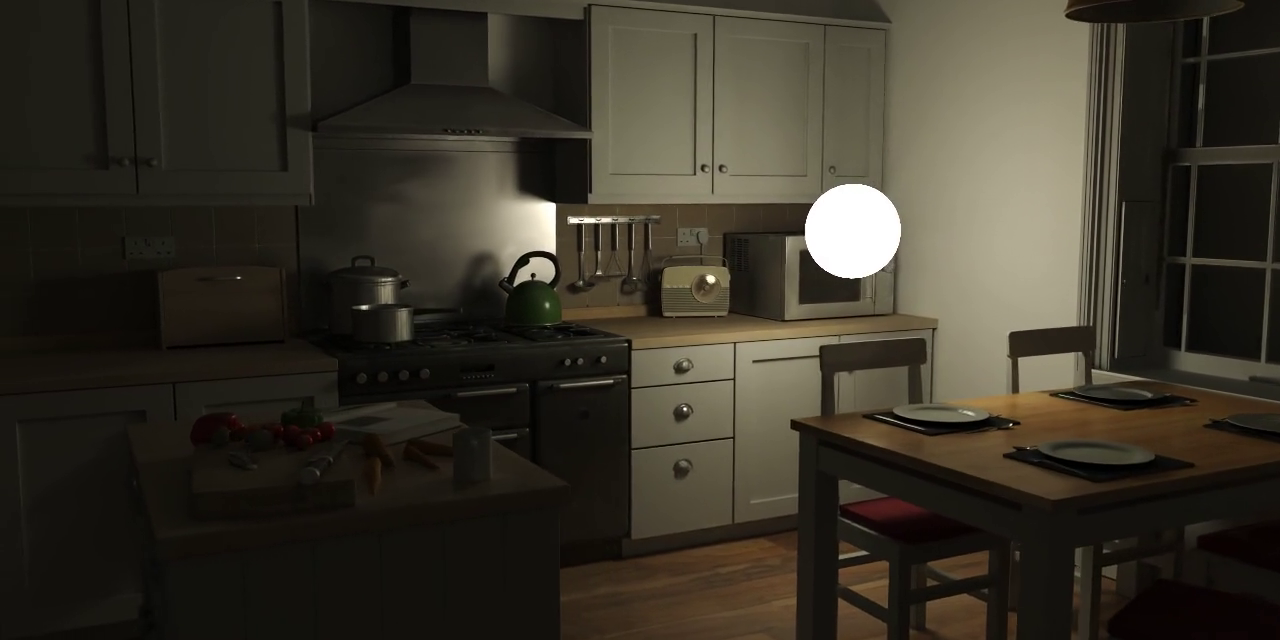}} & \frame{\includegraphics[width=0.3\textwidth]{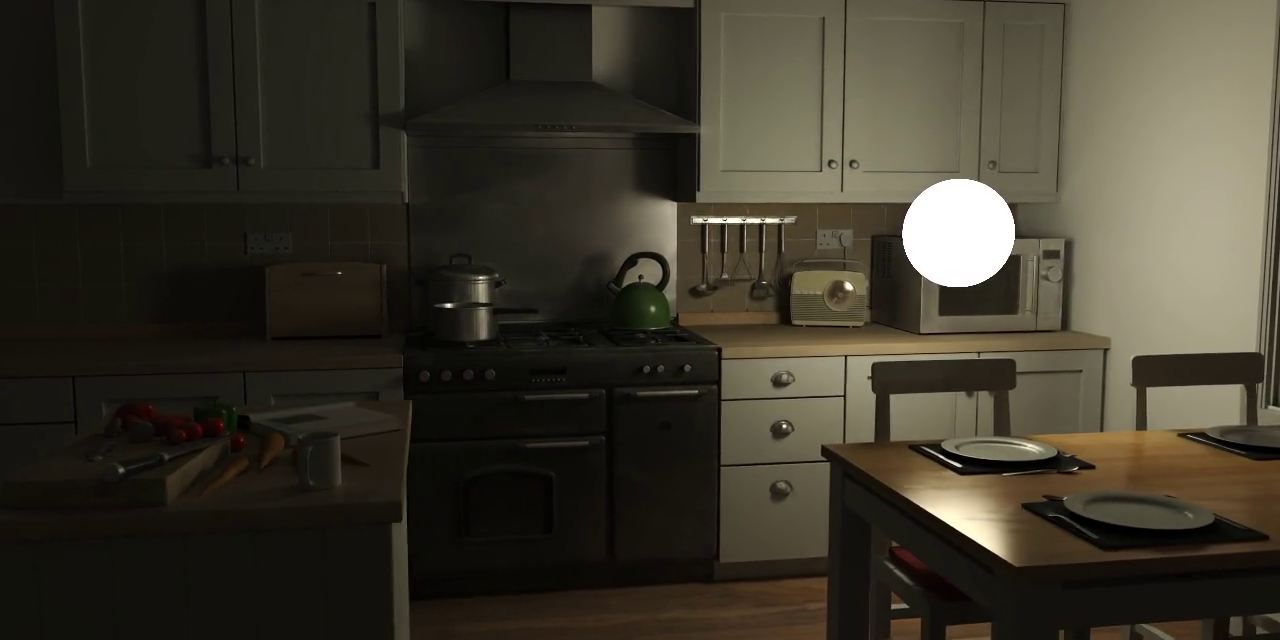}} & \frame{\includegraphics[width=0.3\textwidth]{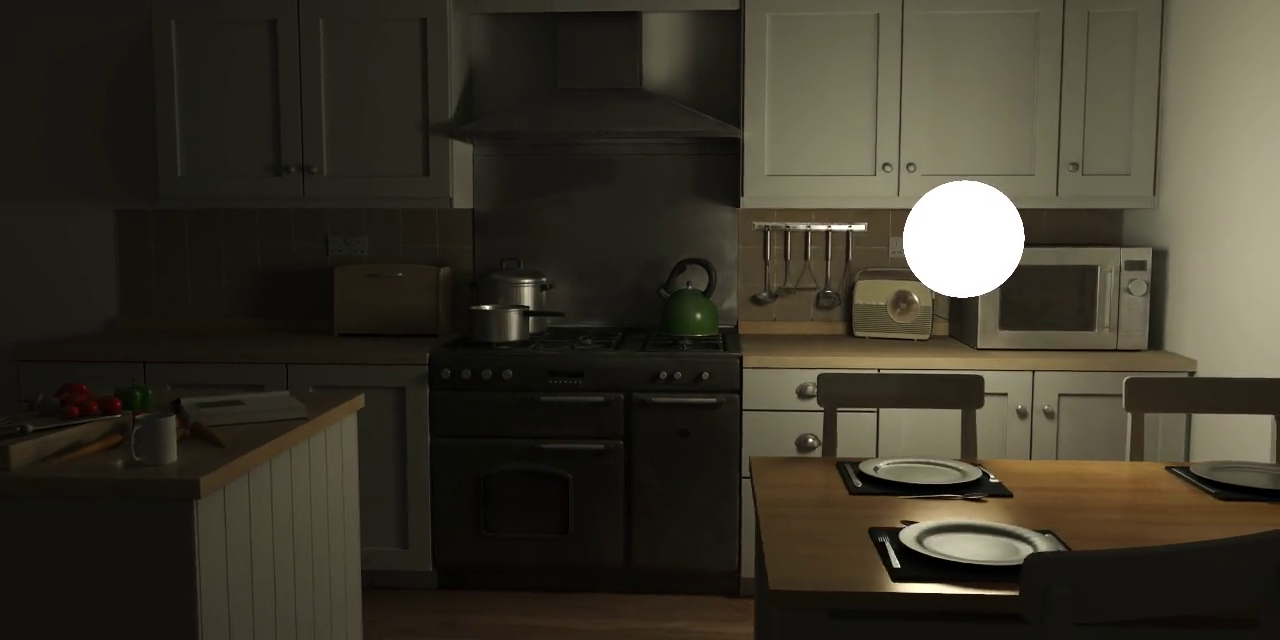}} & \frame{\includegraphics[width=0.3\textwidth]{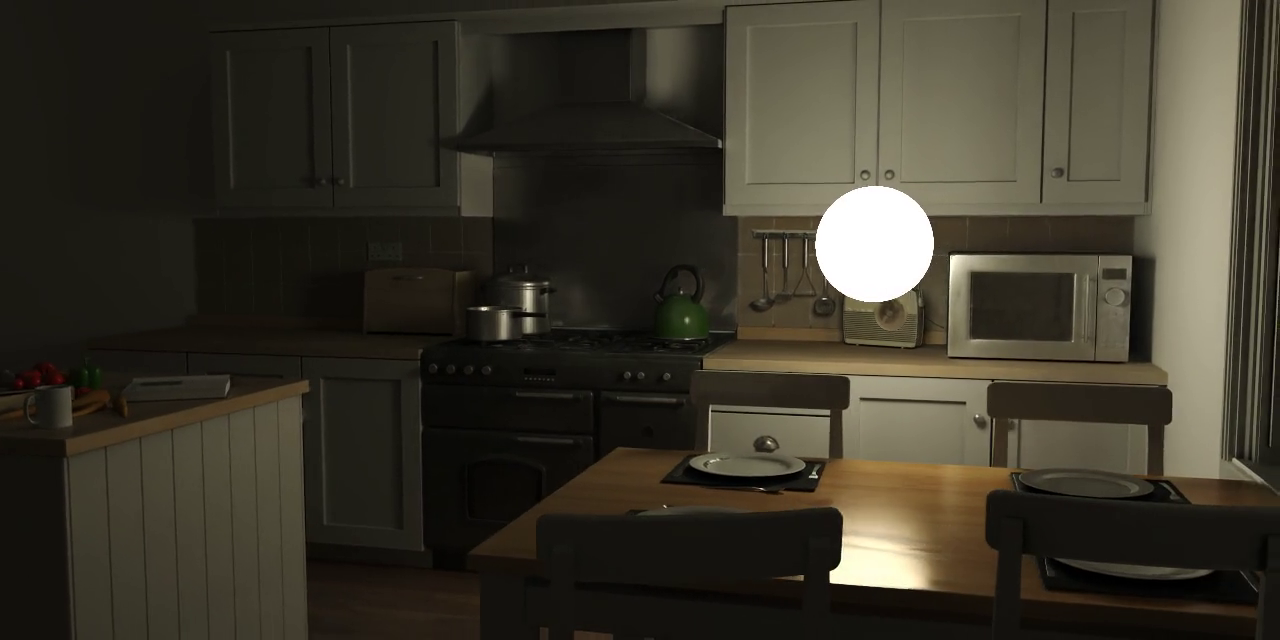}}  \\

        \raisebox{2\normalbaselineskip}[0pt][0pt]{\rotatebox[origin=c]{90}{Insert}} 
        & \frame{\includegraphics[width=0.3\textwidth]{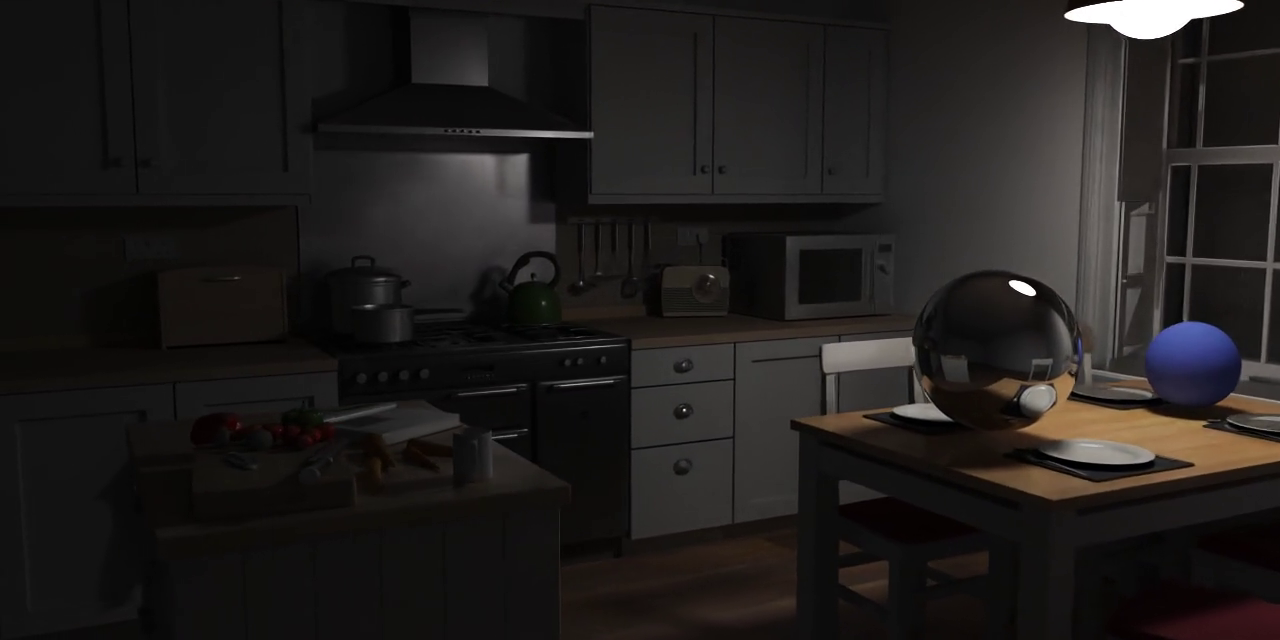}} & \frame{\includegraphics[width=0.3\textwidth]{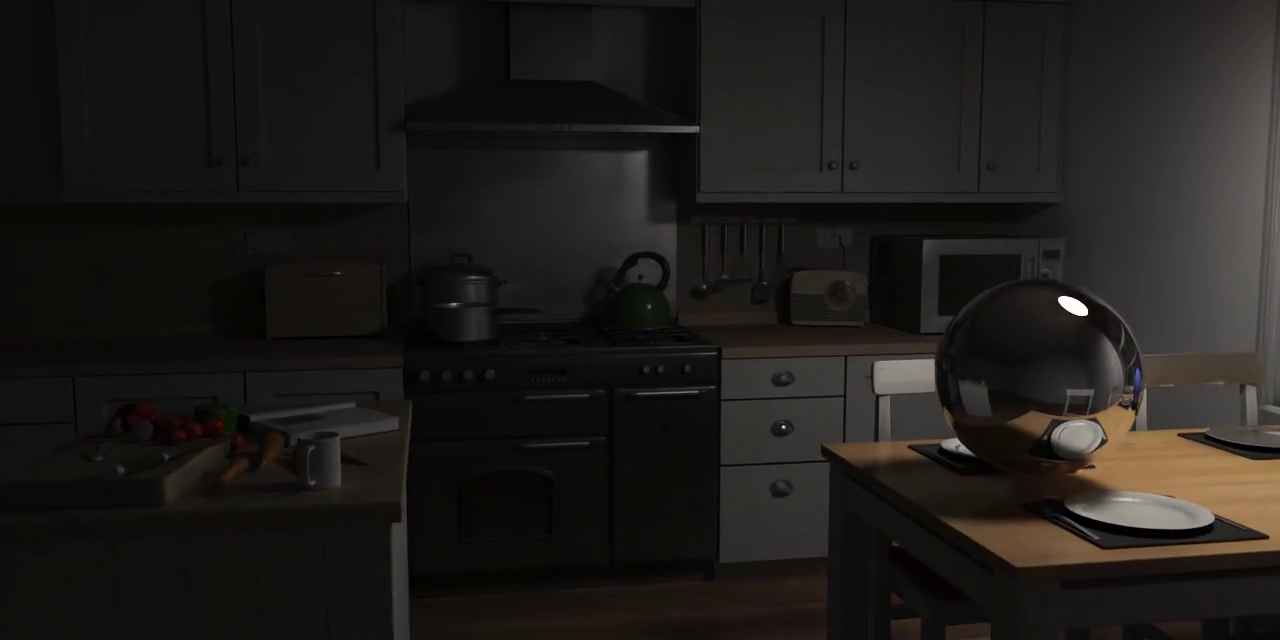}} & \frame{\includegraphics[width=0.3\textwidth]{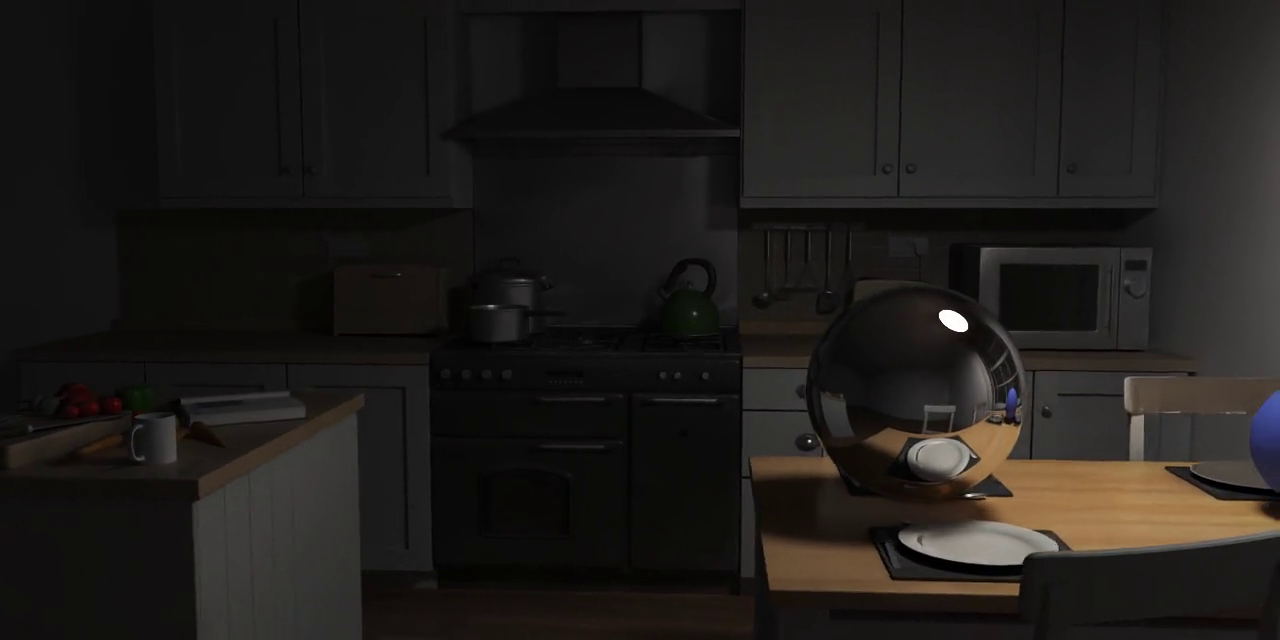}} & \frame{\includegraphics[width=0.3\textwidth]{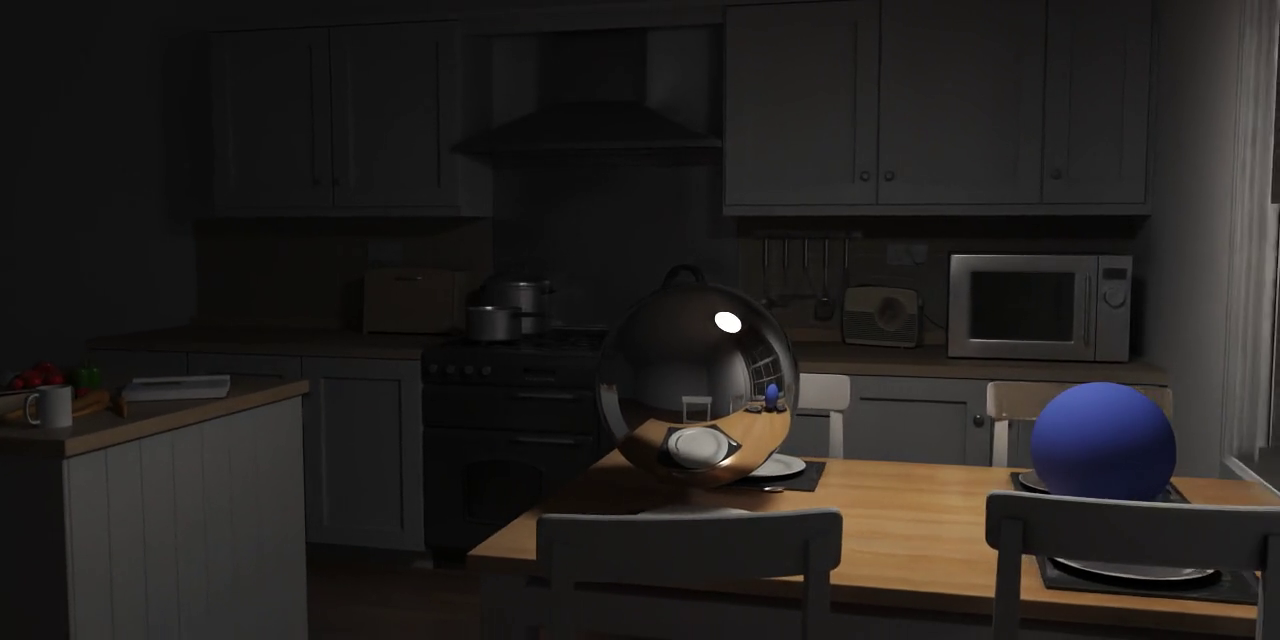}}
    \end{tabular}
    }
    \caption{\textbf{Relighting and object insertion in kitchen.} From top to bottom, we visualize the reconstruction (1st row), relighting (2nd), and object insertion (3rd).}
    \label{tab:supp_relight_kitchen}
\end{figure}

Our method estimates accurate surface material and spatially varying HDR illumination from LDR images, enabling various applications such as relighting and object insertion.
We provide the qualitative results of real-world scenes in \cref{tab:supp_relight_conferenceroom}, \cref{tab:supp_relight_bathroom2}, \cref{tab:supp_relight_room2}, \cref{fig:compare_li_neilfpp}, \cref{tab:supp_relight_kitchen}, where we sample novel camera trajectories and render the scene at different time steps.
The results demonstrate effective modeling of specular reflections on smooth surfaces (like `mirror' and `whiteboard') upon introducing new light sources.
Moreover, our method accurately simulates inter-reflections between the scene and the inserted objects, significantly elevating the realism of object insertion.
To summarize, we show that IRIS can render real-world scenes under various illumination from different viewpoints.
For more interactive visualizations and comparisons, please check our supplementary webpage \href{https://iris-ldr.github.io}{https://iris-ldr.github.io}.

\begin{figure*}[t]
    \centering\setlength{\tabcolsep}{0.1em}
    \resizebox{1.0\textwidth}{!}{%
    \begin{tabular}{@{}lcccc|cccc@{}}

    & NeILF~\cite{yao2022neilf} & FIPT-LDR* & Ours & Ground Truth & Li et al~\cite{li2022physically} & FIPT-LDR* & Ours & Ground Truth   \\[0.2em]
    
    \raisebox{2.0\normalbaselineskip}[0pt][0pt]{\rotatebox[origin=c]{90}{\footnotesize Reconstruction}} & \frame{\includegraphics[width=0.2\textwidth]{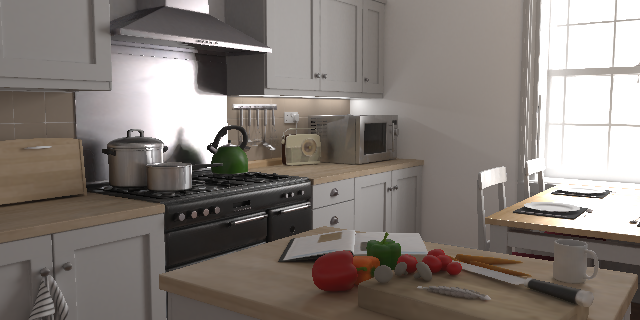}} & \frame{\includegraphics[width=0.2\textwidth]{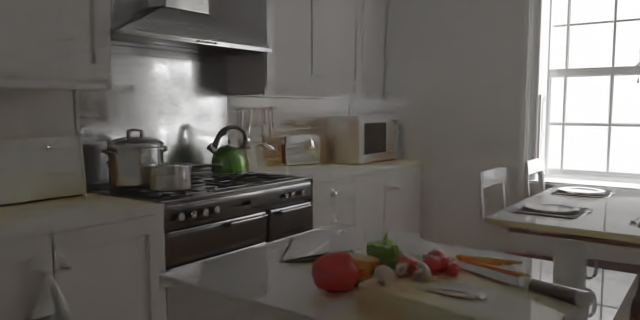}} & \frame{\includegraphics[width=0.2\textwidth]{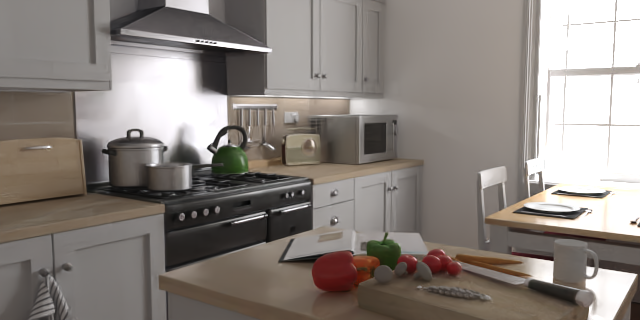}} & \frame{\includegraphics[width=0.2\textwidth]{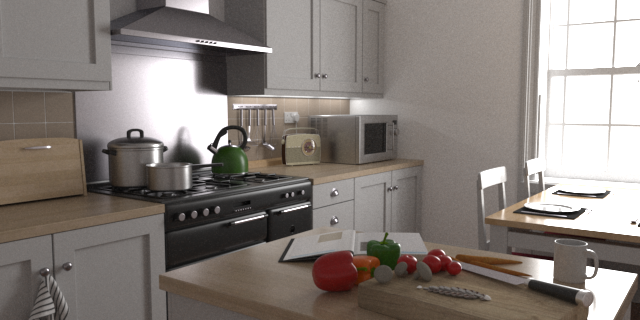}} & \frame{\includegraphics[width=0.2\textwidth]{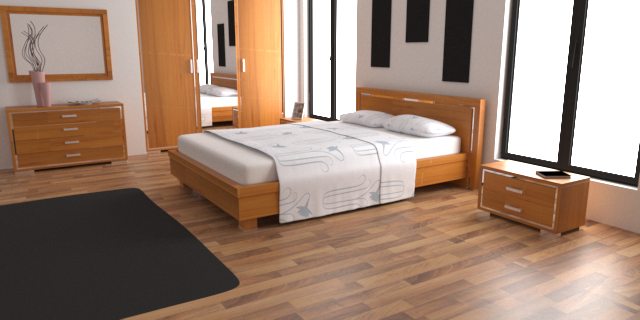}} & \frame{\includegraphics[width=0.2\textwidth]{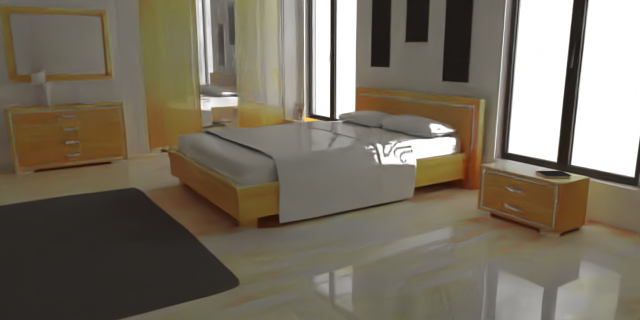}} & \frame{\includegraphics[width=0.2\textwidth]{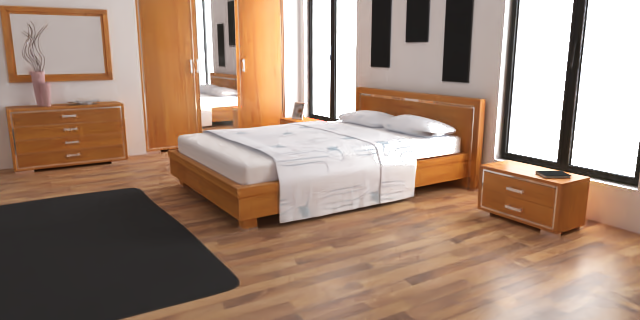}} & \frame{\includegraphics[width=0.2\textwidth]{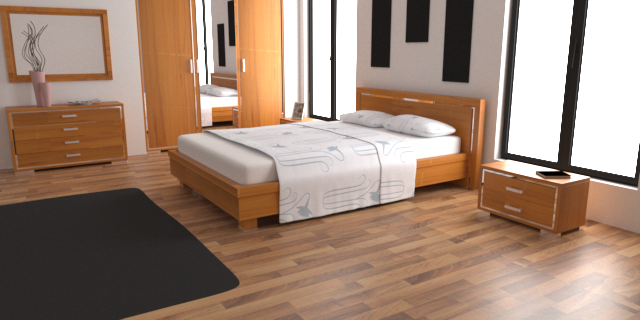}} \\
    
    \raisebox{2.0\normalbaselineskip}[0pt][0pt]{\rotatebox[origin=c]{90}{\footnotesize Material $\ba'$}} & 
    \frame{\includegraphics[width=0.2\textwidth]{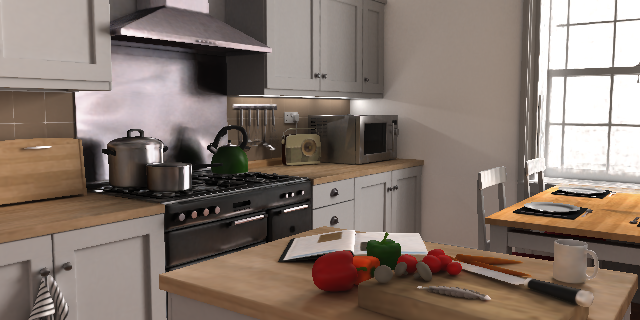}} & \frame{\includegraphics[width=0.2\textwidth]{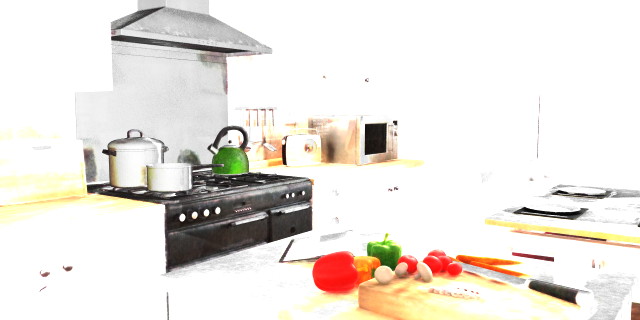}} & \frame{\includegraphics[width=0.2\textwidth]{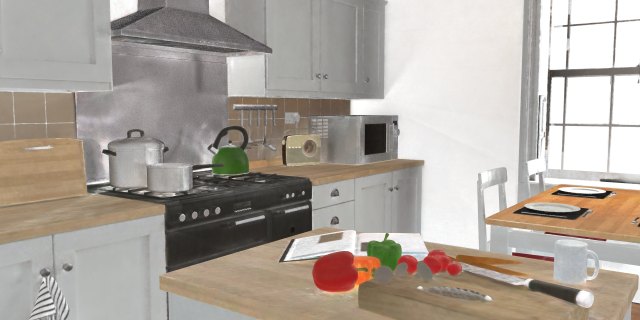}} & \frame{\includegraphics[width=0.2\textwidth]{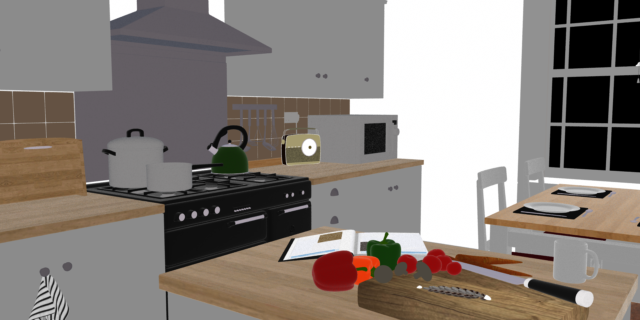}} & \frame{\includegraphics[width=0.2\textwidth]{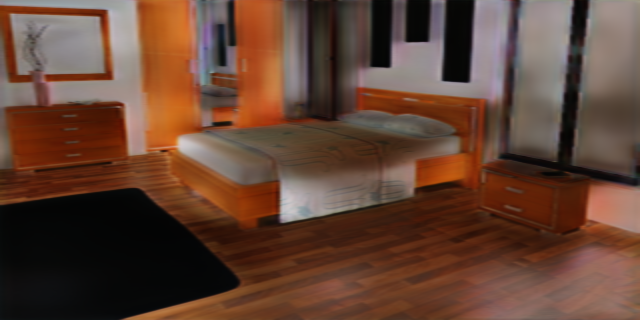}} & \frame{\includegraphics[width=0.2\textwidth]{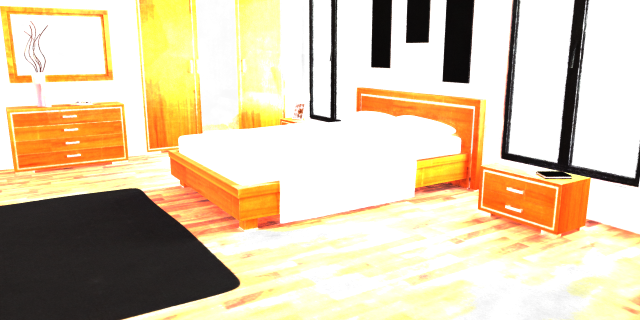}} & \frame{\includegraphics[width=0.2\textwidth]{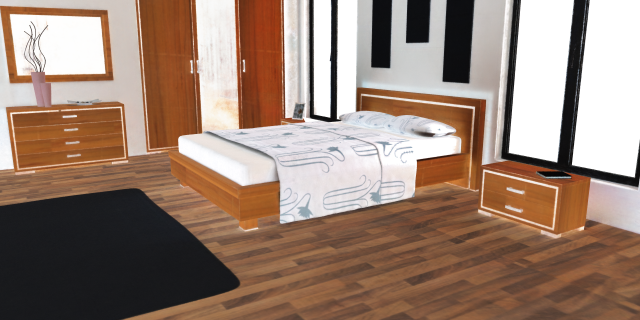}} & \frame{\includegraphics[width=0.2\textwidth]{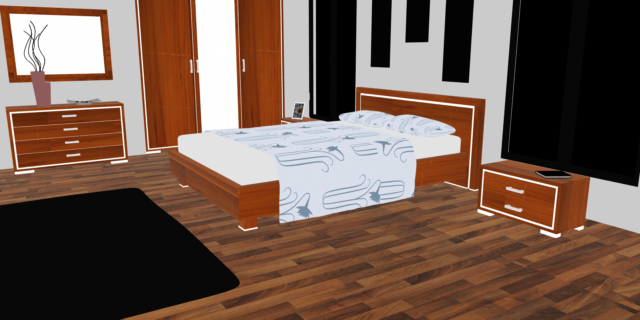}} \\
    
    \raisebox{2.0\normalbaselineskip}[0pt][0pt]{\rotatebox[origin=c]{90}{\footnotesize Roughness $\sigma$ }}& 
    \frame{\includegraphics[width=0.2\textwidth]{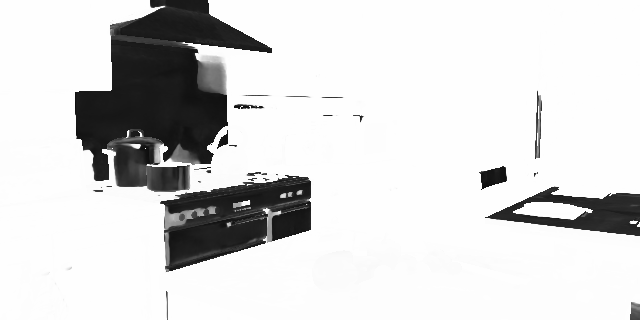}} & \frame{\includegraphics[width=0.2\textwidth]{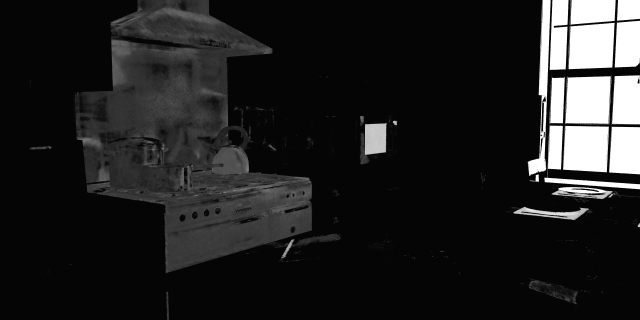}} & \frame{\includegraphics[width=0.2\textwidth]{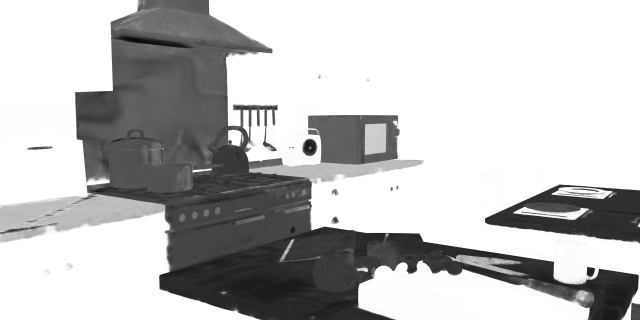}} & \frame{\includegraphics[width=0.2\textwidth]{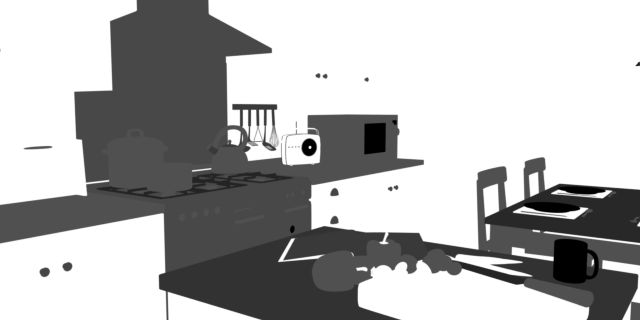}} & \frame{\includegraphics[width=0.2\textwidth]{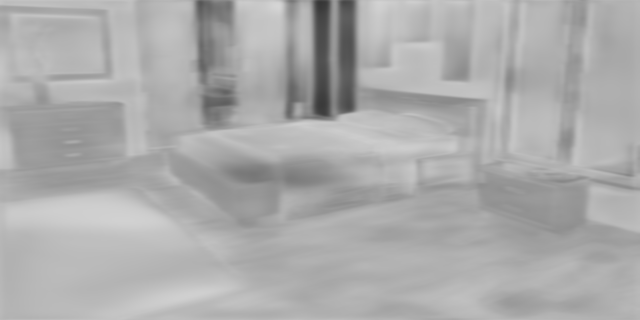}} & \frame{\includegraphics[width=0.2\textwidth]{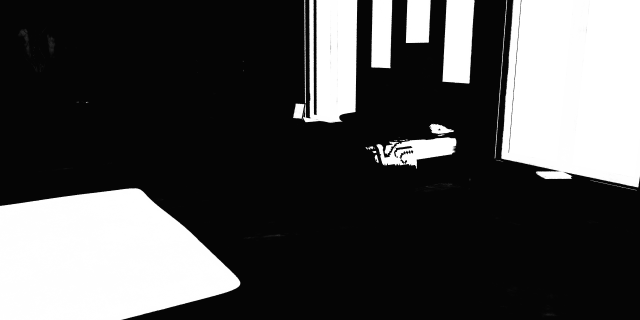}} & \frame{\includegraphics[width=0.2\textwidth]{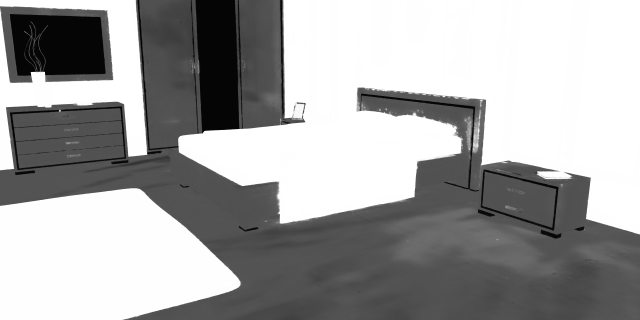}} & \frame{\includegraphics[width=0.2\textwidth]{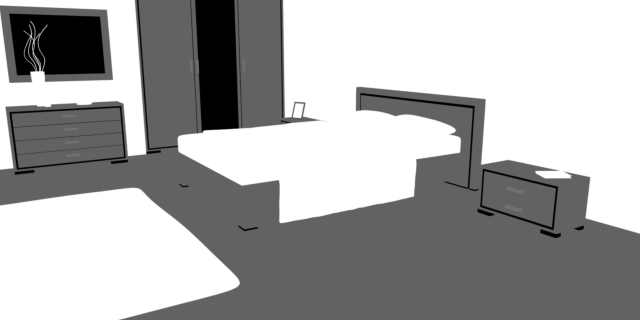}} \\

     \raisebox{2.0\normalbaselineskip}[0pt][0pt]{\rotatebox[origin=c]{90}{\footnotesize Emission map}}& 
     \frame{\includegraphics[width=0.2\textwidth]{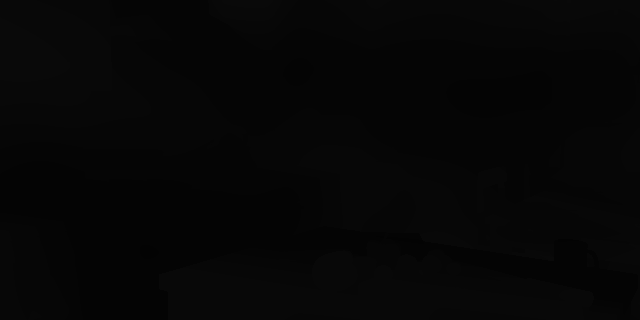}} & \frame{\includegraphics[width=0.2\textwidth]{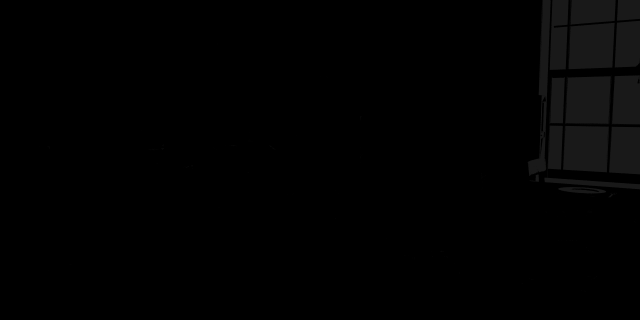}} & \frame{\includegraphics[width=0.2\textwidth]{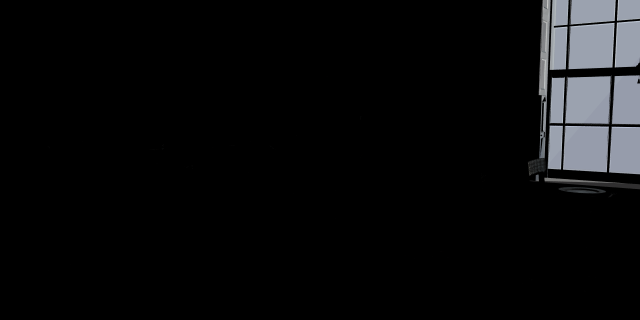}} & \frame{\includegraphics[width=0.2\textwidth]{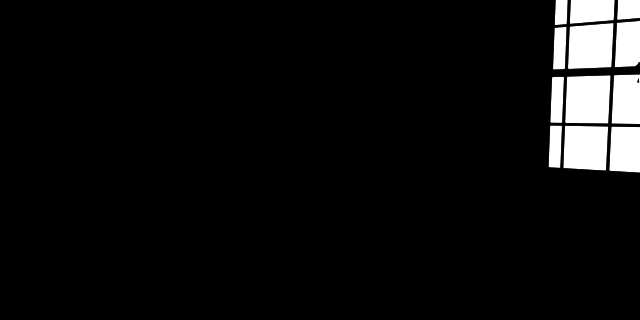}} & \frame{\includegraphics[width=0.2\textwidth]{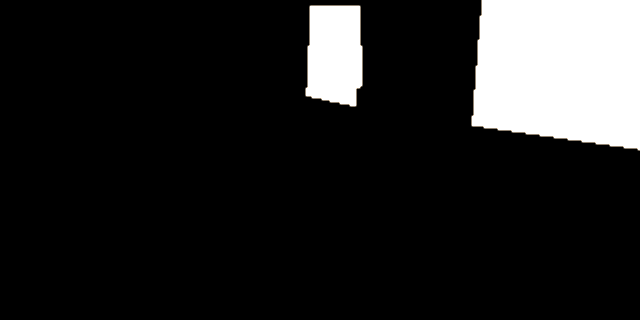}} & \frame{\includegraphics[width=0.2\textwidth]{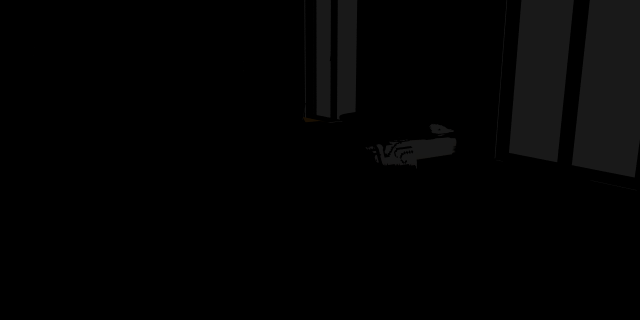}} & \frame{\includegraphics[width=0.2\textwidth]{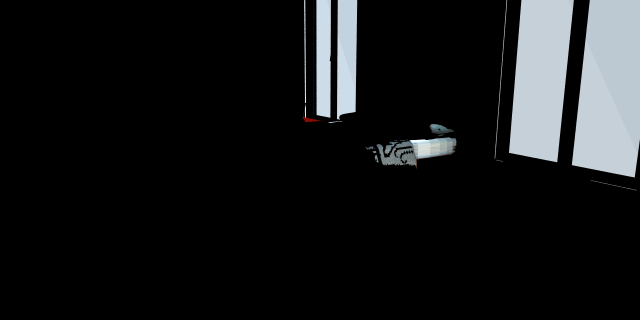}} & \frame{\includegraphics[width=0.2\textwidth]{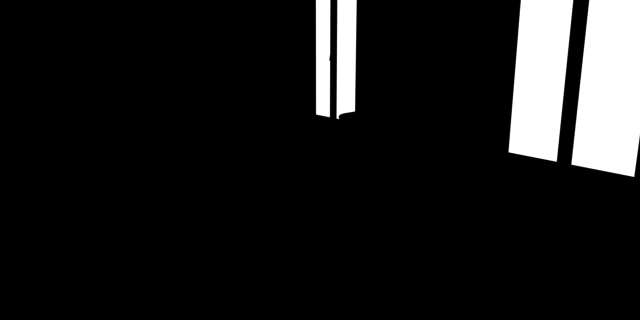}} \\

     & & & & & & & & \\

     \raisebox{2.0\normalbaselineskip}[0pt][0pt]{\rotatebox[origin=c]{90}{\footnotesize Reconstruction}} & \frame{\includegraphics[width=0.2\textwidth]{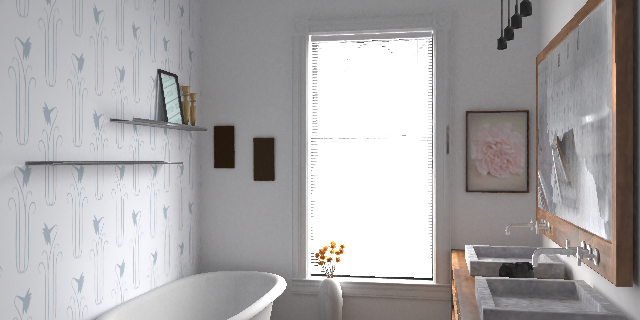}} & \frame{\includegraphics[width=0.2\textwidth]{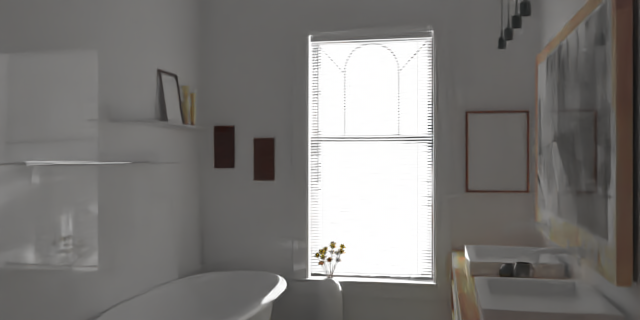}} & \frame{\includegraphics[width=0.2\textwidth]{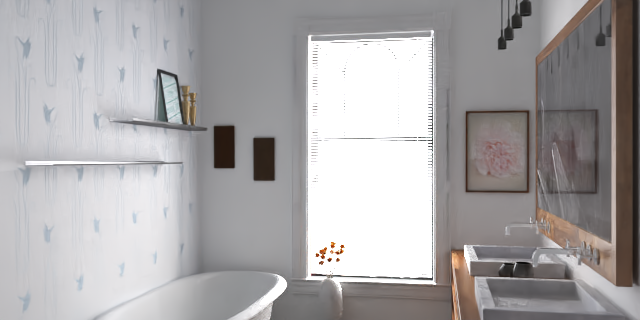}} & \frame{\includegraphics[width=0.2\textwidth]{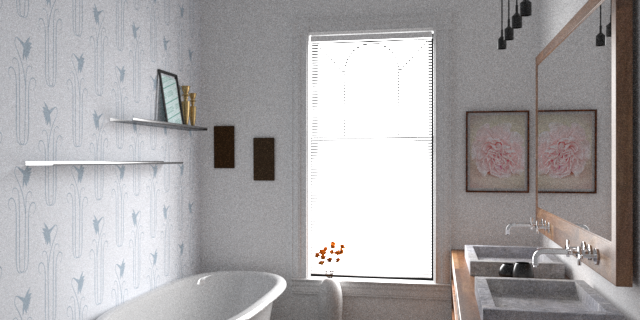}} & \frame{\includegraphics[width=0.2\textwidth]{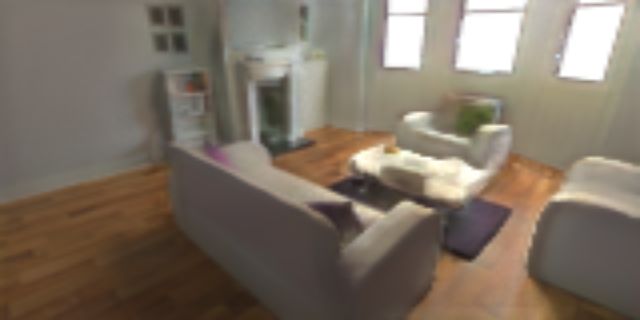}} & \frame{\includegraphics[width=0.2\textwidth]{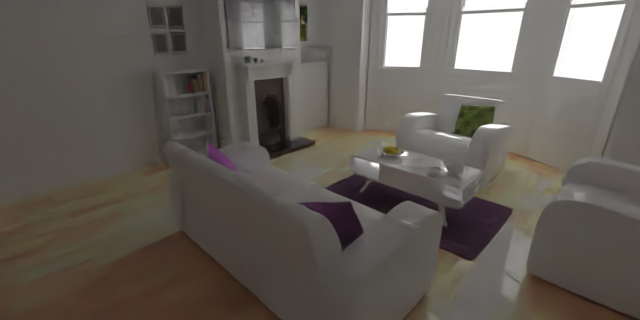}} & \frame{\includegraphics[width=0.2\textwidth]{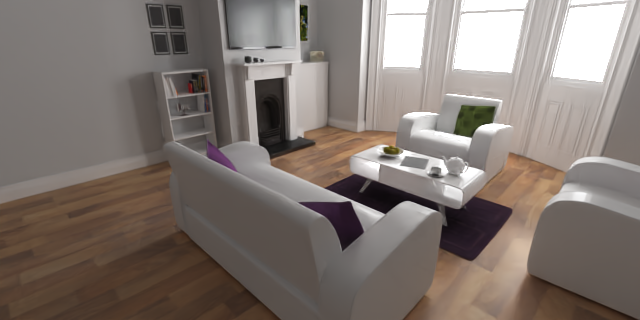}} & \frame{\includegraphics[width=0.2\textwidth]{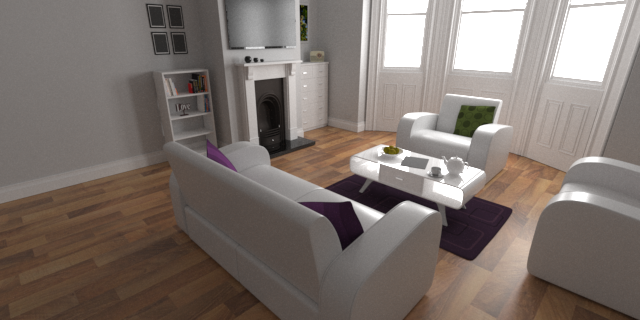}} \\
    
    \raisebox{2.0\normalbaselineskip}[0pt][0pt]{\rotatebox[origin=c]{90}{\footnotesize Material $\ba'$}} & 
    \frame{\includegraphics[width=0.2\textwidth]{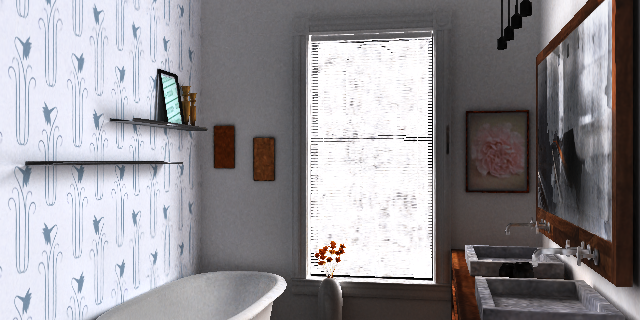}} & \frame{\includegraphics[width=0.2\textwidth]{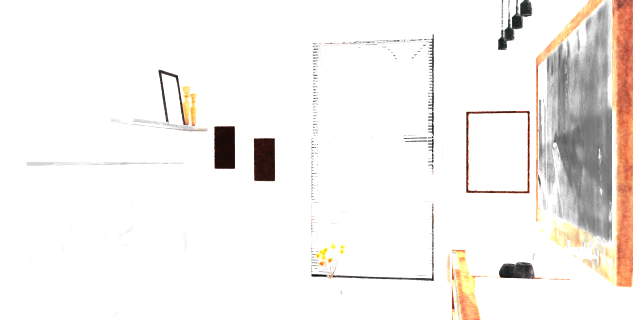}} & \frame{\includegraphics[width=0.2\textwidth]{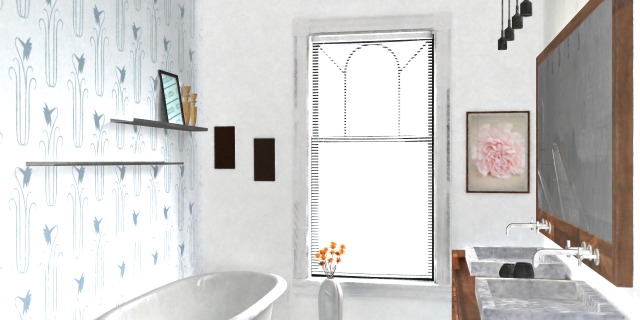}} & \frame{\includegraphics[width=0.2\textwidth]{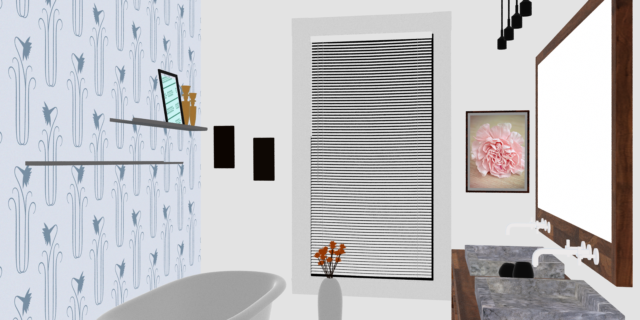}} & \frame{\includegraphics[width=0.2\textwidth]{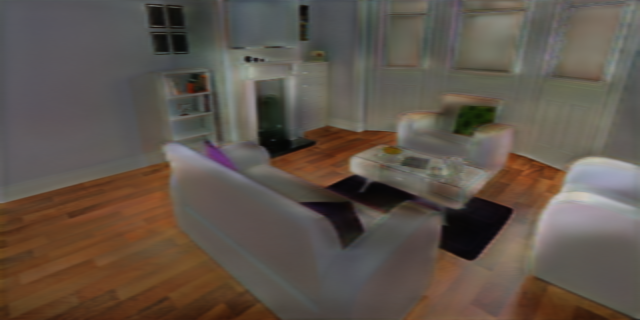}} & \frame{\includegraphics[width=0.2\textwidth]{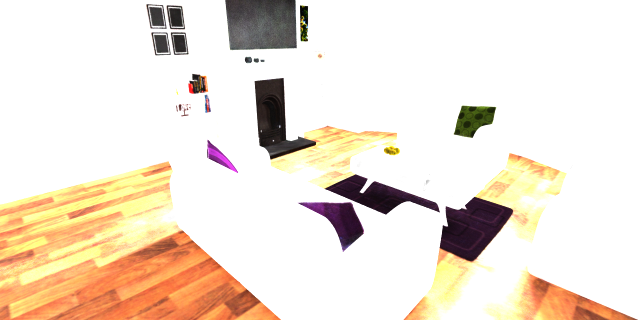}} & \frame{\includegraphics[width=0.2\textwidth]{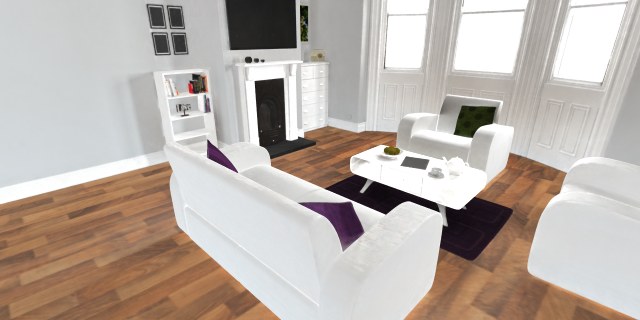}} & \frame{\includegraphics[width=0.2\textwidth]{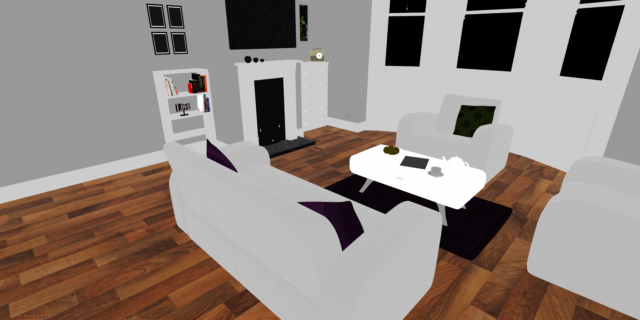}} \\
    
    \raisebox{2.0\normalbaselineskip}[0pt][0pt]{\rotatebox[origin=c]{90}{\footnotesize Roughness $\sigma$ }}& 
    \frame{\includegraphics[width=0.2\textwidth]{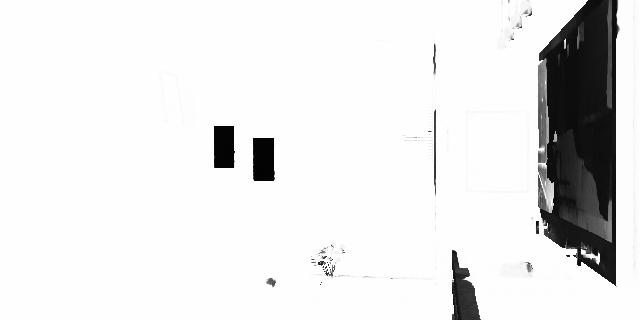}} & \frame{\includegraphics[width=0.2\textwidth]{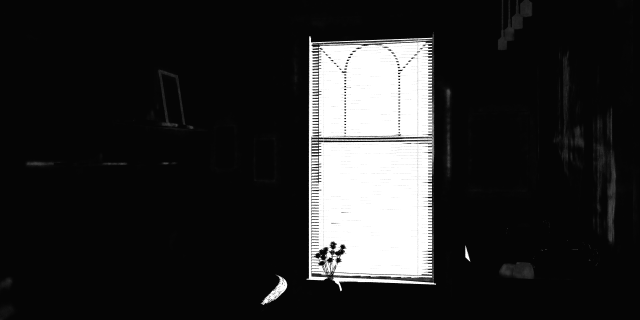}} & \frame{\includegraphics[width=0.2\textwidth]{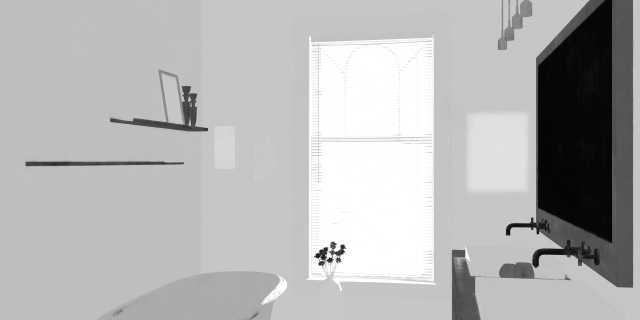}} & \frame{\includegraphics[width=0.2\textwidth]{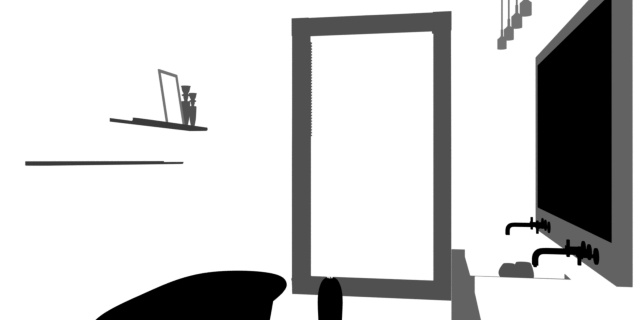}} & \frame{\includegraphics[width=0.2\textwidth]{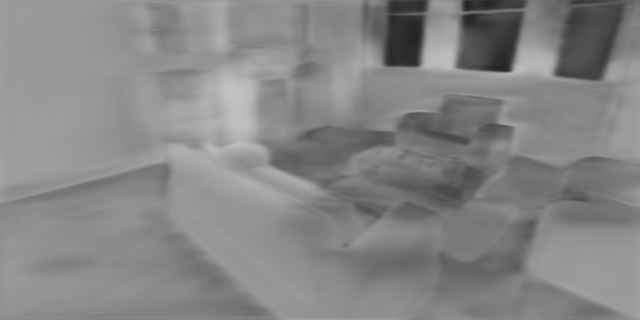}} & \frame{\includegraphics[width=0.2\textwidth]{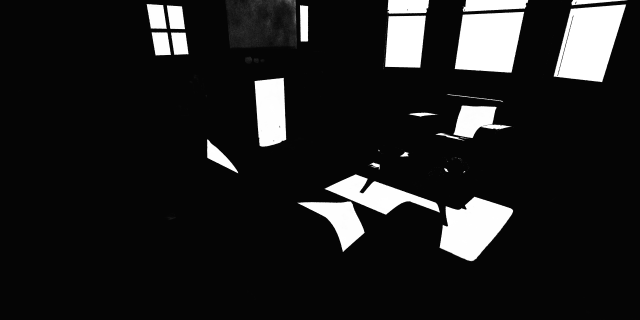}} & \frame{\includegraphics[width=0.2\textwidth]{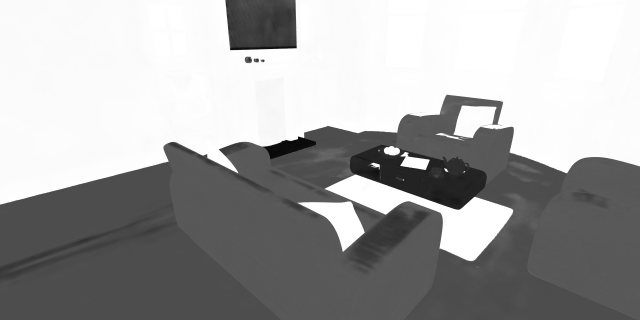}} & \frame{\includegraphics[width=0.2\textwidth]{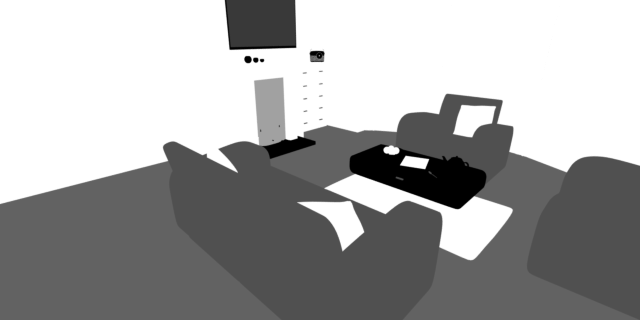}} \\

     \raisebox{2.0\normalbaselineskip}[0pt][0pt]{\rotatebox[origin=c]{90}{\footnotesize Emission map}}& 
     \frame{\includegraphics[width=0.2\textwidth]{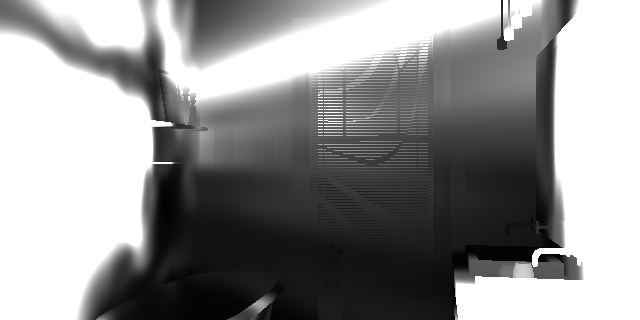}} & \frame{\includegraphics[width=0.2\textwidth]{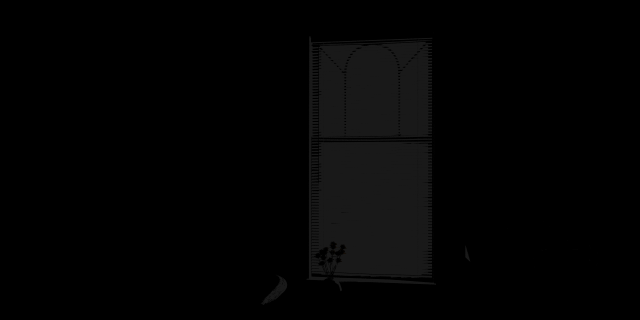}} & \frame{\includegraphics[width=0.2\textwidth]{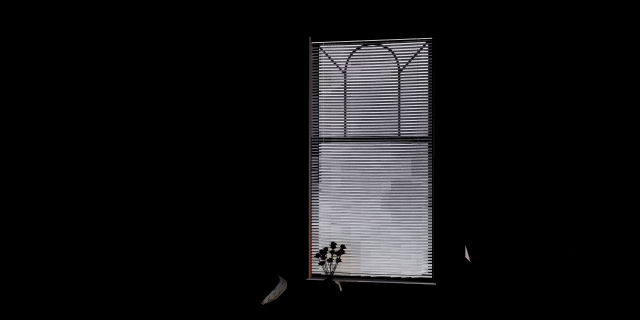}} & \frame{\includegraphics[width=0.2\textwidth]{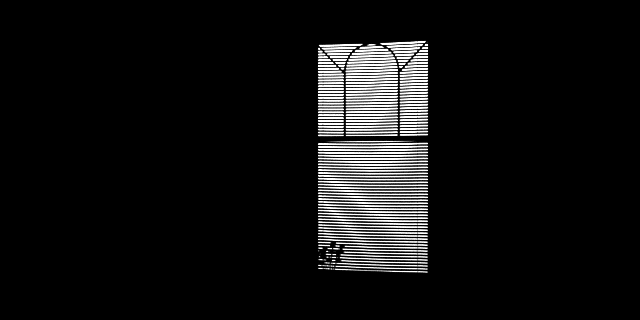}} & \frame{\includegraphics[width=0.2\textwidth]{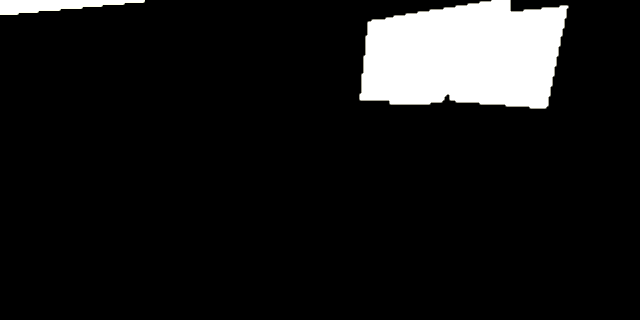}} & \frame{\includegraphics[width=0.2\textwidth]{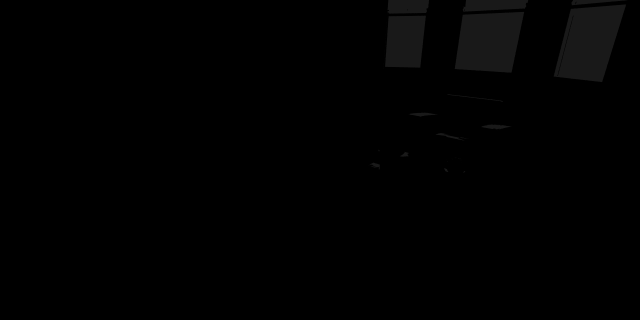}} & \frame{\includegraphics[width=0.2\textwidth]{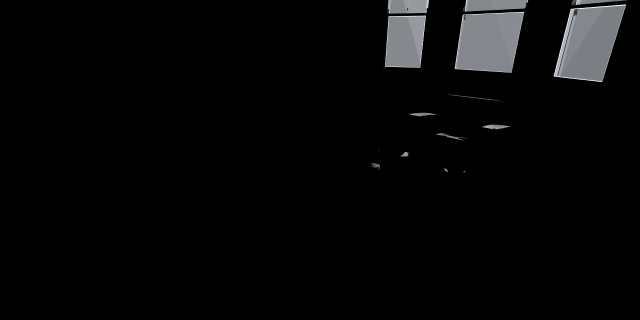}} & \frame{\includegraphics[width=0.2\textwidth]{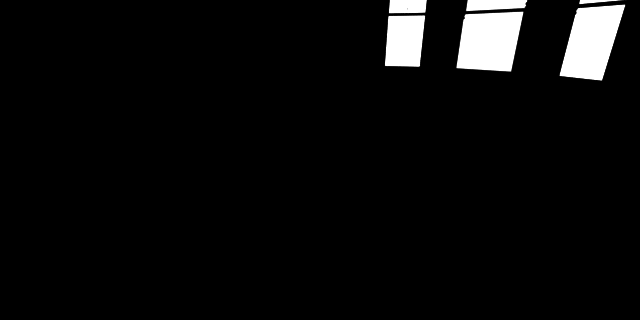}} \\

    \end{tabular}%
    }\vspace{-2mm}
    \caption{\label{fig:syn_intrinsic}%
        \textbf{Intrinsic decomposition of synthetic scenes \cite{wu2023factorized}.}
        From top to bottom, we show reconstruction, material reflectance $\ba'$, roughness $\sigma$, and emission maps.        
        For the emission map, we show normalized HDR emission, such that it is not saturated and differences become visible. With LDR images as input, IRIS successfully recovers the HDR lighting and accurate surface material. 
    }
    \vspace{-1mm}
\end{figure*}

\begin{figure*}[t]
    \centering\setlength{\tabcolsep}{0.1em}
    \resizebox{1.0\textwidth}{!}{%
    \begin{tabular}{@{}lcccc|cccc@{}}

    & NeILF~\cite{yao2022neilf} & FIPT-LDR* & Ours & Ground Truth & Li et al. \cite{li2022physically} & FIPT-LDR* & Ours & Ground Truth \\[0.2em]

    \raisebox{2.0\normalbaselineskip}[0pt][0pt]{\rotatebox[origin=c]{90}{\footnotesize Bathroom}}& 
    \frame{\includegraphics[width=0.2\textwidth]{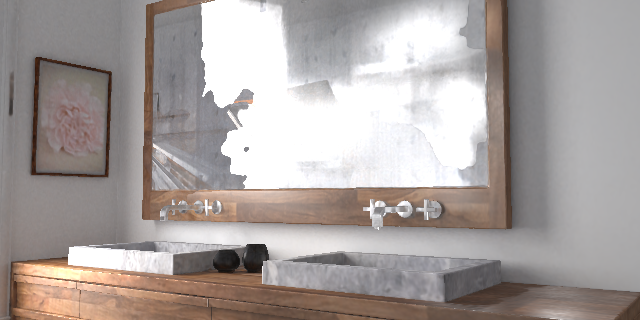}} & \frame{\includegraphics[width=0.2\textwidth]{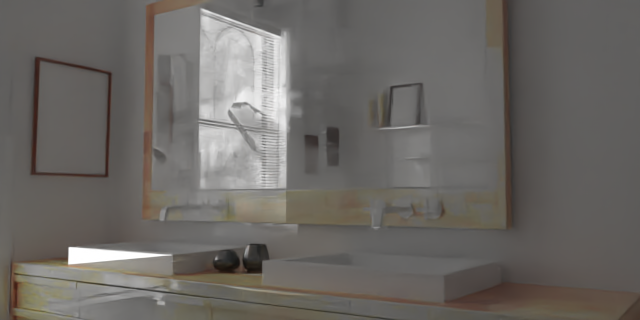}} & \frame{\includegraphics[width=0.2\textwidth]{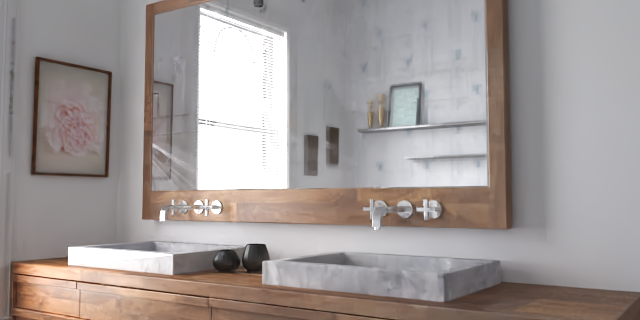}} & \frame{\includegraphics[width=0.2\textwidth]{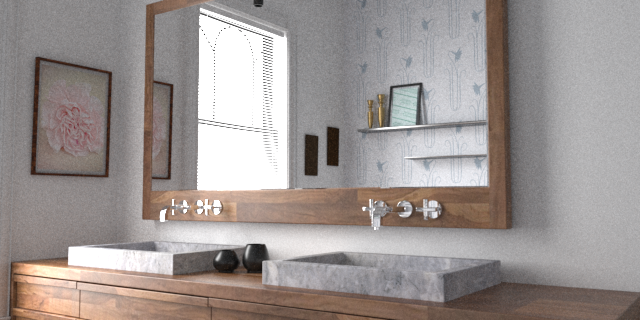}} & 
    
    \frame{\includegraphics[width=0.2\textwidth]{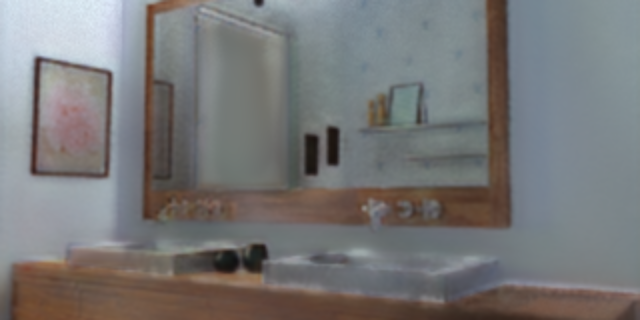}} & \frame{\includegraphics[width=0.2\textwidth]{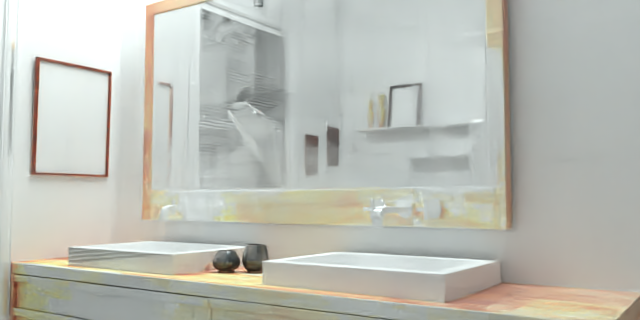}} & \frame{\includegraphics[width=0.2\textwidth]{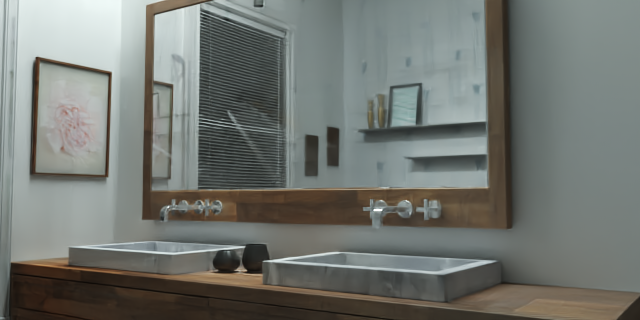}} & \frame{\includegraphics[width=0.2\textwidth]{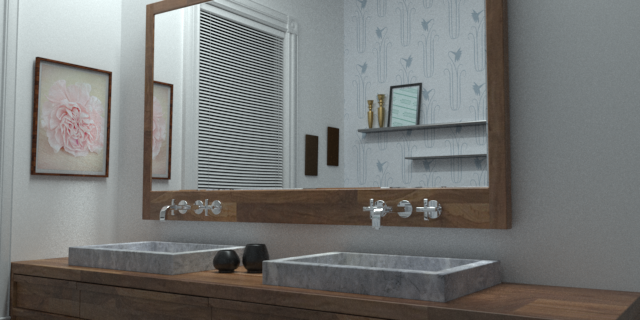}} \\
    
    \raisebox{2.0\normalbaselineskip}[0pt][0pt]{\rotatebox[origin=c]{90}{\footnotesize Bedroom}} &
    \frame{\includegraphics[width=0.2\textwidth]{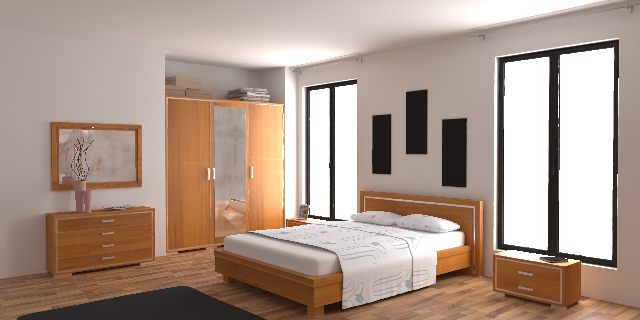}} & \frame{\includegraphics[width=0.2\textwidth]{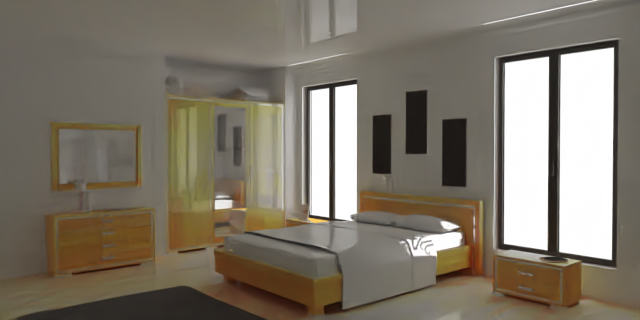}} & \frame{\includegraphics[width=0.2\textwidth]{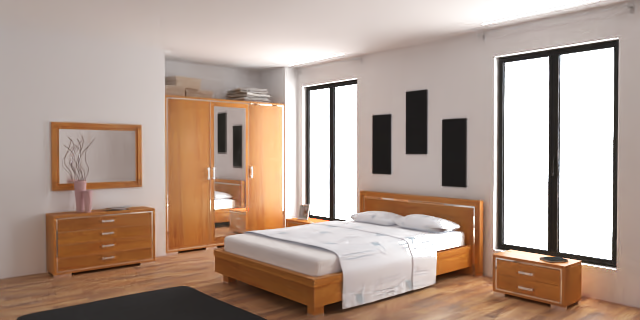}} & \frame{\includegraphics[width=0.2\textwidth]{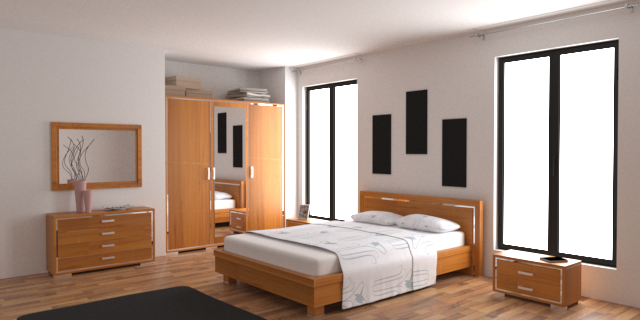}} & 
    
    \frame{\includegraphics[width=0.2\textwidth]{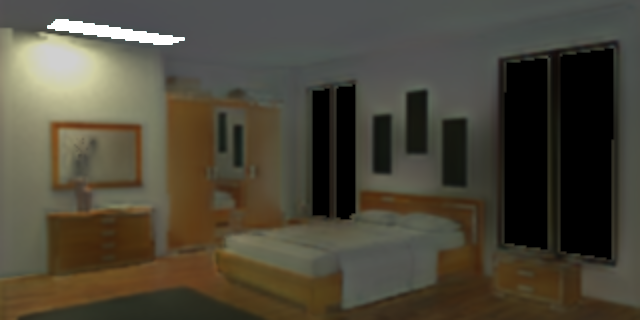}} & \frame{\includegraphics[width=0.2\textwidth]{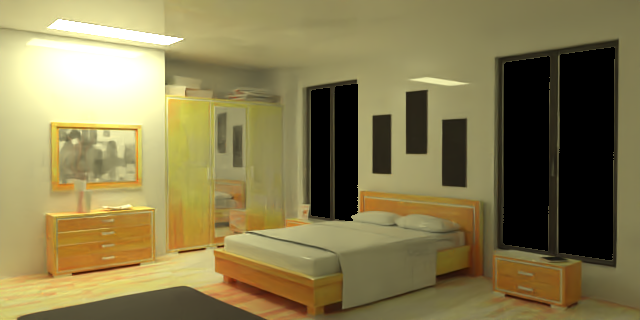}} & \frame{\includegraphics[width=0.2\textwidth]{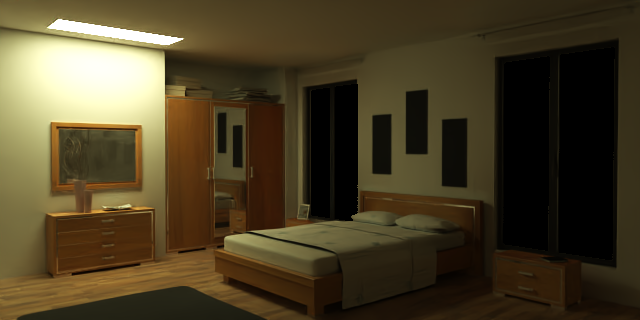}} & \frame{\includegraphics[width=0.2\textwidth]{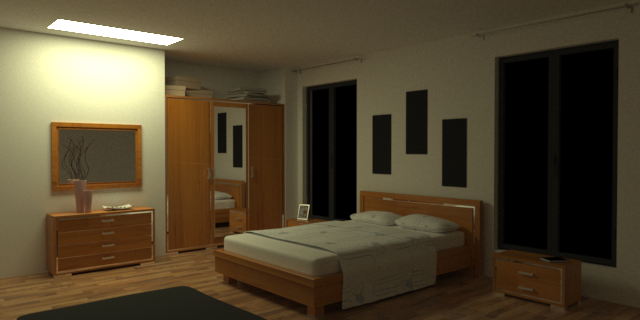} }\\
    
    \raisebox{2.0\normalbaselineskip}[0pt][0pt]{\rotatebox[origin=c]{90}{\footnotesize Kitchen}}& 
    \frame{\includegraphics[width=0.2\textwidth]{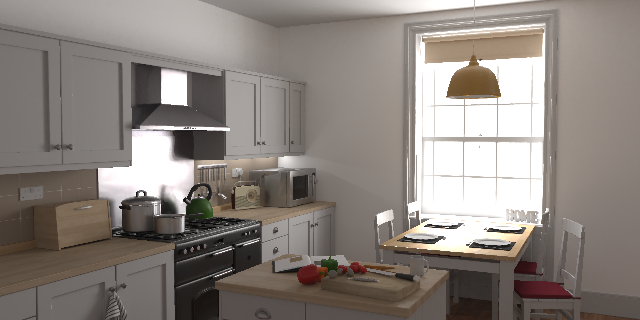}} & \frame{\includegraphics[width=0.2\textwidth]{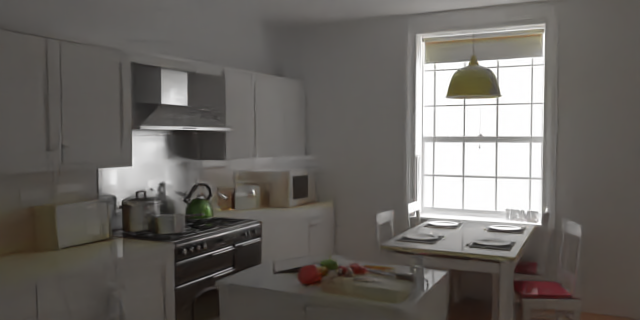}} & \frame{\includegraphics[width=0.2\textwidth]{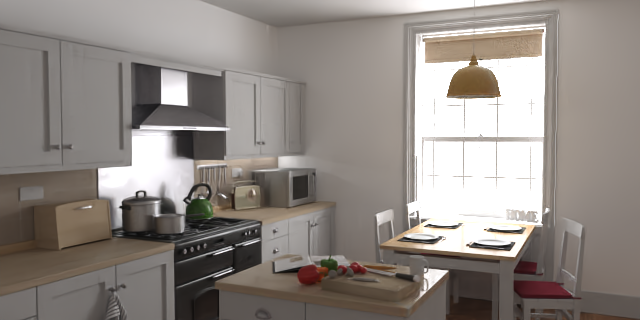}} & \frame{\includegraphics[width=0.2\textwidth]{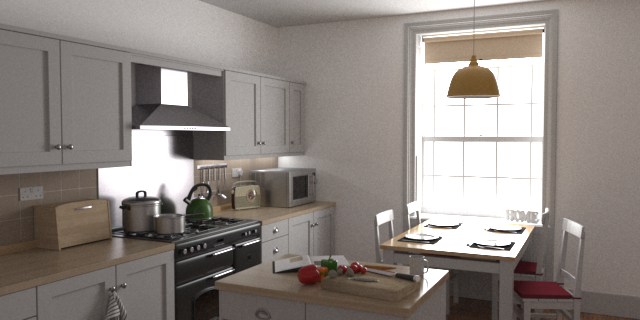}} & 
    
    \frame{\includegraphics[width=0.2\textwidth]{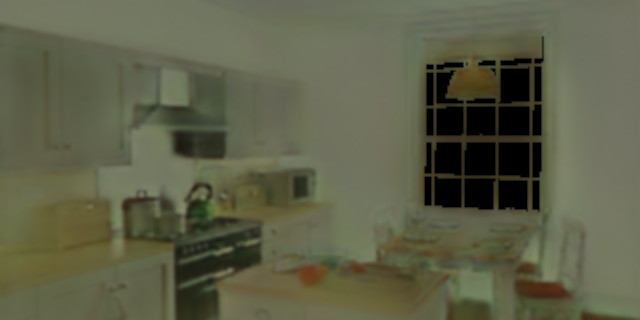}} & \frame{\includegraphics[width=0.2\textwidth]{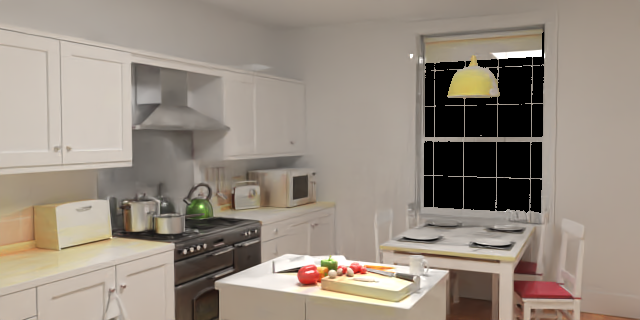}} & \frame{\includegraphics[width=0.2\textwidth]{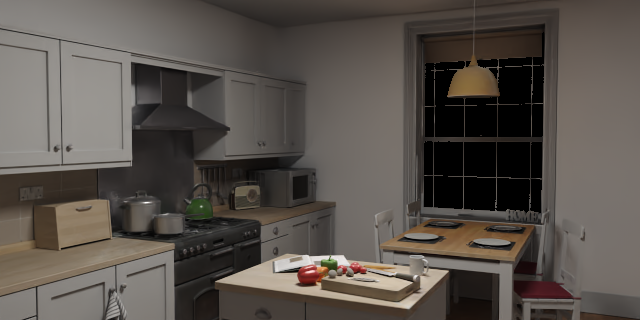}} & \frame{\includegraphics[width=0.2\textwidth]{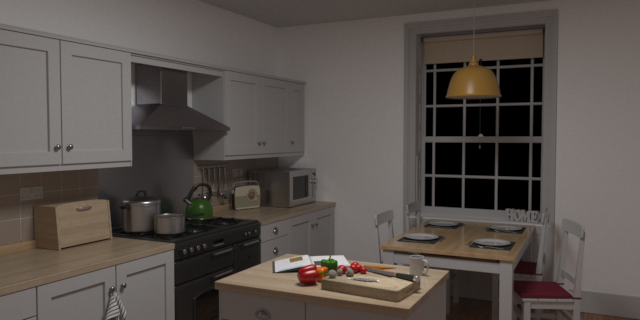}} \\

    \raisebox{2.0\normalbaselineskip}[0pt][0pt]{\rotatebox[origin=c]{90}{\footnotesize Livingroom}}& 
    \frame{\includegraphics[width=0.2\textwidth]{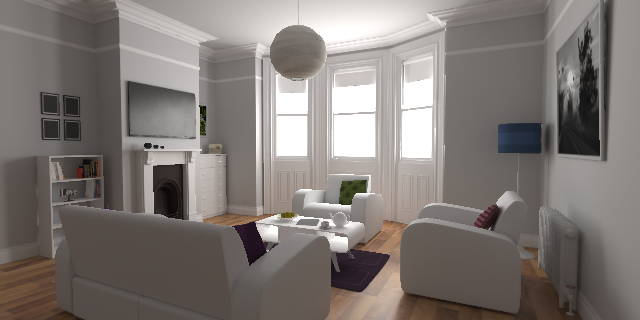}} & \frame{\includegraphics[width=0.2\textwidth]{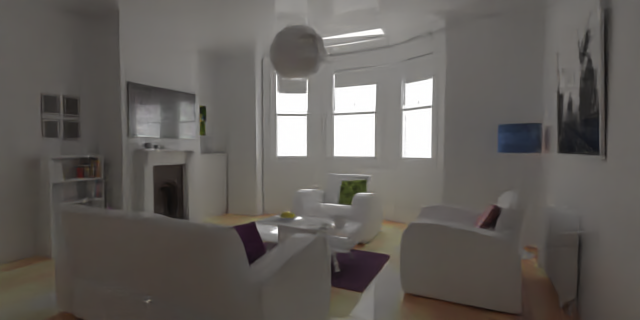}} & \frame{\includegraphics[width=0.2\textwidth]{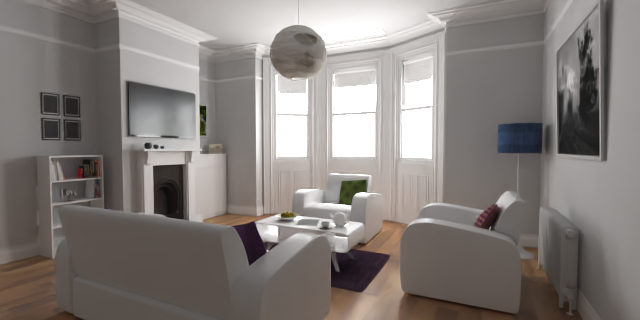}} & \frame{\includegraphics[width=0.2\textwidth]{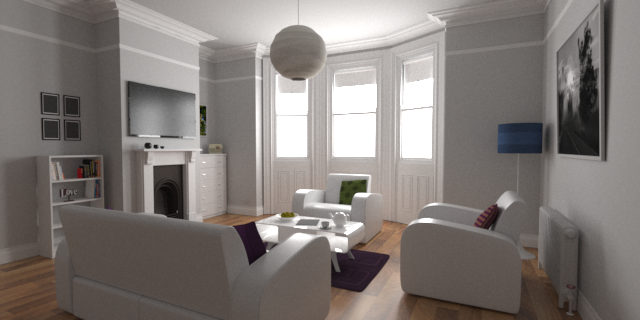} }& 
    
    \frame{\includegraphics[width=0.2\textwidth]{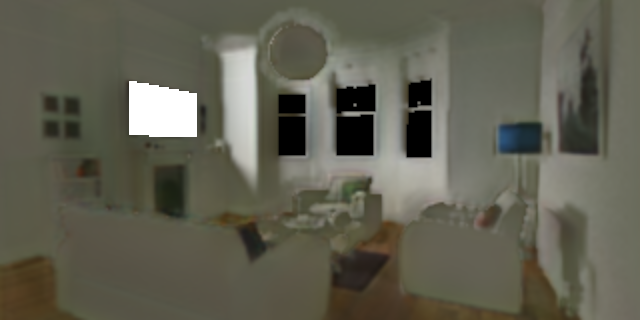}} & \frame{\includegraphics[width=0.2\textwidth]{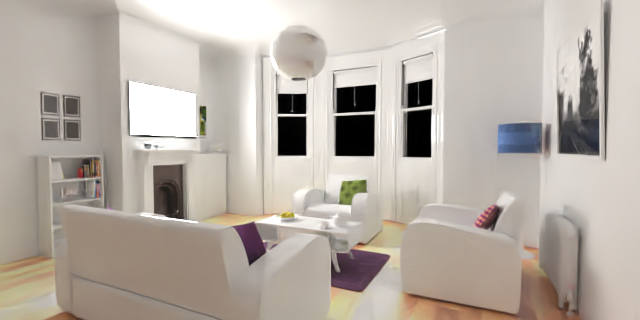}} & \frame{\includegraphics[width=0.2\textwidth]{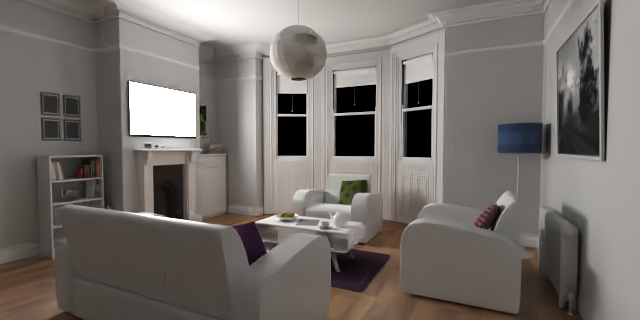}} & \frame{\includegraphics[width=0.2\textwidth]{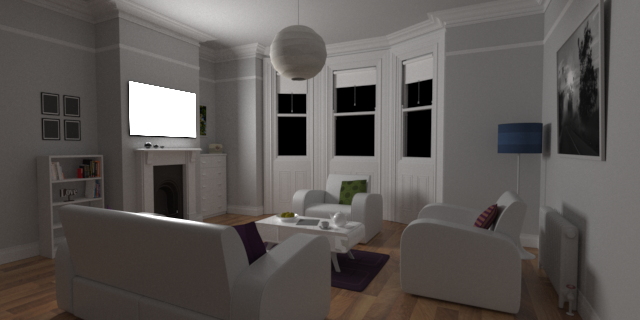}} \\
    \end{tabular}%
    }\vspace{-1mm}
    \caption{\label{fig:syn_nvs_relight}%
        \textbf{Novel-view synthesis and relighting results on synthetic scenes \cite{wu2023factorized}.}
        The novel view synthesis results are shown in the left four columns, and the relighting of the same novel view are shown in the right four columns. 
    }
\end{figure*}

\section{Additional Evaluation Results}
\begin{table}
    \caption{\label{tab:syn_intrinsic_full}%
        \textbf{BRDF-emission comparison on synthetic data.}
        FIPT-LDR* is provided with the GT emitter mask as additional input.
        The best metrics among LDR methods are highlighted in bold.
    }
    \centering\setlength{\tabcolsep}{6pt}
    \resizebox{1.0\linewidth}{!}{%
    \begin{tabular}{llccccc}
    \toprule
        & & $\bk_\text{d}$ & $\ba'$ & $\sigma$ & \multicolumn{2}{c}{$\bL_\text{e}$} \\
        & Method &\multicolumn{3}{c}{PSNR $\uparrow$} & IoU $\uparrow$ & L2 $\downarrow$\\
    \midrule
       \multirow{5}{*}{Kitchen}& Li et al~\cite{li2022physically}& 15.75& 12.64& 10.15& 0.43& 1.410 \\
       & NeILF~\cite{yao2022neilf}& 16.63& 13.73& 14.77& --- & --- \\
       & FIPT-LDR*& 15.77& \padd 8.97& \padd 5.94& \textbf{0.58} & 0.450\\
       & Ours & \textbf{23.22}& \textbf{17.52}& \textbf{20.35}& \textbf{0.58} & \textbf{0.203}\\
       \cmidrule{2-7}
       & FIPT-HDR~\cite{wu2023factorized}& 34.34& 27.05& 24.55&  0.88& 0.010\\
    \midrule
    \multirow{5}{*}{Bedroom}& Li et al~\cite{li2022physically} &  18.90& 15.10& 11.38& 0.34& 2.784 \\
       & NeILF~\cite{yao2022neilf}& 16.85& 13.99& 16.03& --- & --- \\
       & FIPT-LDR*& 18.38& \padd 9.60& \padd 5.82&  \textbf{0.77}& 0.245\\
       & Ours &  \textbf{26.44}&  \textbf{20.95}& \textbf{26.47}& \textbf{0.77}& \textbf{0.043}\\
       \cmidrule{2-7}
       & FIPT-HDR~\cite{wu2023factorized}& 28.98 & 25.86 & 23.53 & 0.92 & 0.004\\
    \midrule
    \multirow{5}{*}{Livingroom}& Li et al~\cite{li2022physically} & 16.78& 14.71& 11.42& 0.17& 3.610 \\
       & NeILF~\cite{yao2022neilf}& 16.06& 13.86& 15.95& --- & --- \\
       & FIPT-LDR*& 11.59& \padd 8.93& \padd 4.08& \textbf{0.77}& 0.240\\
       & Ours & \textbf{18.09}& \textbf{15.45}& \textbf{25.28}& \textbf{0.77}& \textbf{0.103}\\
       \cmidrule{2-7}
       & FIPT-HDR~\cite{wu2023factorized}& 28.42&  27.47& 30.44& 0.95& 0.005\\
    \midrule
    \multirow{5}{*}{Bathroom}& Li et al~\cite{li2022physically} & 15.50& 13.60& 12.24& 0.45& 1.351 \\
       & NeILF~\cite{yao2022neilf}& 17.85& 14.49& \textbf{21.09}& --- & ---\\
       & FIPT-LDR* & 16.21& 11.46& \padd 4.12& \textbf{0.62}& 0.187\\
       & Ours & \textbf{21.56} &  \textbf{17.74} & 13.43& \textbf{0.62} & \textbf{0.135}\\
       \cmidrule{2-7}
       & FIPT-HDR~\cite{wu2023factorized}& 28.06& 23.54& 26.97& 0.68& 0.080\\
    \bottomrule
    \end{tabular}%
    }
\end{table}

\begin{table*}[t]
	\caption{\label{tab:nvs_synthetic_full}%
		\textbf{Complete quantitative results of novel view synthesis and relighting on synthetic scenes}
	}
    \centering
    \resizebox{1.0\textwidth}{!}{%
    \begin{tabular}{llc@{~~}c@{~~}cc@{~~}c@{~~}cc@{~~}c@{~~}cc@{~~}c@{~~}c}
    \toprule
    &  & \multicolumn{3}{c}{Kitchen} & \multicolumn{3}{c}{Bedroom} & \multicolumn{3}{c}{Livingroom} & \multicolumn{3}{c}{Bathroom} \\
    & Method & PSNR $\uparrow$ & SSIM $\uparrow$ & LPIPS $\downarrow$ & PSNR $\uparrow$ & SSIM $\uparrow$ & LPIPS $\downarrow$ & PSNR $\uparrow$ & SSIM $\uparrow$ & LPIPS $\downarrow$ & PSNR $\uparrow$ & SSIM $\uparrow$ & LPIPS $\downarrow$ \\
    \midrule 
    \multirow{5}{*}{NVS} &  NeILF~\cite{yao2022neilf} & 29.309 & 0.910 &  \textbf{0.187} & \textbf{29.651} & \textbf{0.944} & 0.095 & \textbf{34.653} & 0.959 & 0.099 & 26.509 & 0.783 & 0.339 \\  
    & I\textsuperscript{2}-SDF~\cite{ZhuHYLLXWTHBW2023} & 24.993 & 0.898 & 0.234 & 25.845 & 0.916 & 0.150 & 27.955 & \textbf{0.962} & \textbf{0.091} & 24.967 & 0.698 & 0.483 \\
    & FIPT-LDR* & 16.372& 0.776& 0.381& 14.536& 0.784& 0.389& 16.146& 0.805& 0.361& 13.665& 0.609& 0.616\\
    & Ours& \textbf{29.730} & \textbf{0.916} & 0.192 & 28.765 & 0.940 & \textbf{0.094} & 31.368 & 0.954 & 0.104 & \textbf{28.008} & \textbf{0.802} & \textbf{0.335} \\
    \cmidrule{2-14}
    & FIPT-HDR~\cite{wu2023factorized}& 29.059 & 0.924 & 0.180 & 27.670 & 0.940 & 0.095 & 28.524 & 0.951 & 0.109 & 29.788 & 0.792 & 0.358 \\
    \midrule 
    \multirow{4}{*}{Relight} & Li22~\cite{li2022physically} & 21.755 & 0.815 & 0.381 & 23.662 & 0.851 & 0.342 & \textbf{21.631} & 0.841 & 0.395 & 22.887 & 0.747 & 0.475 \\
    & FIPT-LDR* & 11.932 & 0.715 & 0.283 & 13.132 & 0.701 & 0.334 & 9.198 & 0.710 & 0.345 & 12.240 & 0.694 & 0.473 \\
    & Ours& \textbf{23.818} & \textbf{0.873} & \textbf{0.143} & \textbf{25.483} & \textbf{0.892} & \textbf{0.166} & 18.478 & \textbf{0.906} & \textbf{0.127} & \textbf{23.664} & \textbf{0.856} & \textbf{0.254} \\
    \cmidrule{2-14}
    & FIPT-HDR~\cite{wu2023factorized}& 27.597 & 0.886 & 0.115 & 28.411 & 0.878 & 0.155 & 32.543 & 0.964 & 0.078 & 27.497 & 0.881 & 0.208 \\
    \bottomrule 
    \end{tabular}%
    }
\end{table*}

\begin{figure}[t]
    \centering\setlength{\tabcolsep}{0.1em}
    \resizebox{1.0\linewidth}{!}{
    \begin{tabular}{@{}ccc@{}}
    IrisFormer~\cite{ZhuLMPC2022} & After initialization & After full optimization \\
    \frame{\includegraphics[width=0.4\linewidth]{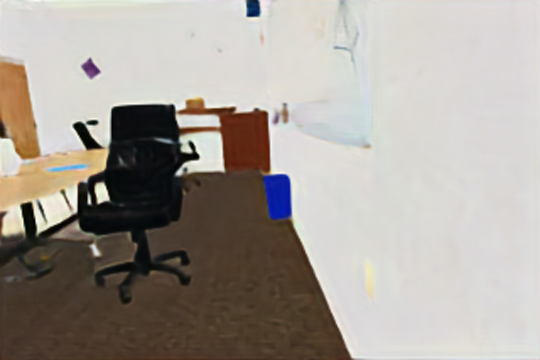}} &
    \frame{\includegraphics[width=0.4\linewidth]{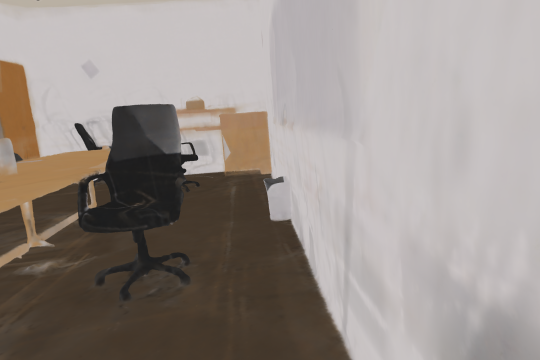}} &
    \frame{\includegraphics[width=0.4\linewidth]{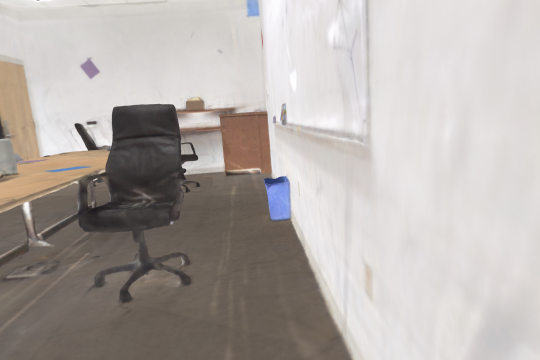}}
    \end{tabular}
    }
    \caption{\textbf{Visualizing albedo $\ba$ during the training.}
    We show that leveraging data-driven IRISformer \cite{ZhuLMPC2022} estimation (left) provides us good albedo initialization (center), and final result is refined with physically-based rendering model.}
    \label{fig:albedo_vis}
\end{figure}

In addition to physically-based inverse rendering techniques like FIPT, methods based on neural radiance fields (NeRF) \cite{mildenhall2020nerf} strive for scene disentanglement by representing indoor scenes' incident radiance fields with a 5D network \cite{yao2022neilf} without constraints.
Recent NeRF-based approaches like I\textsuperscript{2}-SDF \cite{ZhuHYLLXWTHBW2023}, NeILF++ \cite{zhang2023neilf++}, and NeFII \cite{wu2023nefii}, much like FIPT, rely on pre-calculated irradiance, and focus on surface rendering to reconstruct scene materials and/or lighting.
However, these methods typically account for only single-bounce light transport, leading to compromised quality in both material and lighting reconstruction.
The complete metrics of inverse rendering are shown in \cref{tab:syn_intrinsic_full} and the complete metrics of novel-view synthesis and relighting are listed in \cref{tab:nvs_synthetic_full}.
Our method achieves comparable novel-view synthesis results and outperforms other baselines for relighting.
The results underscore the effectiveness of our method in accurately decomposing intrinsic elements from LDR images.
As for computational efficiency, the whole training takes 57\,mins on a single RTX 4090, compared to 298\,mins for NeILF \cite{yao2022neilf} and 50\,mins for FIPT \cite{wu2023factorized}.

\section{Qualitative Results of Synthetic Scenes}
To verify the effectiveness of inverse rendering, we compare IRIS with several baselines on synthetic scenes provided by FIPT \cite{wu2023factorized}, which provide ground-truth geometry, material properties, and lighting.
\cref{fig:syn_intrinsic} shows the qualitative results of inverse rendering, including image reconstruction, material reflectance $\ba'$, roughness $\sigma$, and emission maps.
While NeILF \cite{yao2022neilf} achieves accurate rendering, it bakes significant shading effects into its diffuse albedo map.
\citet{li2022physically} generate a noisy BRDF from a single image input.
FIPT* tends to underestimate illumination intensity, overestimating the reflectance $\ba'$ as compensation.
In contrast, our method successfully recovers high-quality HDR emission from LDR input, resulting in precise intrinsic decomposition.

\section{Additional Ablation Study}

\begin{table}[t]
    \caption{\label{tab:ablation_crf}%
        \textbf{Ablation of CRF modeling.}
    }
  \vspace{-2mm}
    \centering\setlength{\tabcolsep}{4pt}
    \begin{tabular}{lccc@{\hspace{16pt}}c}
    \toprule
         & \multicolumn{3}{c}{PSNR $\uparrow$} & $L_2$ $\downarrow$ \\
        Method & $\bk_\text{d}$ & $\ba'$ & $\sigma$ & CRF  \\
	\midrule
    
    Constant exposure & 24.24 & 19.11 & \textbf{27.42} & 4.074  \\
    Mean CRF $\bar{g}$& 23.61 & 19.55 & 15.25 & 4.240 \\
    Gamma $1/2.2$     & 23.65 & 20.05 & 15.72 & 3.683 \\
    Full model & \textbf{26.82} & \textbf{23.43} & 26.63 & \textbf{1.363}  \\
    \bottomrule
    \end{tabular}%
\end{table}

\begin{figure}[t]
    \centering
    \includegraphics[width=1.0\linewidth]{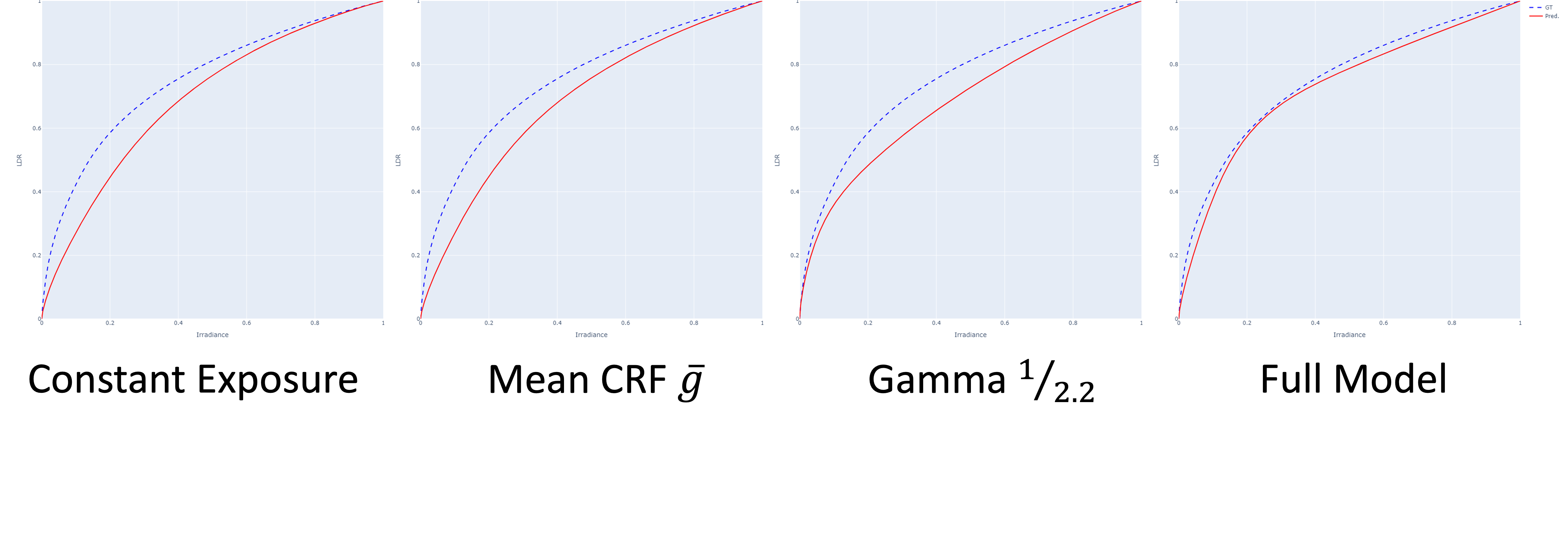}
    \vspace{-5mm}
    \caption{\textbf{CRF comparison visualization.}
    The blue dash lines are the ground-truth CRF, and the red lines are the estimated CRF after the optimization.
    We compare with three variants of CRF modeling settings. 
    We show that the full model with varying exposure and learnable CRF model can approximate the ground truth quite well.
    }\label{fig:crf_vis}
\end{figure}

Our method explicitly models the HDR–LDR conversion and estimates the CRF from input images, and thus achieves better inverse rendering quality.
To further validate the design choices, we conduct an ablation study on the CRF modeling strategy and evaluate inverse rendering from input images with varying exposure levels, which is collected with the strategy described in \cref{sec:details}.
We visualize the CRFs estimation with different modeling techniques in \cref{fig:crf_vis}, corresponding to \cref{tab:ablation_crf}.
The results show that from input images captured with varying exposure, our method can recover ground-truth CRF, demonstrating the effectiveness and importance of CRF modeling.
We parametrize the CRF as a continuous and monotonically increasing function across the domain $(0, 1)$, sample 1024 points between 0 and 1, and calculate the L2 distance between the function values and the ground truth.
We compare with three CRF alternatives: (1) constant exposure input, (2) Mean CRF $\bar{g}$ (the mean CRF from 201 empirical CRF functions measured in the real world~\cite{GrossN2004}), and (3) Gamma $1/2.2$ ($g(x) = x^{1/2.2}$, as used in FIPT \cite{wu2023factorized}).
Our method outperforms the single exposure approach, suggesting the benefits of using varying exposure values to enhance dynamic range.
It also achieves better results than constant CRF functions, justifying joint CRF optimization's merits.

\section{Factorized Light Transport}
We follow the rendering equation \cite{kajiya1986rendering} to model physically-based light transport for realistic rendering:
\begin{equation} \label{eq:rendering-supp}
        \bL_\text{o}(\bx, \bomega_\text{o}) \hspace{-.2em}=\hspace{-.2em}\bL_\text{e}(\bx, \bomega_\text{o}) + \hspace{-.2em}\int_{\Omega^+}\hspace{-.5em}\bL_\text{i}(\bx, \bomega_\text{i})f(\bx, \bomega_\text{i}, \bomega_\text{o})d\bomega_\text{i} \text{,}
\end{equation}
where $\bL_\text{o}$ is the radiance observed along a ray $(\bx, \bomega_\text{o})$ for a 3D position $\bx$ and a direction $\bomega_\text{o}$,
$\bL_\text{e}$ is the emission term,
$\bL_\text{r} = \int_{\Omega^+}\bL_\text{i}(\bx, \omega_\text{i})f(\bx, \bomega_\text{i}, \bomega_\text{o})d\bomega_\text{i}$ is the reflectance term, and
$f(\bx, \bomega_\text{i}, \bomega_\text{o})$ is the BRDF.
While $\bL_\text{i}$ encapsulates recursive incident radiance computation, we represent spatially-varying materials using the Cook–Torrance BRDF \cite{cook1982reflectance}:
\begin{align}\label{eq:brdf}
    f(\bx, \bomega_\text{i}, \bomega_\text{o}) = \frac{\bk_\text{d}(\bx)}{\pi}(\bn \cdot \bomega_\text{i})_{+} + 
    \frac{F \cdot D \cdot G}{4(\bn \cdot \bomega_\text{o})} \text{,}
\end{align}
where $D(\bh, \bn, \sigma(\bx))$ describes the distribution of microfacet orientations, 
$G(\bomega_\text{i}, \bomega_\text{o}, \bn, \sigma(\bx))$ encodes the masking and shadowing effects between microfacets, and
$F(\bomega_\text{i}, \bh, \bk_\text{s}(\bx))$ is the Fresnel reflection term.
The recursive integral in \cref{eq:rendering-supp} is computationally expensive and usually approximated with Monte–Carlo path tracing \cite{kajiya1986rendering, lafortune1996mathematical} with multiple bounces.
The rendering equation can be accelerated by factorizing the BRDF term from the integral \cite{krivanek2022practical, seyb2020design, wang2021learning}:
\begin{equation}
    \bL_\text{r}(\bx, \bomega_\text{o}) \!=\!
    \bk_\text{d}\bL_\text{d}(\bx) \!+\!
    \bk_\text{s}\bL_\text{s}^0(\bx, \bomega_\text{o},\sigma) \!+\!
    \bL_\text{s}^1(\bx, \bomega_\text{o},\sigma) \text{,}
\end{equation}
where we decompose the reflectance term into a diffuse shading term $\bL_\text{d}(\bx) \!=\! \int_{\Omega^+}\bL_\text{i}(\bx, \bomega_\text{i})\frac{(\bn\cdot\bomega_\text{i})_+}{\pi}d\bomega_\text{i}$, as well as two specular terms
\begin{align}
    \bL_\text{s}^0(\bx, \bomega_\text{o},\sigma) &= \int_{\Omega^+}\bL_\text{i}(\bx, \bomega_\text{i})\frac{F_0DG}{4(\bn\cdot \bomega_\text{o})}d\bomega_\text{i} \text{,} \\
    \bL_\text{s}^1(\bx, \bomega_\text{o},\sigma) &= \int_{\Omega^+}\bL_\text{i}(\bx, \bomega_\text{i})\frac{F_1DG}{4(\bn\cdot \bomega_\text{o})}d\bomega_\text{i} \text{,}
\end{align}
where $F_0 \!=\! 1 \!-\! F_1$ and $F_1 \!=\! (1 \!-\! \bh\cdot\bomega_\text{i})^5$.
$\bk_\text{d}(\bx)$ is diffuse reflectance and $\bk_\text{s}(\bx)$ is specular reflectance calculated from the BRDF.
$\bL_\text{s}^*$ is further approximated by linearly interpolating the shading maps pre-computed at various roughness $\sigma$ levels: 
$\bL_{s}^*(\cdot,\sigma) = \texttt{LERP}(\{ \bL_{s}^*(\cdot,\sigma_i) \}_{i=1}^{6}, \sigma )$,
where $\{\sigma_i\}_{i=1}^{6}$ is uniformly sampled between $(0, 1)$.
With the factorization formulation, the shading maps $\bL_\text{d}, \{\bL_\text{s}^0(\cdot, \sigma_i), \bL_\text{s}^1(\cdot, \sigma_i)\}_{i=1}^{6}$ can be pre-computed and allow for more efficient and stable optimization of material properties and HDR lighting.

\begin{table}
	\caption{\label{tab:notations}%
		\textbf{Notation table.}
	}
    \centering
    \resizebox{1.0\linewidth}{!}{%
    \begin{tabular}{cl}
    \toprule
        Symbol & Description \\ \midrule
       $(\cdot)_{+}$ & dot product clamped to positive value \\
       $\bomega_\text{i}$  & incident light direction \\
       $\bomega_\text{o}$  & outgoing light direction \\
       $\mathbf{h}$ & half vector $(\bomega_\text{i}+\bomega_\text{o})/\|\bomega_i+\bomega_o\|_2$\\
       $\mathbf{n}$  & surface normal \\
       $\bx$ & 3D position \\
       $\ba(\bx)$ & surface albedo (base color) \\
       $m(\bx)$ & surface metallicness \\
       $\sigma(\bx)$ & surface roughness \\
       $\bk_\text{d}(\bx)$ & diffuse reflectivity $\ba(\bx)(1 - m(\bx))$ \\
       $\bk_\text{s}(\bx)$ & specular reflectivity $\ba(\bx)m(\bx) + 0.04(1 - m(\bx))$ \\
       $D(\cdot)$ & GGX normal distribution \\
       $F(\cdot)$ & Schlick's approximation of Fresnel coefficients \\
       $G(\cdot)$ & Geometry (shadow-masking) term \\
    \bottomrule
    \end{tabular}
    }
\end{table}

\section{Implementation Details}\label{sec:details}
To clarify the equations in the paper, we describe the mathematical expressions and associated physical meanings in \cref{tab:notations}.

\paragraph{Varying exposure data generation}
In real-world photography pipelines, exposure levels are adjusted by manipulating camera settings, such as shutter speed, aperture size, and ISO, to capture bright and dark regions.
While the FIPT dataset \cite{wu2023factorized} assumes single exposure and utilizes a simplistic camera response function (CRF) model defined as $\texttt{CRF}(x) = x^{1/2.2}$, our approach simulates a capturing process that is both more realistic and challenging.
To create LDR images of synthetic scenes for CRF metric calculation, we split the HDR images of the same scene into five exposure levels $\{\Delta t_i\}_{i=1}^5, \text{s.t. } \Delta t_i < \Delta t_{i+1}$, where the brightest HDR image corresponds to $\Delta t_0$, and conversely, the darkest to $\Delta t_5$, effectively mimicking an auto-exposure mechanism.
Subsequently, we apply each exposure level to the HDR image and convert it into LDR format with the CRF derived from real-world sensors \cite{GrossN2004}.

\paragraph{Direct illumination $\bL_\text{e}(\bx)$} %
We first identify the mesh faces $\{\mathbf{f}_i\}$ of emitters with the emitter mask $M_\text{e}(\bx)$ defined on the mesh faces.
We associate a learnable 3-dimensional parameter for each face: $\be(\mathbf{f}) \in \bbR^3$, representing the emitted light radiance.
These parameters are then optimized during the HDR emission restoration phase.

\paragraph{BRDF} %
The surface material is represented as a neural field: $(\ba, m, \sigma) = \bF(\bx)$, the model architecture of which is based on Instant-NGP \cite{muller2022instant}.

\paragraph{Shading Baking}

Intuitively, the ray tracing continues if it encounters a non-emissive, specular surface (identified by a roughness threshold of 0.6), and stops otherwise.
The radiance at the endpoint adheres to Eq. 11 %
in the main paper.
The view-independent term $\bL_\text{SLF}(\bx)$ effectively approximates the global illumination on diffuse surfaces, which also expedites the rendering process.
This is because rays typically reach diffuse surfaces within a few bounces, eliminating the need for further path tracing:
\begin{equation}
\begin{split}
    \bL_{\text{i}}(\bx, \bomega) &= \bL_{\text{end}}(\bx_n)\displaystyle\prod_{i=1}^{n-1}f(\bx_{i+1} \to \bx_i) \\
    \text{s.t. } \sigma(\bx_i) &\leq 0.6,  M_\text{e}(\bx_i)=0, \forall i < n \text{,}
\end{split}
\end{equation}
where $\{\bx_i\}_{i=1}^n$ are the intersected points along the paths.

\begin{figure}[t]
    \centering\setlength{\tabcolsep}{0.1em}
    \resizebox{1.0\linewidth}{!}{
    \begin{tabular}{lcccc}
       & Reconstruction & Reflectance $\ba'$ & Roughness $\sigma$ & Emission map\\
       
    \raisebox{1.0\normalbaselineskip}[0pt][0pt]{\rotatebox[origin=c]{90}{GT}} & \frame{\includegraphics[width=0.25\linewidth]{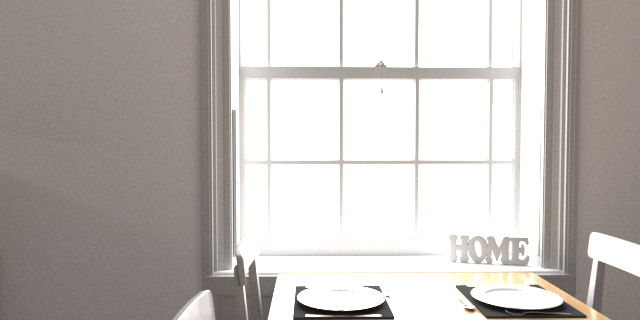}} & \frame{\includegraphics[width=0.25\linewidth]{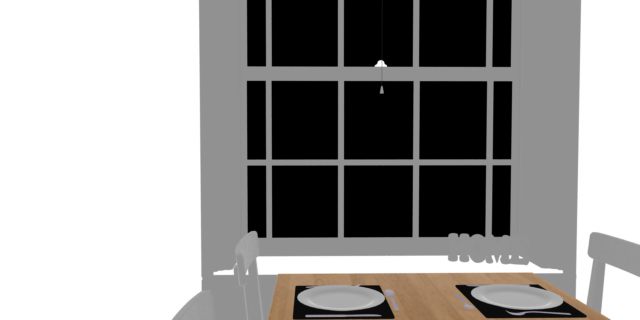}} & \frame{\includegraphics[width=0.25\linewidth]{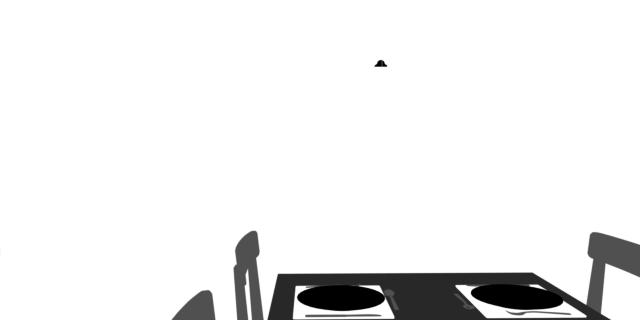}} &\frame{\includegraphics[width=0.25\linewidth]{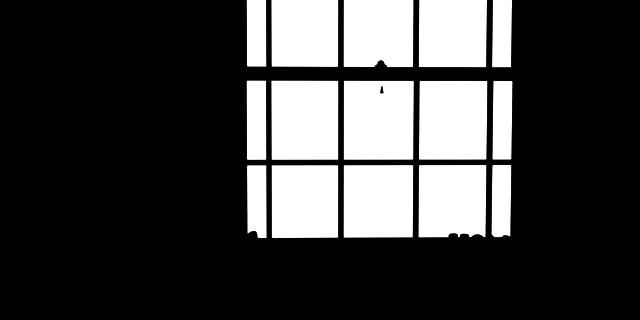}}  \\
    
    \raisebox{1.0\normalbaselineskip}[0pt][0pt]{\rotatebox[origin=c]{90}{Ours}} & \frame{\includegraphics[width=0.25\linewidth]{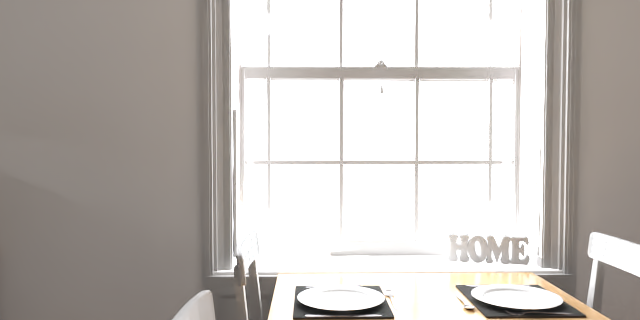}} & \frame{\includegraphics[width=0.25\linewidth]{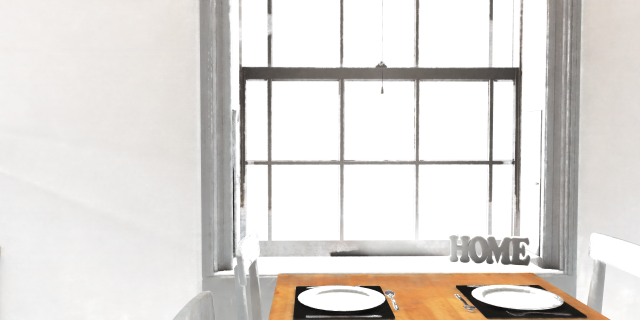}} & \frame{\includegraphics[width=0.25\linewidth]{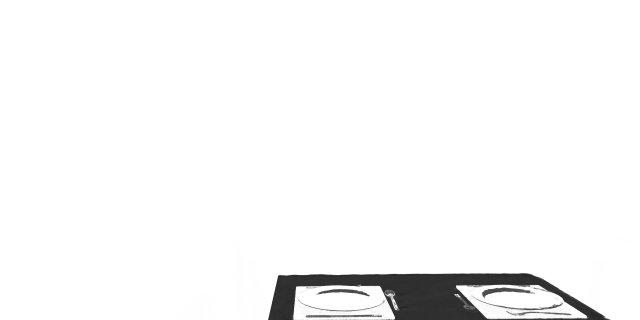}} & \frame{\includegraphics[width=0.25\linewidth]{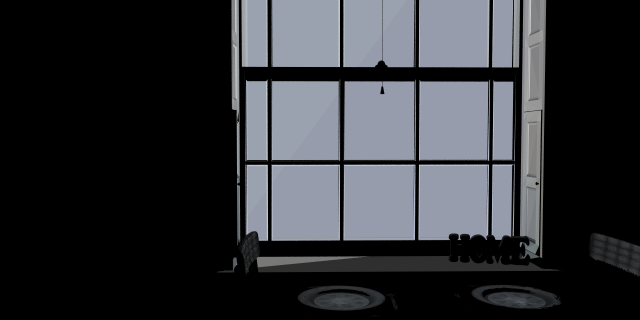}}  \\
      
    \end{tabular}
    }
    \caption{\textbf{Failure cases.}}
    \label{tab:failure}
\end{figure}

\section{Limitations}
Our emitter mask estimation may be inaccurate, especially when images are largely saturated.
An incorrect mask cannot be recovered from as masks are not further optimized.
Our CRF model is global, and it cannot capture complex non-local tone-mapping or while-balance changes.
Addressing these issues would allow for a truly practical method for inverse rendering, which is left for future work.

\end{document}